\documentclass[10pt,twocolumn,letterpaper]{article}

\usepackage[pagenumbers]{cvpr} 

\usepackage{booktabs}
\usepackage{graphicx}

\usepackage{amsmath,amsthm,amssymb}
\usepackage{color}
\usepackage{subcaption}
\usepackage{pifont}
\usepackage{diagbox}
\usepackage{multirow}
\usepackage{enumitem}
\usepackage{algorithm}
\usepackage{algorithmic}
\usepackage{wrapfig}
\usepackage{array}
\usepackage{balance}

\newcolumntype{x}[1]{>{\centering\arraybackslash\hspace{0pt}}p{#1}}
\DeclareMathOperator*{\argmin}{arg\,min}

\usepackage{xcolor}         
\definecolor{ultramarine}{RGB}{18, 10, 143}
\usepackage[pagebackref=true,breaklinks=true,letterpaper=true,colorlinks,bookmarks=false,citecolor=blue,linkcolor=red]{hyperref}

\newcommand{\bfsection}[1]{\noindent\textbf{#1.}}

\usepackage[capitalize]{cleveref}
\crefname{section}{Sec.}{Secs.}
\Crefname{section}{Section}{Sections}
\Crefname{table}{Table}{Tables}
\crefname{table}{Tab.}{Tabs.}

\def\cvprPaperID{***} 
\def\confName{CVPR}
\def\confYear{2023}

\begin{document}

\title{QuantArt: Quantizing Image Style Transfer Towards High Visual Fidelity}

\author{Siyu Huang$^{1}$\thanks{} \qquad
Jie An$^{2}$\footnotemark[1] \qquad
Donglai Wei$^{3}$ \qquad
Jiebo Luo$^{2}$ \qquad
Hanspeter Pfister$^{1}$ \\
$^1$Harvard University \quad
$^2$University of Rochester \quad
$^3$Boston College
\\
{\tt\small \{huang,pfister\}@seas.harvard.com} \quad
{\tt\small \{jan6,jluo\}@cs.rochester.edu} \quad 
{\tt\small donglai.wei@bc.edu} 
}

\twocolumn[{%
\renewcommand\twocolumn[1][]{#1}%
\maketitle
\begin{center}
    \centering
    \captionsetup{type=figure}
    \vspace{-2.5em}
    \includegraphics[width=1\linewidth]{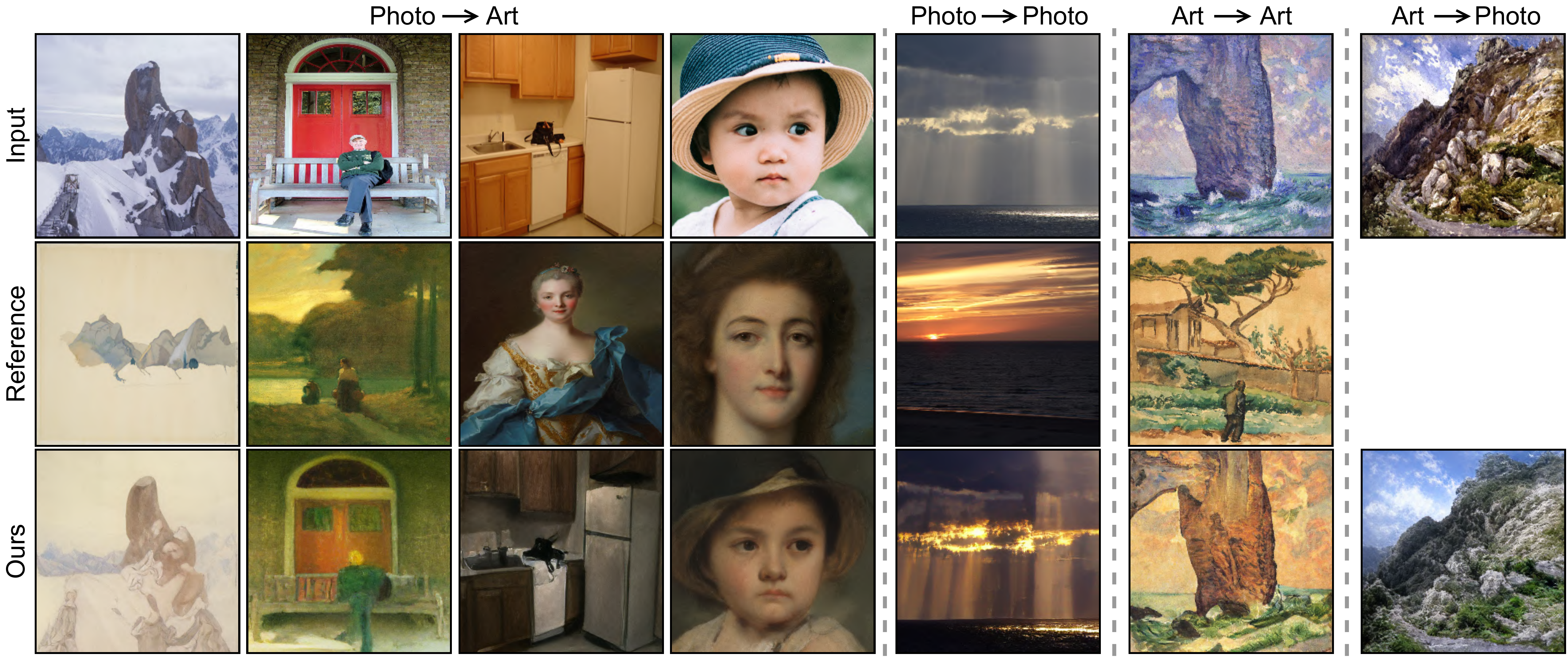}
    \vspace{-2em}
    \captionof{figure}{The proposed QuantArt method produces impressive arbitrary style transfer results on various image style transfer tasks.}
    \label{fig:teaser}
\end{center}%
}]

\begin{abstract}
The mechanism of existing style transfer algorithms is by minimizing a hybrid loss function to push the generated image toward high similarities in both content and style. However, this type of approach cannot guarantee visual fidelity, i.e., the generated artworks should be indistinguishable from real ones. In this paper, we devise a new style transfer framework called QuantArt for high visual-fidelity stylization. QuantArt pushes the latent representation of the generated artwork toward the centroids of the real artwork distribution with vector quantization. By fusing the quantized and continuous latent representations, QuantArt allows flexible control over the generated artworks in terms of content preservation, style similarity, and visual fidelity. Experiments on various style transfer settings show that our QuantArt framework achieves significantly higher visual fidelity compared with the existing style transfer methods.
\end{abstract}

\vspace{-1em}

\section{Introduction}
Image style transfer aims at transferring the artistic style of a reference image to a content image, where the output image should have the style (\eg, colors, textures, strokes, and tones) of the reference and the content information of the content image. Great advances~\cite{gatys2016image,gatys2017controlling,chen2017stylebank,adaattn,luan2017deep,chen2021diverse,zhang2022exact} have been made in the area of image style transfer, where the arbitrary style transfer (AST) has become one of the main research focuses. Given a trained model, AST algorithms~\cite{adain,linearWCT,stytr2} can perform style transfer on arbitrary unseen content-style pairs in a zero-shot manner, such that it enables more practical applications\footnote{The codes of this paper are available at \url{https://github.com/siyuhuang/QuantArt}}.

Existing AST algorithms, including the statistics-based methods~\cite{adain,wct,wct2,artflow} and the patch-based methods~\cite{styleswap,avatar}, deliver remarkable style transfer results by matching the artistic style information of the stylized image and the style reference. However, taking the high-fidelity artwork generation as the ultimate goal of image style transfer, all existing methods can still be improved since there are few mechanisms to guarantee a high artistic fidelity of the stylized image. A few existing work~\cite{chang2020domain,xu2021drb} accommodate the adversarial loss~\cite{gan} into the style transfer framework to enhance the image quality. However, the performance improvement is hindered by the heterogeneous optimization objectives of high image quality and faithful image stylization.

In this work, we introduce visual fidelity as a new evaluation dimension of style transfer. It is formulated as the similarity between the stylized image and the real artwork dataset, and it is orthogonal to the two widely studied evaluation dimensions including style similarity and content preservation. Motivated by the vector-quantized image representation~\cite{vqvae,vqvae2,vqgan}, if the latent feature of generation is closer to one of the cluster centers in the real distribution, it is harder for humans to distinguish it from the real images, \ie, having better visual fidelity. We propose to learn an artwork codebook, \ie, a global dictionary, to save the discrete cluster centers of all artworks. The continuous representations of images are converted to the discrete encodings in the artwork codebook via vector quantization, ensuring that it is not only close to the given style reference but also close to one of the learned cluster centers in the real distribution. 
  
We further propose a framework called Quantizing Artistic Style Transfer (QuantArt) to achieve flexible control of the three evaluation dimensions mentioned above. QuantArt first extracts both content and style features using separate encoders, respectively. Next, it applies vector quantization to both content and style features to fetch discrete codes in the learned codebooks. Then, the content and style codes are transferred to the stylized feature with a specially designed feature style transfer module called Style-Guided Attention. Before feeding into the decoder, the stylized feature is quantized again with the artwork codebook, ensuring a high visual-fidelity stylization by approaching the cluster centers of the real artwork distribution. By fusing the continuous and quantized stylized features with the content features before the decoder, QuantArt allows users to arbitrarily trade off between the style similarity, visual fidelity, and content reservation of the style transfer results.
In the experiments, the proposed method significantly increases the visual fidelity of generations in various image style transfer settings including photo-to-art, art-to-art, photo-to-photo, and art-to-photo (see Fig.~\ref{fig:teaser}).
The contribution of the proposed method can be summarized as follows:
\begin{itemize}
    \setlength\itemsep{0em}
    \item We define  {\it visual fidelity} as a new evaluation dimension of style transfer and propose a high visual-fidelity style transfer algorithm based on vector quantization.
    \item We design a framework based on both discrete and continuous style transfer architectures, which allow users to flexibly control style similarity, content preservation, and visual fidelity of the stylization result. 
    \item The extensive experiments demonstrate that our method achieves higher visual fidelity and comparable style similarity with respect to the state-of-the-art style transfer methods.
\end{itemize}

\section{Related Work}

\noindent  
\textbf{Image style transfer.}
Image style transfer is a challenging topic that has been studied for decades~\cite{hertzmann1998painterly,winnemoller2006real,li2017diversified,survey}. Gatys et al.~\cite{gatys2016image} first adopted convolutional neural networks (CNNs) for image style transfer by matching the statistics of content and style features extracted by CNNs. Among the neural style transfer (NST) algorithms~\cite{gatys2016image,gatys2017controlling,luan2017deep}, arbitrary style transfer (AST)~\cite{adain,styleswap,huang2022parameter,chen2021diverse,zhang2022exact} has drawn much attention from researchers in recent years due to its zero-shot image stylization manner. 

Existing AST algorithms can be generally categorized into two types: the statistics-based methods~\cite{in,cbn} and the patch-based methods~\cite{frigo2016split,liao2017visual,wang2022divswapper,gu2018arbitrary}. The statistics-based methods minimize the distance of global feature statistics between the generation and the style image, where the feature statistics can be Gram matrices~\cite{gatys2016image,gatys2017controlling}, histograms~\cite{heeger1995pyramid,histogramloss}, wavelets~\cite{portilla2000parametric,wct2}, mean-std statistics~\cite{adain,cin}, and covariance matrices~\cite{wct}. The statistics-based methods are highly efficient in capturing the global style information.
The patch-based methods search for appropriate patches in style images to reconstruct the transferred images. StyleSwap~\cite{styleswap} and Avatar-Net~\cite{avatar} are two typical patch-based methods, which iteratively swap the content feature patches with the nearest-matched feature patches of the reference image. 
Compared to the statistics-based approaches, the patch-based methods produce better texture synthesis quality as they directly adopt patches from style images. However, it requires the content and style images to have similar local semantic structures. 

In general, the existing AST algorithms aim at matching the styles of the generation and the reference, where the visual fidelity of the generation cannot be guaranteed. In this work, we introduce visual fidelity as a new evaluation dimension of style transfer and propose a novel AST framework, \ie, QuantArt, to enhance the visual fidelity of generations via pushing the latent feature toward the centroids of artwork distributions. QuantArt can also alleviate the stylization artifact issue, as the outlier styles are replaced with the nearest style centroid in the latent space. 

\noindent  
\textbf{Photorealistic style transfer.}
The proposed method can also handle the photorealistic style transfer task. The artistic style transfer algorithms often fail for this task, since the stylizated image would contain warping distortions that are redundant for the photorealism scenario. Motivated by this, several methods~\cite{an2020ultrafast,xia2020joint,xia2021real,dstn} have been specially designed. Luan et al.~\cite{luan2017deep} first introduced a locally affine transformation as the photorealism loss term. PhotoWCT~\cite{photowct} proposes a closed-form post-processing algorithm to further smooth the stylized results. WCT$^2$~\cite{wct2} eliminates the post-processing stage via the Wavelet Corrected Transfer module. Distinct from these approaches, our method does not require to impose any additional regularization or post-processing step for photorealistic style transfer, thanks to the highly effective quantized image representation. 

\noindent  
\textbf{Vector-quantized image representation.}
The vector-quantized generative models~\cite{vqgan,yu2021vector} are originally developed for compact yet effective image modeling. VQ-VAE~\cite{vqvae,vqvae2} devises a vector-quantized autoencoder to represent an image with a set of discrete tokens. VQGAN~\cite{vqgan} improves VQVAE with the adversarial learning scheme~\cite{gan}. This work adopts vector quantization as an efficient learnable implementation of artwork distribution clustering. The vector quantization pushes the latent feature to be closer to the real artwork distribution, resulting in higher visual-fidelity image stylization.

\section{Visual Fidelity for Image Style Transfer}

Image style transfer aims to transfer a content image $c$, \eg, a photo, to a stylized image $y$ by a given style reference $s$, \eg, an artwork image. Existing image style transfer algorithms mainly focus on either content preservation or style similarity between the generated and input images. However, we argue that the visual realness of generated images is also a vital factor for style transfer performance. We formulate the three critical performance indicators of image style transfer as

\begin{figure}[t]
    \centering
    \includegraphics[width=1\linewidth]{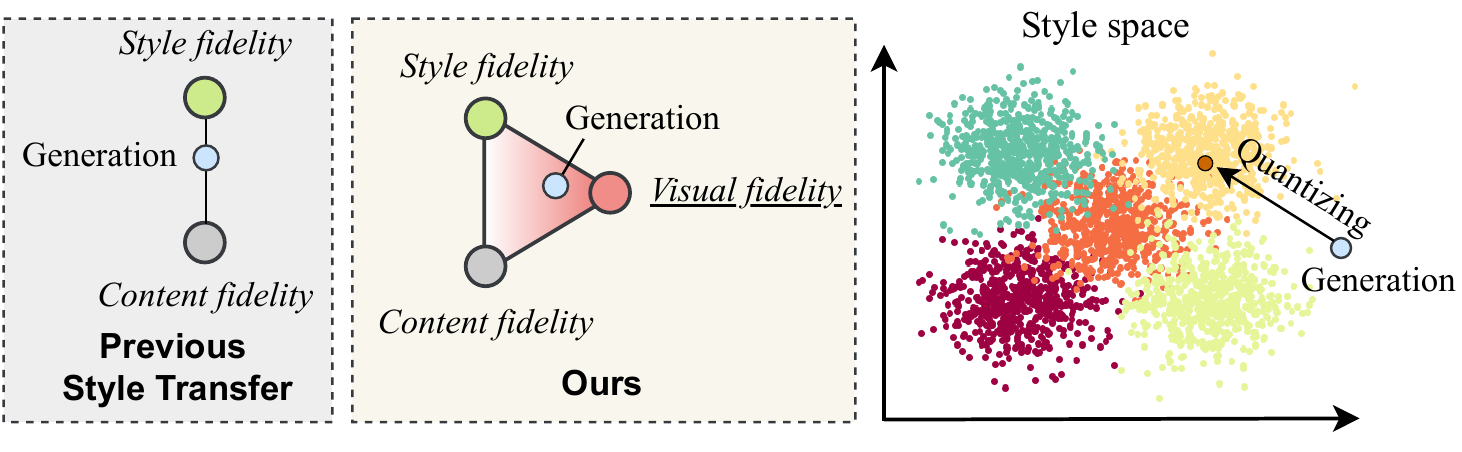} \\
    \includegraphics[width=1\linewidth]{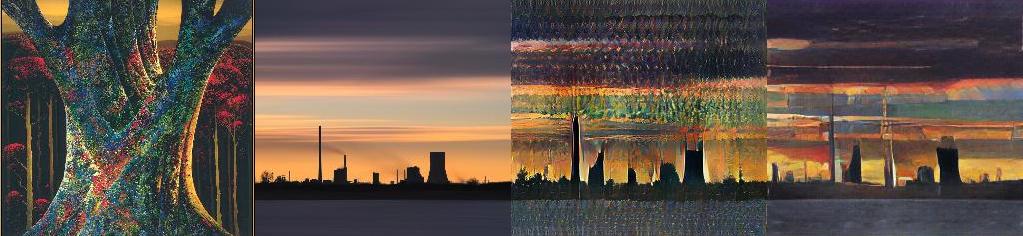}
    \vspace{-1.3em}\\
    \scriptsize
    \newcommand\w{0.2}
    \begin{tabular}    {x{\w\linewidth}x{\w\linewidth}x{\w\linewidth}x{\w\linewidth}}
     Reference & Input & \textit{w/o} Quantizing &  \textit{w/} Quantizing
    \end{tabular}
    \vspace{-1em}
    \caption{
    \textbf{(Top left)} This work introduces {\it visual fidelity} as an orthogonal evaluation dimension to content fidelity and style fidelity. 
    \textbf{(Top right)} Our style transfer method enables a trade-off between style and visual fidelities via quantizing the generation in the style space. 
    \textbf{(Bottom)} An example of image style transfer \textit{w/o} and \textit{w/} latent feature quantizing, respectively.
    }
    \vspace{-.5em}
    \label{fig:anneal}
\end{figure}

\begin{itemize}
    \setlength\itemsep{0em}
    \item \emph{Style fidelity}: The style similarity between $y$ and $s$, which is often evaluated by the Gram matrix~\cite{gatys2016image,li2017diversified}.
    \item \emph{Content fidelity}: The content similarity between $y$ and $c$, which is often evaluated by the perceptual loss~\cite{johnson2016perceptual} or the LPIPS distance~\cite{lpips}.
    \item \emph{Visual fidelity}: The realness of the generated image $y$. Since all real artwork images belong to a distribution $\mathcal{T}$, a generated image $y$ tends to have a higher visual fidelity if it is closer to distribution $\mathcal{T}$.
\end{itemize} 
Many existing literature~\cite{adain,avatar,sanet,kolkin2019style,adaattn} shows that there is a trade-off between the style fidelity and content fidelity in image style transfer. Analogous to this, there is a trade-off between visual fidelity and style fidelity. As an example shown in the bottom of Fig.~\ref{fig:anneal}, the neural style transfer algorithm ($\alpha=0$) faithfully renders the style textures of the tree in the reference image. However, it lowers the visual fidelity of the generated image. 

\begin{figure*}[t]
    \centering
    \begin{minipage}{0.325\linewidth}
    \centering
    \includegraphics[width=1\linewidth]{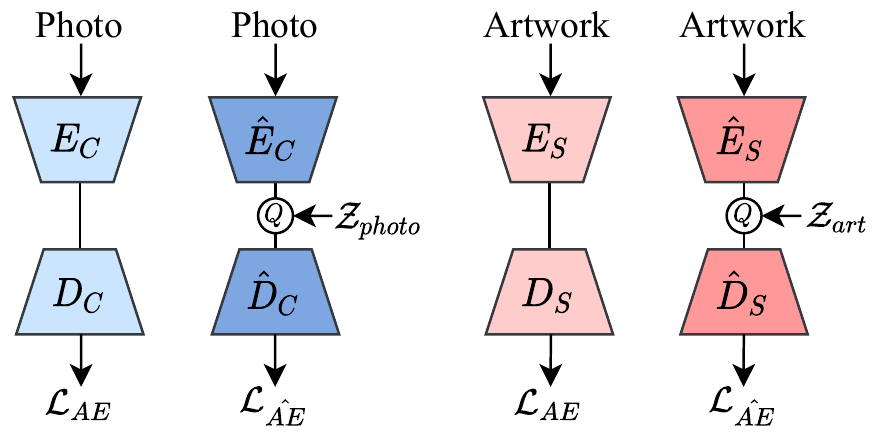}
    \vspace{-1.35em}
    \subcaption{\emph{Training Stage-1}: learning the auto-encoders}
    \vspace{0.3em}
    \hrule
    \vspace{0.2em}
    \includegraphics[width=1.14\linewidth]{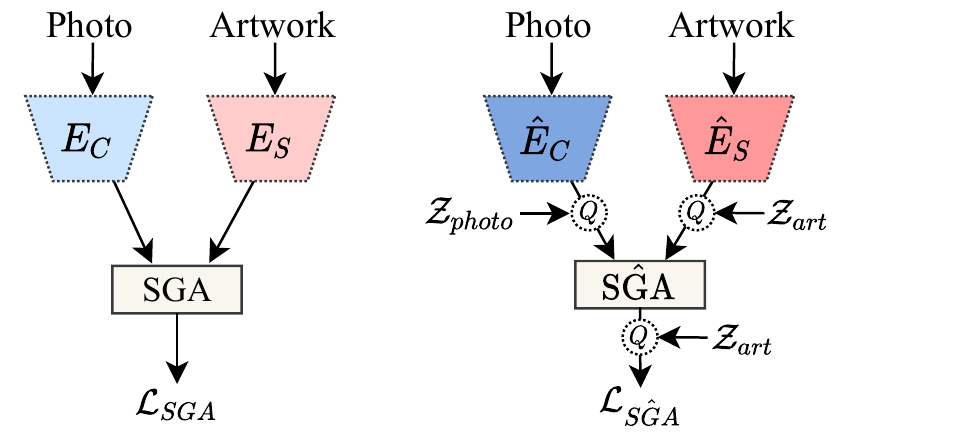}
    \vspace{-1em}
    \subcaption{\emph{Training Stage-2}: learning the SGA modules}
    \end{minipage}
    \vrule
    \hfill
    \begin{minipage}{0.655\linewidth}
    \hspace{0.1em}
    \includegraphics[width=1\linewidth]{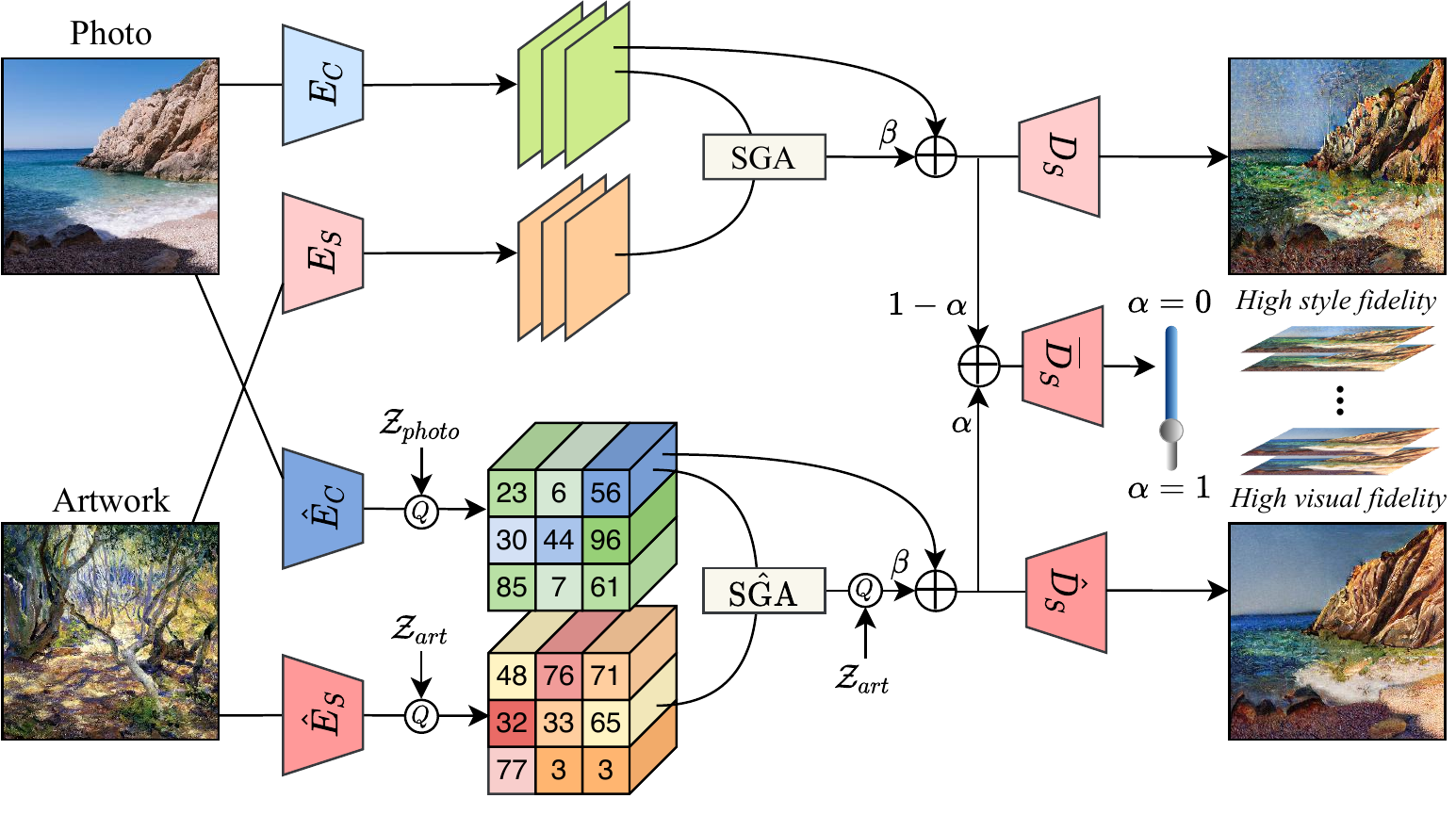}
    \subcaption{\emph{Inference Stage}}
    \end{minipage}
    \caption{The training and inference pipelines of QuantArt. 
    \textbf{(a)} The first training stage, where we learn the auto-encoders and codebooks for photo and artwork images, respectively.
        \textcircled{\scriptsize{$Q$}} denotes the vector quantization operator in Eq.~\ref{eq:vq}.
    \textbf{(b)} The second training stage, where we learn the SGA-based style transfer modules.
    The dashed lines denote the parameters of encoders and the codebooks are frozen in this stage.
    \textbf{(c)} In the inference phase, one can trade off the content, style and visual fidelities by simply adjusting the parameters $\alpha,\beta \in [0,1]$.
    }
    \vspace{-.5em}
    \label{fig:framework}
\end{figure*}

To increase the visual fidelity of neural style transfer, in this work we propose a novel framework named QuantArt to learn to cluster the artwork distribution $\mathcal{T}$ in the representation space, where the centroids of all clusters form an \emph{artwork codebook} $\mathcal{Z}_{art}$. When making inferences, we replace the feature map at each position with its nearest centroid in $\mathcal{Z}_{art}$. In this way, the feature of the generated artwork is pushed closer to the real artwork distribution, thus leading to better visual fidelity. This nearest centroid search and replacing operation is implemented by the vector quantization used in~\cite{vqvae,vqvae2,vqgan}. Besides, the idea of pushing the latent feature closer to the centroid of real distribution is partially motivated by the low-temperature sampling used in GANs~\cite{gan,stylegan,huang2020generating, zhou2019text} and diffusion models~\cite{ddpm,guo2022shadowdiffusion}. As illustrated in the top right part of Fig.~\ref{fig:anneal}, one can increase the visual fidelity by pushing the generation $y$ to be close to one of the centroids of the artwork distribution $\mathcal{T}$ but far away from the reference $s$. We introduce more details of QuantArt in the following Section~\ref{sec:method}.

\section{Our Approach}
\label{sec:method}
\subsection{Framework Overview}

In this work, we propose a novel framework dubbed Quantizing Artistic Style Transfer (QuantArt) to enhance the visual fidelity of generations in image style transfer. As illustrated in Fig.~\ref{fig:framework}, QuantArt adopts four auto-encoders to extract the continuous/quantized features of photo/artwork images respectively, two codebooks to store the cluster centers of photo and artwork distributions, and two SGA modules to transfer the styles of feature representations. The training of QuantArt consists of two stages. In the first training stage (see Fig.~\ref{fig:framework}(a)), we learn the auto-encoders and the codebooks by reconstructing the photo and artwork images, respectively. In the second training stage (see Fig.~\ref{fig:framework}(b)), we train the SGA modules based on the extracted feature representations. In the inference phase, as illustrated in Fig.~\ref{fig:framework}(c), users can easily trade off the style and visual fidelity of generations by adjusting the discretization level $\alpha \in [0,1]$ between the SGA outputs. In the following, we discuss more details of QuantArt.

\subsection{Learning Auto-Encoders and Codebooks}
\label{sec:vq}
We first extract the features of the content image $c$ and the style reference $s$ with two convolutional encoders $E_C$ and $E_S$, then decode the features back into images $c_{rec}$ and $s_{rec}$ with two convolutional decoders $D_C$ and $D_S$, respectively as
\begin{equation}
    c_{rec} = D_C(E_C(c)), ~~~~~ s_{rec} = D_S(E_S(s)).
\end{equation}
This is distinct from the general neural style transfer methods, which use the VGG~\cite{vgg} or ResNet~\cite{resnet} network pre-trained on natural image datasets (\eg, ImageNet~\cite{imagenet}) as the image encoder to extract features of both content and style images. 
To optimize encoder $E_C$ and decoder $D_C$, the reconstruction loss is
\begin{equation}
    \mathcal{L}_{AE}(E_C,D_C) = ||c_{rec} -c || + \mathcal{L}_{adv}(E_C,D_C,\mathbb{D}_C),
    \label{eq:ae}
\end{equation}
where $\mathcal{L}_{adv}$ is the adversarial training loss and $\mathbb{D}_C$ is the corresponding discriminator network, 
\begin{equation}
    \mathcal{L}_{adv}(E_C, D_C, \mathbb{D}_C) = \log \mathbb{D}_C(c) + \log(1-\mathbb{D}_C(c_{rec})).
   \label{eq:GANloss}
\end{equation}
$E_S$ and $D_S$ are also optimized by the reconstruction loss $\mathcal{L}_{AE}(E_S, D_S)$ as formulated in Eq.~\ref{eq:ae}. 

\begin{figure}[t]
    \centering
    \includegraphics[width=0.85\linewidth]{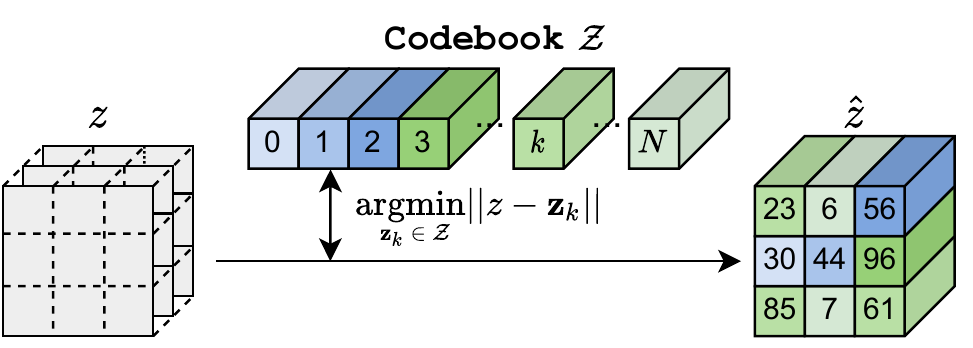}
    \caption{The vector quantization operator $Q_\mathcal{Z}(z)$ in Eq.~\ref{eq:vq}.
    }
    \vspace{-.5em}
    \label{fig:vq}
\end{figure}

Next, we build two codebooks $\mathcal{Z}_{photo},\mathcal{Z}_{art} \in \mathbb{R}^{N \times d}$ to model the distributions of the photo dataset and artwork dataset, where $N$ is the number of entries in the codebook and $d$ is the dimension of each entry. To enable a better representation performance of the quantized features, we use two extra encoders to extract the features, respectively as
\begin{equation}
    z_c = \hat{E}_C(c), ~~~~~~~ z_s = \hat{E}_S(s).
\end{equation}
We then apply vector quantization~\cite{vqvae} to the latent features to get the quantized features $\hat{z}_c$ and $\hat{z}_s$, where the vector quantization operator $Q_{\mathcal{Z}}(z)$ is formulated as
\begin{equation} 
    Q_{\mathcal{Z}}(z):= \argmin_{\mathbf{z}_k \in \mathcal{Z}} || z - \mathbf{z}_k||,
    \label{eq:vq}
\end{equation}
where $\mathbf{z}_k$ is the $k$-th entry in the codebook $\mathcal{Z}$. As illustrated in Fig.~\ref{fig:vq}, $z$ is replaced with the nearest neighbour entry in the codebook $\mathcal{Z}$ via $Q_{\mathcal{Z}}(z)$. 
The quantized features $\hat{z}_c$ and $\hat{z}_s$ are collected from codebooks $\mathcal{Z}_{photo}$ and $\mathcal{Z}_{art}$ as
\begin{equation} 
    \hat{z}_c = Q_{\mathcal{Z}_{photo}}(z_c), ~~~~~ \hat{z}_s = Q_{\mathcal{Z}_{art}}(z_s), 
\end{equation}
The quantized features $\hat{z}_c$ and $\hat{z}_s$ are decoded into images $\hat{c}_{rec}$ and  $\hat{s}_{rec}$ via the decoders $\hat{D}_C$ and $\hat{D}_S$, as
\begin{equation}
    \hat{c}_{rec} = \hat{D}_C(\hat{z}_c), ~~~~~ \hat{s}_{rec} = \hat{D}_S(\hat{z}_s).
\end{equation}
By following~\cite{vqvae}, we optimize the codebook $\mathcal{Z}_{photo}$ jointly with the reconstruction loss in Eq.~\ref{eq:ae}. The reconstruction loss for encoder $\hat{E}_C$, decoder $\hat{D}_C$, and codebook $\mathcal{Z}_{photo}$ is
\begin{align}
    \label{eq:aehat}
    \mathcal{L}_{\hat{AE}}(\hat{E}_C,\hat{D}_C,\mathcal{Z}_{photo}) & =  \mathcal{L}_{AE}(\hat{E}_C,\hat{D}_C)~+ \\
    & ||\mathtt{sg}[z_c]-\hat{z}_c]|| + ||\mathtt{sg}[\hat{z}_c]-z_c]||, \notag
\end{align}
where $\mathtt{sg}[\cdot]$ indicates the stop gradient operator. The second term in Eq.~\ref{eq:aehat} optimizes the codebook $\mathcal{Z}_{photo}$, while, the third term in Eq.~\ref{eq:aehat} forces the latent feature $z_c$ to be close to the nearest neighbor entry in $\mathcal{Z}_{photo}$. $\hat{E}_S$, $\hat{D}_S$, and $\mathcal{Z}_{art}$ are also optimized by the loss function $\mathcal{L}_{\hat{AE}} (\hat{E}_S, \hat{D}_S, \mathcal{Z}_{art})$ as formulated in Eq.~\ref{eq:aehat}.

After training the Stage-1 models, we can compute the continuous features and quantized features for the content and style images. In the following Section~\ref{sec:sga}, we discuss how to perform feature-level style transfer based on the extracted features.

\begin{figure}[t]
    \centering
    \includegraphics[width=0.75\linewidth]{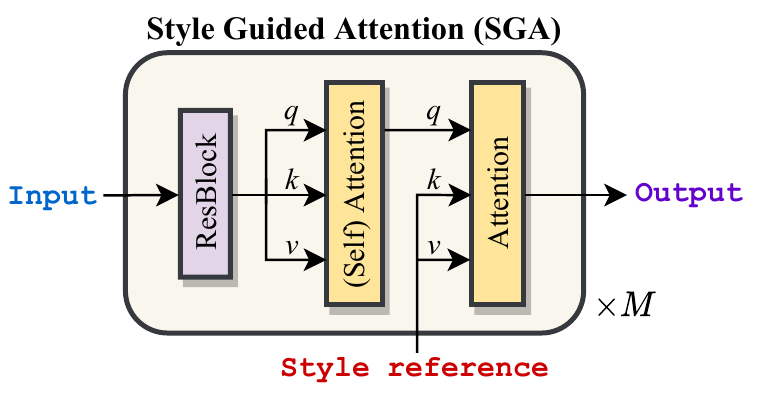}
    \vspace{-.6em}
    \caption{The SGA module proposed for feature style transfer.
    }
    \vspace{-.6em}
    \label{fig:sga}
\end{figure}

\subsection{Style Transfer with Style Guided Attention}
\label{sec:sga}

To perform effective style transfer for both the continuous and quantized features, we propose a feature-level style transfer module dubbed style-guided attention (SGA). Fig.~\ref{fig:sga} shows an illustration of the SGA module. The module takes the content feature $z_c \in \mathbb{R}^d$ and style reference feature $z_s \in \mathbb{R}^d$ as inputs, then outputs the stylized feature vector $z_y \in \mathbb{R}^d$. It consists of three blocks including a ResBlock used in ResNet-18~\cite{resnet} and two attention blocks~\cite{transformer} each with a residual connection. The attention block accepts the query $q$, key $k$, and value $v$ as inputs,
\begin{equation}
    \text{Attn}(q,k,v) = \text{softmax} (f_q(q)f_k(k))f_v(v)+q,
    \label{eq:attention}
\end{equation}
where $f_q,f_k,f_v$ are the embedding layers. 
The first attention block is a self-attention block that all its input heads $q,k,v$ accept the transformed content feature $\tilde{z}_c = \text{ResBlock}(z_c)$. The second attention block takes $\tilde{z}_c$ as $q$, and the style reference feature $z_s$ as $k$ and $v$. The output of the SGA module is formulated as
\begin{equation}
    z_y = \text{SGA}(z_c, z_s) = \text{Attn}(\text{Attn}(\tilde{z}_c,\tilde{z}_c,\tilde{z}_c),z_s,z_s).
\end{equation}
Compared with the existing attention-based style transfer modules~\cite{sanet,adaattn}, our SGA module adopts an additional self-attention block to model the global information of the quantized codes. Our attention block also enables a residual connection between the input and output to better preserve the content details. In practice, we repeat the SGA module for $M$ times to achieve a better style transfer performance.

The objective function for $\text{SGA}(z_c, z_s)$ includes the content loss $\mathcal{L}_{content}$, style loss $\mathcal{L}_{style}$, and adversarial training loss $\mathcal{L}_{featadv}$
\begin{equation}
    \mathcal{L}_{SGA} = \mathcal{L}_{content} + \mathcal{L}_{style} + \mathcal{L}_{featadv}
    \label{eq:sgaloss}
\end{equation}
where
\begin{equation}
    \mathcal{L}_{content} = ||z_y-z_c||,
\end{equation}
\begin{equation}
    \mathcal{L}_{style} = || \mu(z_y) - \mu(z_s)|| + || \sigma(z_y) - \sigma(z_s) ||,
\end{equation}
\begin{equation}
    \mathcal{L}_{featadv} = \log \mathbb{D}_{SGA}(z_s) + \log(1-\mathbb{D}_{SGA}(z_y)).
\end{equation}
$\mu(\cdot)$ and $\sigma(\cdot)$ denote the channel-wise mean and standard deviation of feature maps, respectively. Note that since the encoders and decoders are not optimized in Stage-2, the content and style losses can be directly computed on the features without using an extra pre-trained network to extract features~\cite{gatys2016image,johnson2016perceptual,adain}. The adversarial loss $\mathcal{L}_{featadv}$ is also computed on the features, and it forces the SGA output to be more close to the distribution of style reference features.

We adopt another SGA module to transfer the quantized features $\hat{z}_c$ and $\hat{z}_s$, as 
\begin{equation}
    \hat{z}_y = Q_{\mathcal{Z}_{art}}(\hat{\text{SGA}}(\hat{z}_c, \hat{z}_s)).
    \label{eq:sgahat}
\end{equation}
$\hat{\text{SGA}}(\cdot,\cdot)$ and $\text{SGA}(\cdot,\cdot)$ have the same network architecture but different parameters. The output of the SGA module is further quantized by the art codebook $Q_{\mathcal{Z}_{art}}$ to ensure that the output is on the latent space of the decoder $\hat{D}_S$. 
The optimization objective of $\hat{\text{SGA}}$ follows Eq.~\ref{eq:sgaloss} with an additional codebook loss, as
\begin{equation}
    \mathcal{L}_{\hat{SGA}} = \mathcal{L}_{SGA} + ||\text{sg}[\hat{z}_y]-z_y]||
\end{equation}

\subsection{Inference} 

Fig.~\ref{fig:framework}(c) illustrates the inference procedure of QuantArt framework. We first extract the continuous and quantized features of input photo and artwork images using the corresponding encoders. Then, the features are transformed to the stylized continuous feature $z_y$ and stylized quantized feature $\hat{z}_y$ by using the corresponding SGA modules. 
To trade off between the content, style reference, and visual fidelity, we have 
\begin{equation}
    z_{test} = \oplus_{\alpha} (\oplus_{\beta}(\hat{z}_y,\hat{z}_c), \oplus_{\beta}(z_y,z_c)),
    \label{eq:interpolatealphabeta}
\end{equation}
where $z_c$ and $\hat{z}_c$ are the content features, and $\oplus$ is the weighted sum operator as $\oplus_p (a,b)=pa+(1-p)b$. $z_{test}$ is decoded into a stylized image with a fused decoder $\bar{D}_S=\oplus_{\alpha}(\hat{D}_S,D_S)$. $\beta$ controls the style fidelity, and a larger $\alpha$ results in higher visual fidelity. In practice, a good trade-off of the fidelity terms depends on both the input images and individual users' preferences. QuantArt($\alpha,\beta$) provides a simple and easy-to-understand handle grip for users to adjust the style transfer results.

\bfsection{Extension to more style transfer tasks} 
In Section~\ref{sec:sga}, we take the artistic image style transfer task as an example to discuss the learning of SGA modules. In addition to artistic style transfer, QuantArt can be applied to other image style transfer tasks such as artwork-to-artwork style transfer and photorealistic style transfer, via training the SGA modules on the basis of the learned auto-encoders and codebooks discussed in Section~\ref{sec:vq}.

\section{Experiments}
\label{sec:experiment}

\subsection{Experiment Settings}
\bfsection{Dataset} 
For photo-artwork, photo-photo, and artwork-artwork style transfer tasks, we use the MS-COCO~\cite{mscoco} as the photo dataset and the WikiArt~\cite{wikiart} as the artwork dataset, by following the existing artistic/photorealistic style transfer methods~\cite{adain,artflow,adaattn,photowct}. For face-to-artwork style transfer, we use the FFHQ~\cite{stylegan} as the photo dataset and the MetFaces~\cite{metfaces} as the artwork dataset. 
For artwork-to-photo transfer, we let the 12k images with the ``genre'' tag of ``landscape'' in the WikiArt~\cite{wikiart} be the artwork dataset, and the LandscapesHQ~\cite{lhq} be the photo dataset, to ensure a semantic alignment between artistic and realistic domains~\cite{tomei2019art2real}. 
All input images are resized to 256$\times$256 pixels.

\bfsection{Network architecture and training} 
The encoder/decoder of QuantArt consists of four blocks, where each block contains two ResBlocks~\cite{resnet} and a downsampling/upsampling layer. The intermediate feature has a spatial size of $16 \times 16$ and a feature dimension of 256. The codebook has $N=1024$ entries and an entry dimension of $d=256$. The style transfer model consists of six SGA modules. 
We implement QuantArt on the PyTorch framework~\cite{pytorch}. For both two training stages, we use an Adam optimizer~\cite{kingma2014adam} with a batch size of 32, a learning rate of $4.5\times10^{-6}$, and a training epoch of 50. The loss weights of $\mathcal{L}_{AE}$, $\mathcal{L}_{codebook}$, $\mathcal{L}_{content}$, $\mathcal{L}_{style}$, and $\mathcal{L}_{adv}$ are set to 1, 1, 1, 10, and 0.8, respectively.

\subsection{Multi-Fidelity Image Style Transfer}
\begin{figure}[t]
    \centering
    \includegraphics[width=1\linewidth]{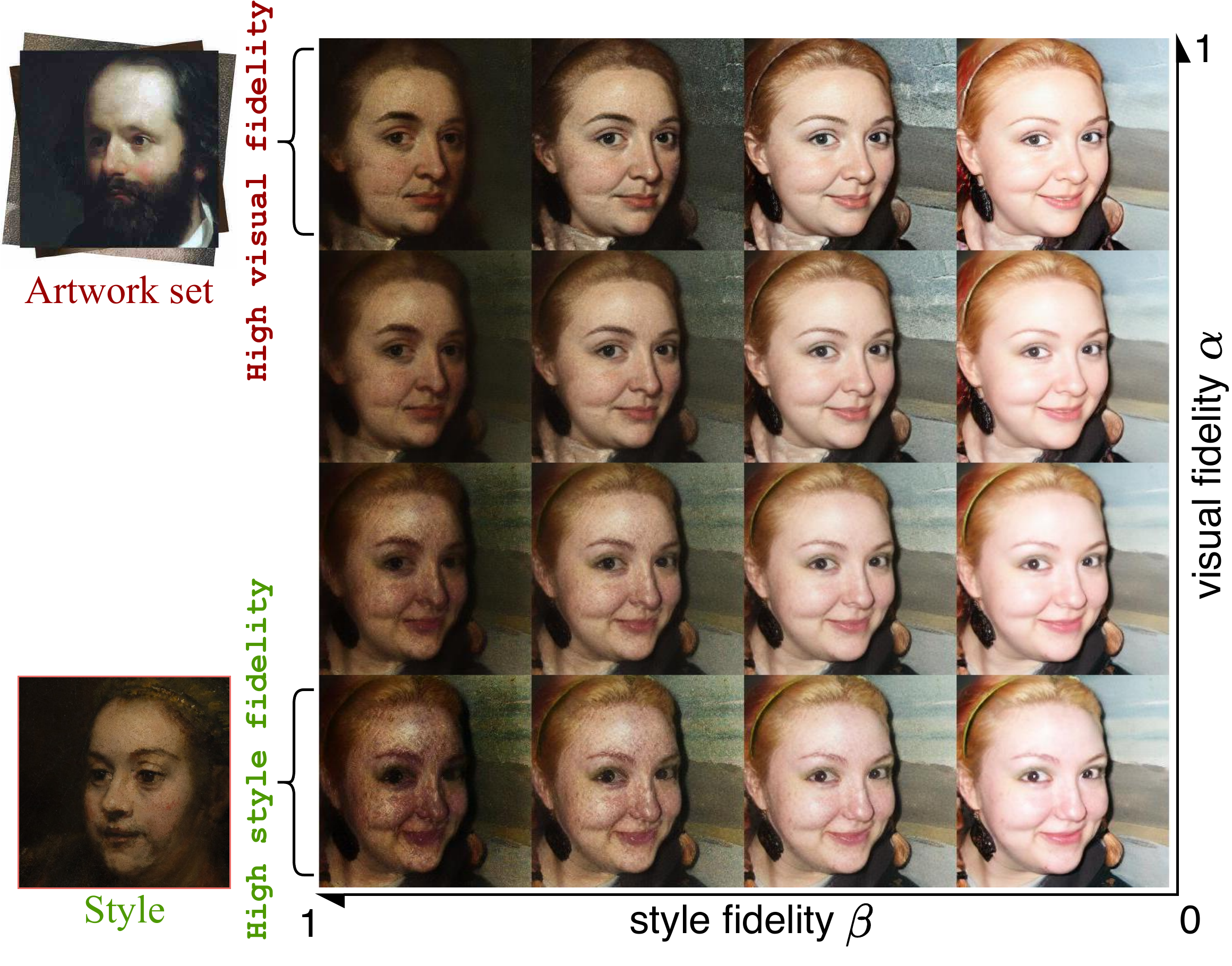}
    \vspace{-2em}
    \caption{QuantArt achieves smooth interpolation among content, style, and visual fidelities by adjusting the parameters $\alpha$ and $\beta$. 
    }
    \vspace{-1em}
    \label{fig:double_interpolation}
\end{figure}

Our QuantArt($\alpha,\beta$) method produces diverse stylization results to meet the preferences of content, style, and visual fidelities for different users.
Taking the Photo-to-Artwork style transfer task as an
example, Fig.~\ref{fig:double_interpolation} shows a trilinear interpolation of the three fidelity terms by uniformly sampling parameter combinations $(\alpha,\beta)$.
When the style fidelity $\beta=0$, the model simply reconstructs the input content image. 
With $\beta=1$ and visual fidelity $\alpha \rightarrow 1$, the generated images look more vivid but lose some texture details in the style reference. Note that neither $\alpha=0$ or $\alpha=1$ performs the best for this example, revealing the necessity of flexibly adjusting the fidelity terms in practical application.

\begin{figure*}[t]
    \centering
    \scriptsize
    \newcommand\w{0.125}
    
    \begin{minipage}{0.015\linewidth}
    \rotatebox{90}{~~~~~~~~~~~\small \bf Photo $\rightarrow$ Artwork}
    \end{minipage}
    \hfill
    \begin{minipage}{0.975\linewidth}
    \includegraphics[width=1\linewidth,height=7.5em]{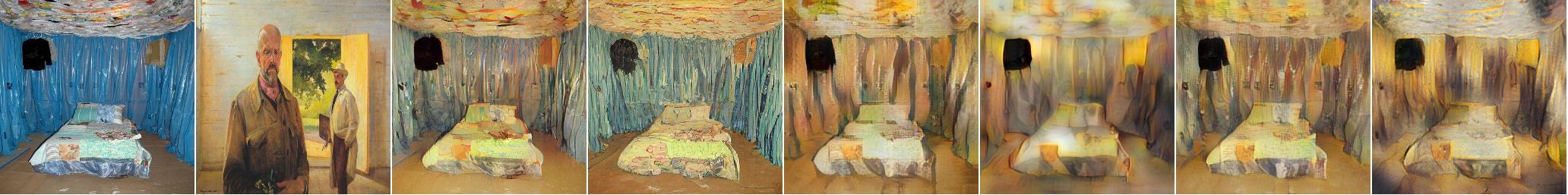}
    \setlength{\tabcolsep}{0pt}
    \vspace{-1.4em}\\
    \begin{tabular}
    {x{\w\linewidth}x{\w\linewidth}x{\w\linewidth}x{\w\linewidth}x{\w\linewidth}x{\w\linewidth}x{\w\linewidth}x{\w\linewidth}}
     Input &  Reference &  Ours ($\alpha=0$) & Ours ($\alpha=1$) &  AdaIN~\cite{adain} &  WCT~\cite{wct} &  LinearWCT~\cite{linearWCT} & DSTN~\cite{dstn} 
    \end{tabular}
    \includegraphics[width=1\linewidth,height=7.5em]{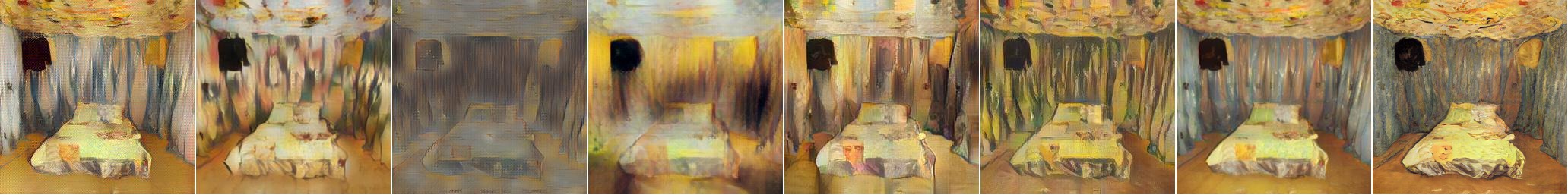}
    \vspace{-1.4em}\\
    \begin{tabular}
    {x{\w\linewidth}x{\w\linewidth}x{\w\linewidth}x{\w\linewidth}x{\w\linewidth}x{\w\linewidth}x{\w\linewidth}x{\w\linewidth}}
     ArtFlow~\cite{artflow} &  EFDM~\cite{efdm} &  StyleSwap~\cite{styleswap} &  Avatar-Net~\cite{avatar} &  SANet~\cite{sanet} &  AdaAttN~\cite{adaattn} &  StyTR2~\cite{stytr2} & CAST~\cite{zhang2022domain}
    \end{tabular}
    \end{minipage}
    \\\hrulefill 
    \vspace{0.3em}
    
    \begin{minipage}{0.015\linewidth}
    \rotatebox{90}{~~~~\small \bf Artwork $\rightarrow$ Artwork}
    \end{minipage}
    \hfill
    \begin{minipage}{0.975\linewidth}
    \includegraphics[width=1\linewidth,height=7.5em]{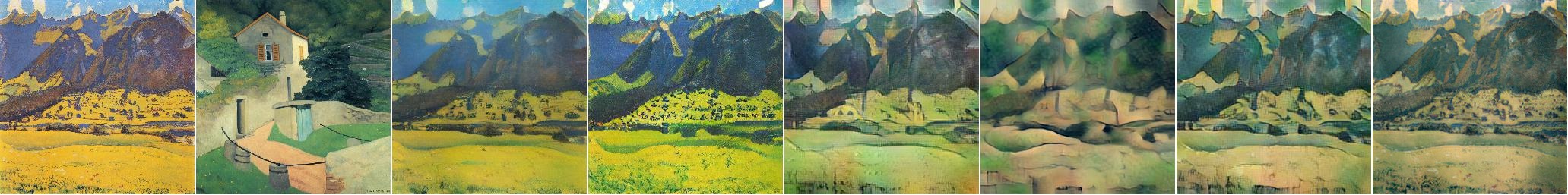}
    \setlength{\tabcolsep}{0pt}
    \vspace{-1.4em}\\
    \begin{tabular}
    {x{\w\linewidth}x{\w\linewidth}x{\w\linewidth}x{\w\linewidth}x{\w\linewidth}x{\w\linewidth}x{\w\linewidth}x{\w\linewidth}}
     Input &  Reference &  Ours ($\alpha=0$) & Ours ($\alpha=1$) &  AdaIN~\cite{adain} &  WCT~\cite{wct} &  LinearWCT~\cite{linearWCT} & DSTN~\cite{dstn} 
    \end{tabular}
    \includegraphics[width=1\linewidth,height=7.5em]{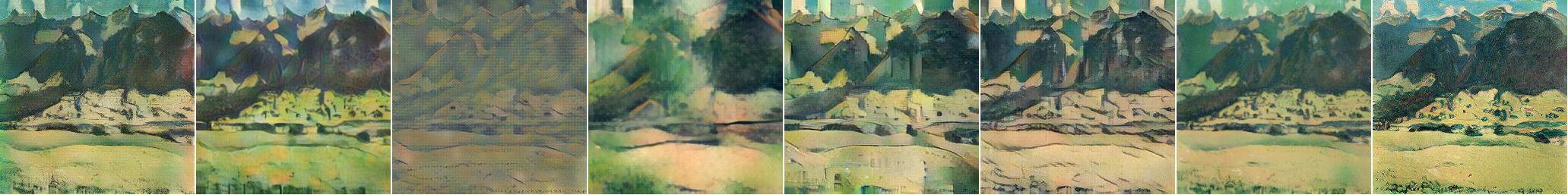}
    \vspace{-1.4em}\\
    \begin{tabular}
    {x{\w\linewidth}x{\w\linewidth}x{\w\linewidth}x{\w\linewidth}x{\w\linewidth}x{\w\linewidth}x{\w\linewidth}x{\w\linewidth}}
     ArtFlow~\cite{artflow} &  EFDM~\cite{efdm} &  StyleSwap~\cite{styleswap} &  Avatar-Net~\cite{avatar} &  SANet~\cite{sanet} &  AdaAttN~\cite{adaattn} &  StyTR2~\cite{stytr2} & CAST~\cite{zhang2022domain}
    \end{tabular}
    \end{minipage}
    \\\hrulefill 
    \vspace{0.3em}

    \begin{minipage}{0.015\linewidth}
    \rotatebox{90}{~~~~~\small \bf Photo $\rightarrow$ Photo}
    \end{minipage}
    \hfill
    \begin{minipage}{0.975\linewidth}
    \includegraphics[width=1\linewidth]{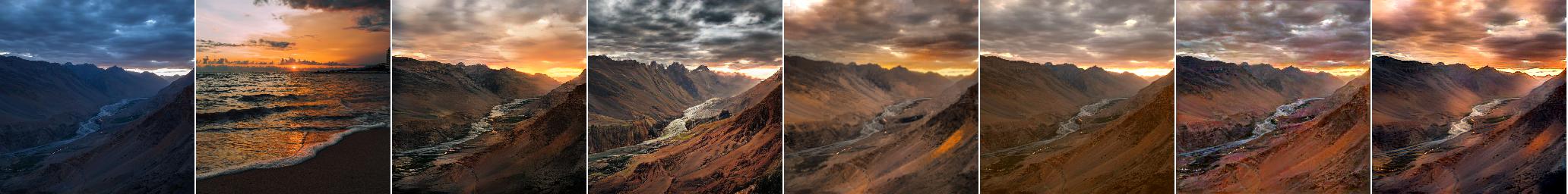}
    \setlength{\tabcolsep}{0pt}
    \vspace{-1.4em}\\
    \renewcommand\w{0.125}
    \begin{tabular}
    {x{\w\linewidth}x{\w\linewidth}x{\w\linewidth}x{\w\linewidth}x{\w\linewidth}x{\w\linewidth}x{\w\linewidth}x{\w\linewidth}}
     Input &  Reference &  Ours ($\alpha=0$) & Ours ($\alpha=1$) &    PhotoWCT~\cite{photowct} &  WCT$^2$~\cite{wct2} & LinearWCT~\cite{linearWCT} & DSTN~\cite{dstn}  
    \end{tabular}
    \end{minipage}

    \vspace{-.8em}
    \caption{Comparisons of the state-of-the-art methods for artistic image style transfer, \ie, Photo-to-Artwork and Artwork-to-Artwork, and photorealistic image style transfer, \ie, Photo-to-Photo.  
    \emph{Zoom in to view the details.}
    }
    \vspace{-.8em}
    \label{fig:qualitative}
\end{figure*}

\begin{table*}[t]
\centering
\caption{Quantitative comparison of the universal style transfer methods. The \textbf{best} and \underline{second best} results are highlighted, respectively.}%
\vspace{-1em}
     \resizebox{1\hsize}{!}{
    \begin{tabular}{lccccccccccccccc}
        \hline
       \multirow{2}{*}{Metric $\downarrow$} & \multirow{2}{*}{AdaIN} &  \multirow{2}{*}{WCT} &  \multirow{2}{*}{LinearWCT} & \multirow{2}{*}{ArtFlow} &  \multirow{2}{*}{EFDM} & \multirow{2}{*}{StyleSwap} &  \multirow{2}{*}{AvatarNet} &  \multirow{2}{*}{SANet} &  \multirow{2}{*}{AdaAttN} &  \multirow{2}{*}{StyTR2} & \multicolumn{3}{c}{Ours ($\alpha,\beta$)}\\ \cline{12-14}
        &&&&&&&&&&&$(0,0)$&$(0,1)$&$(1,1)$\\\hline
        LPIPS $\downarrow$ & 0.681 & 0.695 & 0.657 & 0.603 & 0.652 & 0.607 & 0.706 & 0.686 & 0.633 & \textbf{0.514} & \textit{0.159} & \underline{0.565} & 0.581 \\ 
        Gram loss\begin{scriptsize} (${\times 10^{3}}$) \end{scriptsize}  $\downarrow$  & 0.163 & 0.282 &	0.172 &	\textbf{0.101} & 0.402 & 1.357 & 0.718 & \underline{0.120} & 0.246 & 0.386 & - & 0.395 & 0.864 \\
        FID $\downarrow$ & 36.618 & 65.193 & 48.156 & 28.899 & 51.070 & 74.168 & 58.178 & 27.080 & 25.894 & 30.893 & - & \underline{25.590} & \textbf{17.787} \\ 
        ArtFID $\downarrow$ & 63.240 & 112.229 & 81.452 & 47.936 & 86.007 & 120.803 & 100.964 & 47.356 & 43.910 & 48.284 & - & \underline{41.623} & \textbf{29.695} \\ \hline 
    \end{tabular}
    }
    \vspace{-3mm}
    
    \label{tab:quantative}
\end{table*}

\begin{table}[h]
\centering
\caption{Inference time (seconds) of a 256$\times$256 image with a NVIDIA Tesla V100 GPU. }
\label{tab:time}
\vspace{-1em}
\resizebox{1\linewidth}{!}{%
\begin{tabular}{cccccccc}
\hline
ArtFlow & StyleSwap & AvatarNet & SANet & AdaAttN & StyTR2 & Ours \\ \hline
 0.068	& 0.043 & 0.124	& \textbf{0.007}	& 0.050	& 0.055	& 0.045 \\ \hline
\end{tabular}%
}
\vspace{-1.5em}
\end{table}

\subsection{Comparing with State-of-the-Art Methods}
For a comprehensive evaluation, we qualitatively and quantitatively compare our QuantArt framework with the state-of-the-art algorithms on tasks including artistic style transfer (\ie, Photo-to-Artwork, Artwork-to-Artwork) and photo-realistic style transfer (\ie, Photo-to-Photo).

\bfsection{Methods for comparison} 
For artistic style transfer, we compare three lines of methods: 1) The feature statistics-based methods, including AdaIN~\cite{adain}, WCT~\cite{wct}, LinearWCT~\cite{linearWCT}, DSTN~\cite{dstn}, ArtFlow~\cite{artflow}, EFDM~\cite{efdm}, and CAST~\cite{zhang2022domain}; 2) The patch swapping methods, including StyleSwap~\cite{styleswap} and AvatarNet~\cite{avatar}; 3) The attention-based methods, including SANet~\cite{sanet}, AdaAttn~\cite{adaattn}, and StyTR2~\cite{stytr2}. For photorealistic style transfer, we compare PhotoWCT~\cite{photowct}, WCT$^2$~\cite{wct2}, LinearWCT~\cite{linearWCT}, and DSTN~\cite{dstn}. 

\bfsection{Qualitative results} 
As shown in Fig.~\ref{fig:qualitative}, on all three tasks, QuantArt with $\alpha=0$ faithfully transfers the texture from input style images, while QuantArt with $\alpha=1$ paints realistic textures thanks to the learned style codebook.
For the Photo-to-Photo example, the content image does not have clearly visible textures, and it is challenging for other methods to hallucinate texture details when the scene in the style reference is lit up. 

\bfsection{Quantitative results} 
We further quantitatively compare the performances of the state-of-the-art methods on the Photo-to-Artwork task. We randomly sample 10K images from the MS-COCO~\cite{mscoco} and WikiArt~\cite{wikiart} datasets, respectively, forming a total of 10K pairs of evaluation images. We use the LPIPS loss~\cite{lpips} to measure the content fidelity between the content and generated images. We use the style loss with a pretrained VGG-19 network~\cite{adain,stytr2} to measure the style fidelity between the style and generated images. We compute the FID metric~\cite{fid} between the style dataset and all the generated images to measure the visual fidelity of algorithms. We also adopt a recently proposed metric $\text{ArtFID} = (1+\text{LPIPS})\cdot(1+\text{FID})$~\cite{artfid} to evaluate the overall stylization performance. 
For a fair comparison, we compare our methods QuantArt(0,1) and QuantArt(1,1) with the previous methods. QuantArt(0,0) serves as a baseline to show the image reconstruction performance. As shown in Table~\ref{tab:quantative}, both our methods QuantArt(0,1) and QuantArt(1,1) achieve competitive performances compared with the baseline methods on all metrics. Specifically, QuantArt(1,1) performs better than other methods for FID and ArtFID, indicating that it can significantly enhance the visual fidelity of style transfer results.
QuantArt(0,1) performs better than QuantArt(1,1) on the Gram loss, empirically demonstrating the trade-off between style and visual fidelities. 
In Table \ref{tab:time}, the inference time of our method is comparable with most of the benchmark universal style transfer methods.

\subsection{Human Evaluation}
Since image style transfer is a highly subjective task, we further examine our model with human evaluation. 

\begin{figure}[t]
    \centering
    \includegraphics[width=1\linewidth]{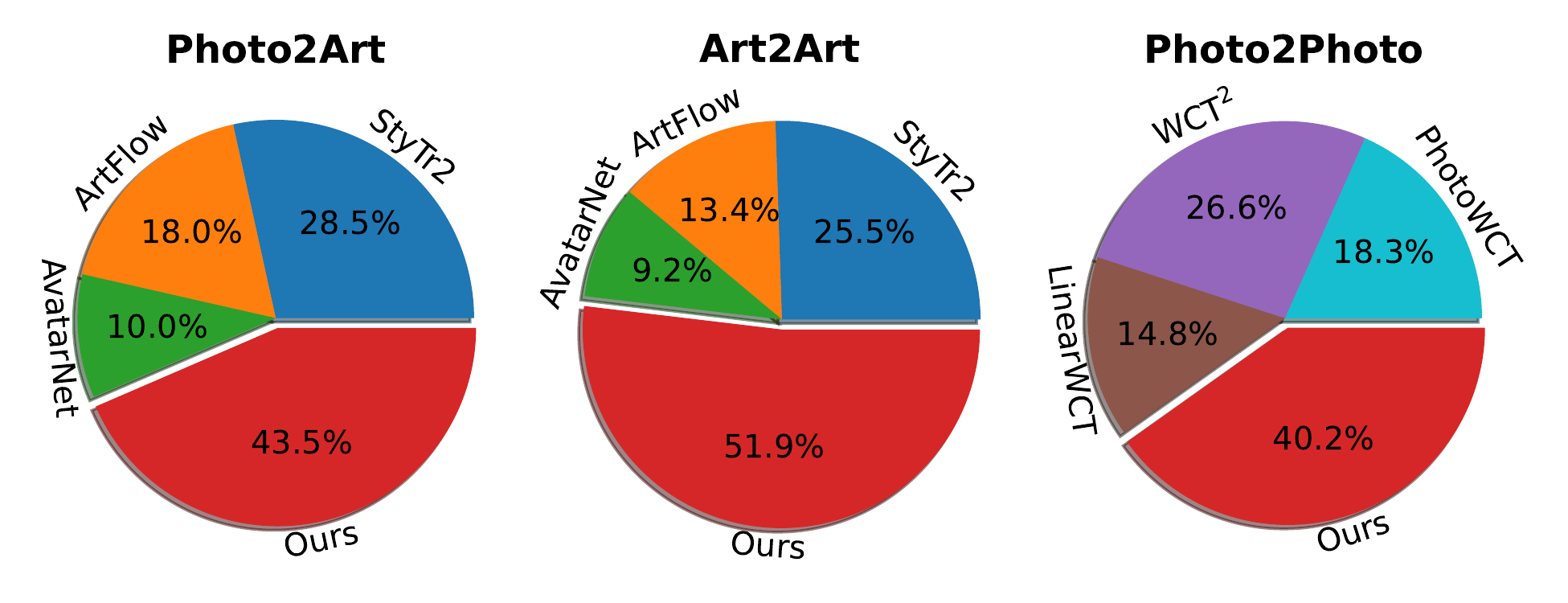}
    \vspace{-2.2em}
    \caption{User study results on three image style transfer tasks. We show the percentages of methods preferred by the participants.
    }
    \vspace{-.5em}
    \label{fig:userstudy}
\end{figure}

\bfsection{User study on method performance}
We perform a user study to subjectively evaluate the performance of different methods on three image style transfer tasks. On Photo-to-Artwork and Artwork-to-Artwork, we compare AvatarNet~\cite{avatar}, ArtFlow~\cite{artflow}, StyTr2~\cite{stytr2}, and our QuantArt(1,1). On Photo-to-Photo, we compare PhotoWCT~\cite{photowct}, WCT$^2$~\cite{wct2}, LinearWCT~\cite{linearWCT}, and QuantArt(1,1). In each example, we show the input image, style reference, and style transfer results of the four comparison methods. The user study participants are asked to select one of the four stylizations to their preference. We show 12 examples to every participant, where each style transfer task has 4 examples. We have collected effective responses from a total of 59 participants. 
As shown in Fig.~\ref{fig:userstudy}, our method significantly outperforms the prior arts on all three tasks with an average preference ratio of 43.5\%, 51.9\%, and 40.2\%, respectively. The results also indicate that the visual fidelity is an important evaluation metrics for human to assess the style transfer algorithms. 

\begin{figure}[t]
    \centering
    \vspace{-1.2em}
    \begin{minipage}{0.46\linewidth}
        \footnotesize \textbf{Q}: Which one is a real artwork? \\
        \includegraphics[width=0.48\linewidth]{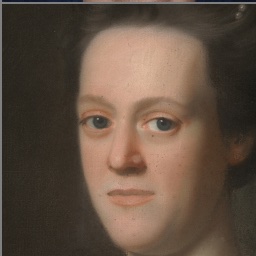}
        \includegraphics[width=0.48\linewidth]{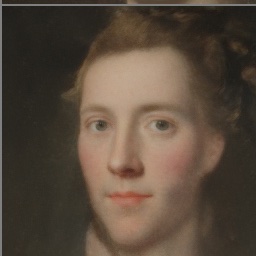}
    \end{minipage}
    \hfill
    \begin{minipage}{0.52\linewidth}
        \includegraphics[width=1\linewidth]{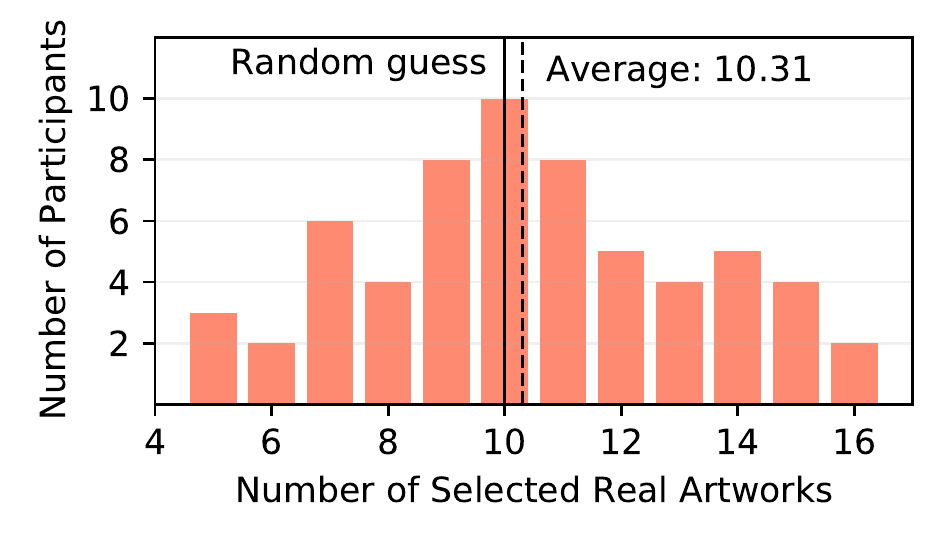}
        \vspace{-2em}
    \end{minipage}
    \\
    \caption{\textbf{(Left)} An example of our Artistic Style Transfer Confusion Test. Only 40.6\% participants successfully distinguished the real artwork in this example. The answer can be found in our supplementary material. \textbf{(Right)} The statistical results with a total of 61 participants, where each participant is asked 20 questions. 
    }
    \vspace{-2em}
    \label{fig:turing}
\end{figure}

\bfsection{Confusion test on visual fidelity}
We further perform a novel user study, dubbed artistic style transfer confusion test, to subjectively evaluate the visual fidelity of the stylized images. The left part of Fig.~\ref{fig:turing} shows an example question of the test. Within each question of the test, we ask the participant to select the real artwork from a pair of \emph{i)} a real artwork image $s$ and \emph{ii)} a photo stylized according to the reference of $s$. The test is more challenging compared with the previous tests designed for image style transfer~\cite{zhang2022domain}, since the fake image should follow the artistic style of the paired real image. 
We generate a total of 50 images using QuantArt(1,1). The content images are randomly selected from the FFHQ~\cite{stylegan}, LandscapeHQ~\cite{lhq}, or MS-COCO~\cite{mscoco} dataset. The style images are randomly selected from the WikiArt~\cite{wikiart} or MetFaces~\cite{metfaces} dataset. To avoid image contents that would lead to unfair comparison, such as the hairstyles of middle ages in the reference image or the modern articles in the content image, we manually filter 36 examples from the 50 examples. Each participant is requested to answer 20 questions, where they can choose either one of the two presented images, or skip the question if she/he feels difficult to make a choice. We have collected effective responses from 61 participants. 

We plot the histogram of participants over the number of correct selections in the right part of Fig.~\ref{fig:turing}. The distribution of the participants roughly follow the normal distribution with the mean number of correct selections 10.31, close to the random guess result (10). It indicates that our method can generate highly realistic artistic images which are difficult for human to identify from the real ones. More examples of this artistic confusion test can be found in the supplementary material.

\begin{table}[t]
\centering
\caption{An ablation study of the design choices of QuantArt.
}
\vspace{-1em}
     \resizebox{1\hsize}{!}{
      \begin{tabular}{lccc}
        \hline
        ~ & LPIPS $\downarrow$ & Gram $\downarrow$ & FID $\downarrow$ \\ 
        Complete model, $\alpha=1$, code size 16$\times$16 & 0.581 & 0.864 & \textbf{17.787} \\ \hline
        ~~~~Remove Quantization in SGA & 0.545 & 0.993 & 18.757 \\ 
        ~~~~Remove all Quantizations (\ie, $\alpha=0$) & 0.565 & 0.395 & 25.590 \\ \hline
        ~~~~Code size 8$\times$8 & 0.656 & 1.078 & 27.741 \\ 
        ~~~~Code size 32$\times$32 & 0.513 & 0.798 & 28.663 \\ \hline
        ~~~~Remove self-attention in SGA & 0.584 & 0.883 & 19.025 \\ 
        ~~~~Remove ResBlock in SGA & 0.608 & 1.841 & 27.177 \\ 
        ~~~~Replace SGA with StyTR2 decoder layer \cite{stytr2} & 0.539 &	0.875 & 26.299 \\ \hline
        ~~~~Shared encoders & 0.588 & 0.788 & 30.806\\
        ~~~~Shared encoders and decoders & 0.593 & 0.891 & 35.830 \\ \hline
    \end{tabular}
    }
    \vspace{-1em}
    \label{tab:ablation}
\end{table}

\subsection{Ablation Study}
As shown in Table~\ref{tab:ablation}, we conduct the following ablation studies to justify the design choices of the proposed framework. 
1) \emph{The effectiveness of vector quantization.} By removing the quantization operators in SGA or removing all the quantization operators, the Gram loss decreases while the FID increases. It is aligned with our motivation that vector quantization trades-off the style fidelity with visual fidelity.
2) \emph{Code patch size.} Each code in QuantArt corresponds to a 16$\times$16 patch of the feature maps. Either increasing or decreasing the results in an increase of FID, indicating that the code size of 16$\times$16 is a good choice for image style transfer.
3) \emph{Components in SGA.} Removing either the self-attention or ResBlock in SGA results in a worse FID. Removing the ResBlock additionally results in a significant increase of Gram loss. Replacing SGA with the decoder layer of StyTR2 \cite{stytr2} leads to a worse FID.
4) \emph{Shared auto-encoders.} QuantArt adopts separate auto-enocoders to extract the continuous and quantized feature representations, respectively (see Fig.~\ref{fig:framework}). The FID drastically increases when we share the encoders or auto-encoders for continuous and quantized representations. Therefore, in this work we use separate auto-encoders for feature extraction.

\vspace{-.5em}
\section{Conclusion}
\vspace{-.3em}
In this paper, we study the problem of enhancing the visual fidelity of image style transfer. We have proposed a quantizing style transfer algorithm to make the latent feature of the generated artwork closer to the real distribution. However, the algorithm may result in a generation that displays fewer style details of the reference image. To address this limitation, we have further presented a style transfer framework called QuantArt, which consists of a continuous branch and a quantized branch, to allow users to arbitrarily control the content preservation, style similarity, and visual fidelity of generated artworks. The experiments on Photo-to-Artwork, Artwork-to-Artwork, and Photo-to-Photo transfer settings have shown that our method achieves higher visual fidelity along with comparable content and style fidelities compared with the state-of-the-art methods.

\vspace{.65em}
\bfsection{Acknoledgements}
This work was supported by NIH grant 5U54CA225088-03. The computations in this paper were run on the FASRC Cannon cluster supported by Harvard.

{\small
\balance
\bibliographystyle{ieee_fullname}
\bibliography{refs}
}

\newpage
\appendix
\onecolumn

\section{More Quantitative Comparisons}
\label{sec:quant}

We quantitatively compare our method with the state-of-the-art methods for three different settings: artwork-to-artwork (Table~\ref{tab:art2art}), photo-to-photo (Table~\ref{tab:real2real}), and artwork-to-photo (Table~\ref{tab:art2real}). Tables~\ref{tab:art2art} and~\ref{tab:real2real} show that our QuantArt(1,1) framework achieves the best FID and ArtFID scores on both settings, which is consistent with the quantitative result of the photo-to-artwork task shown in the paper. The consistent results in different settings reinforce our finding that feature quantization can lead to a higher visual fidelity in image style transfer. QuantArt(0,1) has a lower Gram loss than QuantArt(1,1) because the visual and style fidelities are orthogonal transfer directions in some cases. To this end, we introduce a stronger control ability to the proposed framework, where the user can easily control the trade-off between visual, style, and content fidelities by adjusting the parameters $\alpha$ and $\beta$.

For the artwork-to-photo task, following Art2Real~\cite{tomei2019art2real}, we do not adopt the style reference, such that the task is formulated as an unpaired image translation problem. We train an SGA module that takes the feature of an artwork image as the content input. The second attention block in SGA is replaced with a self-attention block. We compare our method with CycleGAN~\cite{cyclegan}, CUT~\cite{cut}, DRIT~\cite{drit} and Art2Real~\cite{tomei2019art2real}, where CycleGAN~\cite{cyclegan}, CUT~\cite{cut} and DRIT~\cite{drit} are benchmark methods for unpaired image translation, and Art2Real~\cite{tomei2019art2real} is the algorithm specially designed for artwork-to-photo translation.
Table \ref{tab:art2real} shows that our QuantArt framework can achieve decent performance without special modification for the artwork-to-photo task, since it directly fetches photorealism patch representations from the learned photo codebook.

\section{More Qualitative Results}
\label{sec:qual}

\begin{figure*}[t]
    \centering
    \scriptsize
    \newcommand\w{0.1426}
    
    \begin{minipage}{1\linewidth}
    \includegraphics[width=1\linewidth]{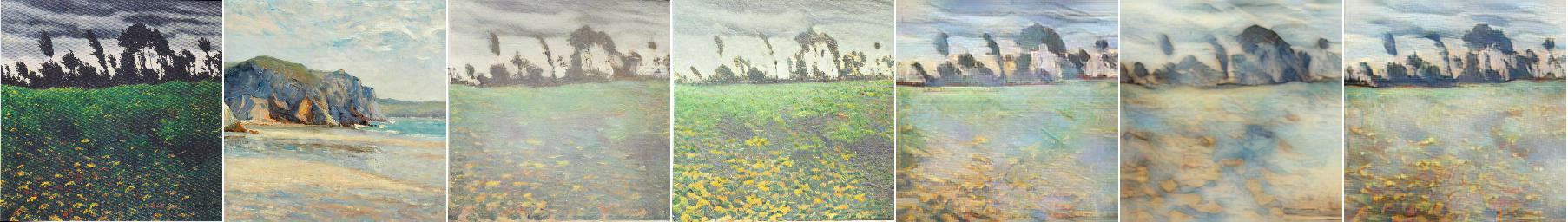}
    \setlength{\tabcolsep}{0pt}
    \vspace{-1.4em}\\
    \begin{tabular}
    {x{\w\linewidth}x{\w\linewidth}x{\w\linewidth}x{\w\linewidth}x{\w\linewidth}x{\w\linewidth}x{\w\linewidth}}
     Input &  Reference &  Ours ($\alpha=0$) & Ours ($\alpha=1$) &  AdaIN~\cite{adain} &  WCT~\cite{wct} &  LinearWCT~\cite{linearWCT} 
    \end{tabular}
    \includegraphics[width=1\linewidth]{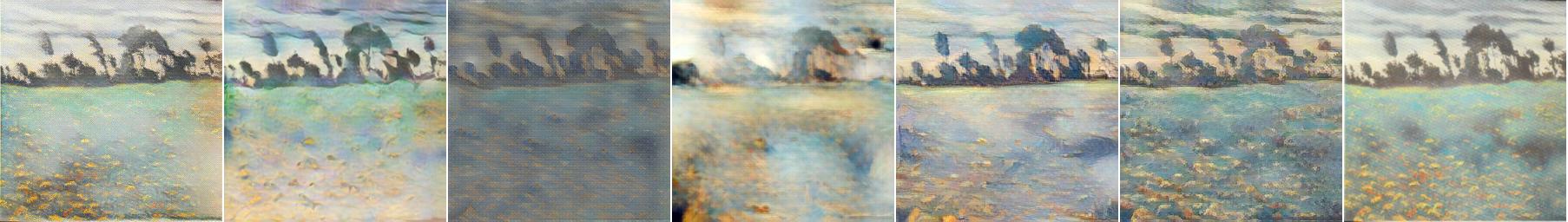}
    \vspace{-1.4em}\\
    \begin{tabular}
    {x{\w\linewidth}x{\w\linewidth}x{\w\linewidth}x{\w\linewidth}x{\w\linewidth}x{\w\linewidth}x{\w\linewidth}}
     ArtFlow~\cite{artflow} &  EFDM~\cite{efdm} &  StyleSwap~\cite{styleswap} &  Avatar-Net~\cite{avatar} &  SANet~\cite{sanet} &  AdaAttN~\cite{adaattn} &  StyTR2~\cite{stytr2}
    \end{tabular}
    \end{minipage}
    \\\hrulefill 
    \vspace{0.3em}

    \begin{minipage}{1\linewidth}
    \includegraphics[width=1\linewidth]{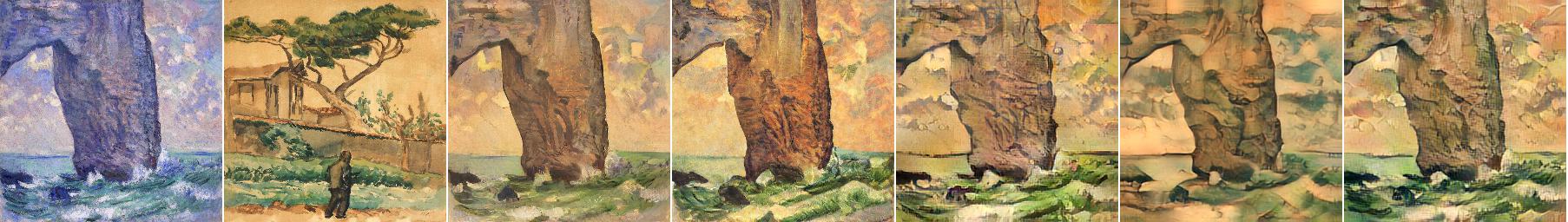}
    \setlength{\tabcolsep}{0pt}
    \vspace{-1.4em}\\
    \begin{tabular}
    {x{\w\linewidth}x{\w\linewidth}x{\w\linewidth}x{\w\linewidth}x{\w\linewidth}x{\w\linewidth}x{\w\linewidth}}
     Input &  Reference &  Ours ($\alpha=0$) & Ours ($\alpha=1$) &  AdaIN~\cite{adain} &  WCT~\cite{wct} &  LinearWCT~\cite{linearWCT} 
    \end{tabular}
    \includegraphics[width=1\linewidth]{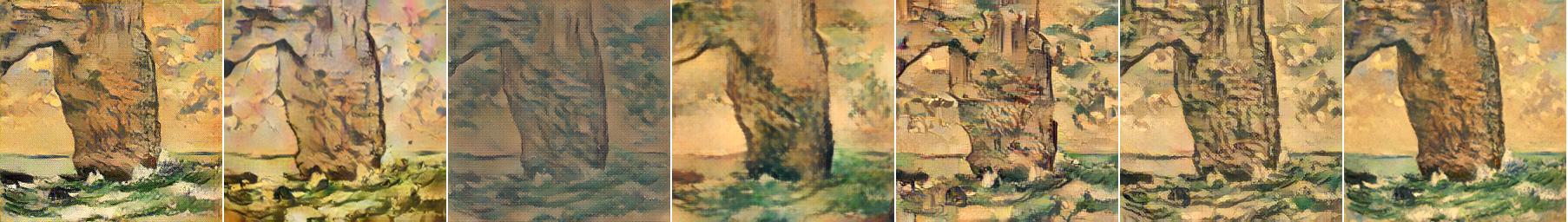}
    \vspace{-1.4em}\\
    \begin{tabular}
    {x{\w\linewidth}x{\w\linewidth}x{\w\linewidth}x{\w\linewidth}x{\w\linewidth}x{\w\linewidth}x{\w\linewidth}}
     ArtFlow~\cite{artflow} &  EFDM~\cite{efdm} &  StyleSwap~\cite{styleswap} &  Avatar-Net~\cite{avatar} &  SANet~\cite{sanet} &  AdaAttN~\cite{adaattn} &  StyTR2~\cite{stytr2}
    \end{tabular}
    \end{minipage}
    \\\hrulefill 
    \vspace{0.3em}

    \begin{minipage}{1\linewidth}
    \includegraphics[width=1\linewidth]{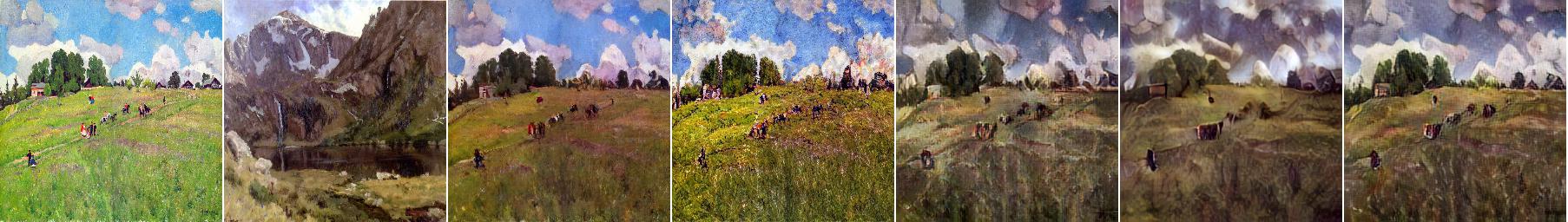}
    \setlength{\tabcolsep}{0pt}
    \vspace{-1.4em}\\
    \begin{tabular}
    {x{\w\linewidth}x{\w\linewidth}x{\w\linewidth}x{\w\linewidth}x{\w\linewidth}x{\w\linewidth}x{\w\linewidth}}
     Input &  Reference &  Ours ($\alpha=0$) & Ours ($\alpha=1$) &  AdaIN~\cite{adain} &  WCT~\cite{wct} &  LinearWCT~\cite{linearWCT} 
    \end{tabular}
    \includegraphics[width=1\linewidth]{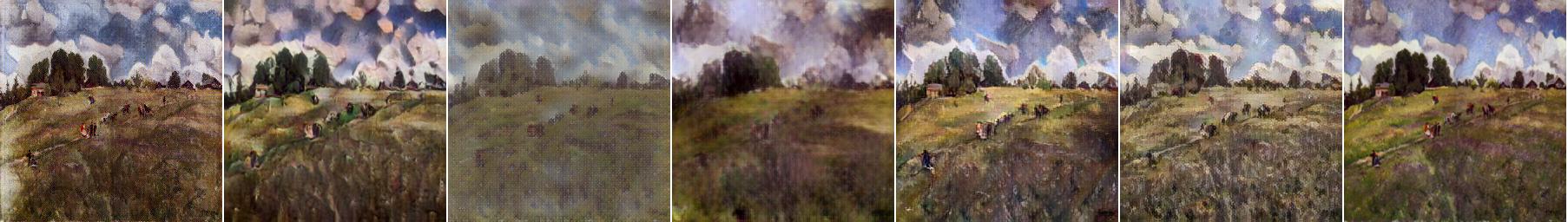}
    \vspace{-1.4em}\\
    \begin{tabular}
    {x{\w\linewidth}x{\w\linewidth}x{\w\linewidth}x{\w\linewidth}x{\w\linewidth}x{\w\linewidth}x{\w\linewidth}}
     ArtFlow~\cite{artflow} &  EFDM~\cite{efdm} &  StyleSwap~\cite{styleswap} &  Avatar-Net~\cite{avatar} &  SANet~\cite{sanet} &  AdaAttN~\cite{adaattn} &  StyTR2~\cite{stytr2}
    \end{tabular}
    \end{minipage}
    
    \caption{More \textbf{artwork-to-artwork} style transfer results by the state-of-the-art methods. 
    }
    \label{fig:art2art1}
\end{figure*}

\begin{figure*}[t]
    \centering
    \scriptsize
    \newcommand\w{0.1426}
    
    \begin{minipage}{1\linewidth}
    \includegraphics[width=1\linewidth]{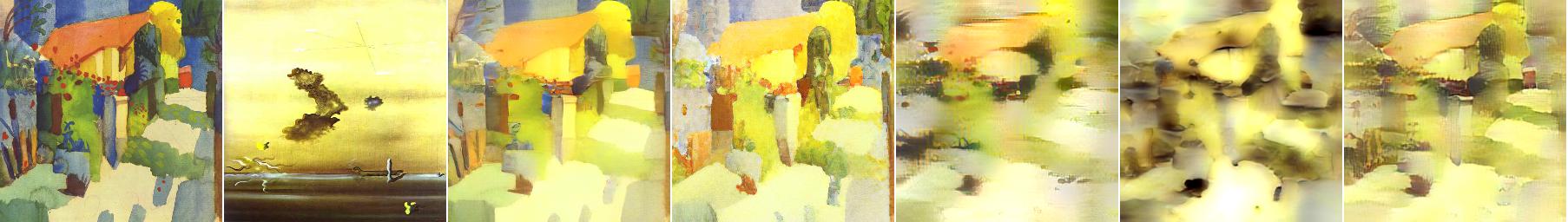}
    \setlength{\tabcolsep}{0pt}
    \vspace{-1.4em}\\
    \begin{tabular}
    {x{\w\linewidth}x{\w\linewidth}x{\w\linewidth}x{\w\linewidth}x{\w\linewidth}x{\w\linewidth}x{\w\linewidth}}
     Input &  Reference &  Ours ($\alpha=0$) & Ours ($\alpha=1$) &  AdaIN~\cite{adain} &  WCT~\cite{wct} &  LinearWCT~\cite{linearWCT} 
    \end{tabular}
    \includegraphics[width=1\linewidth]{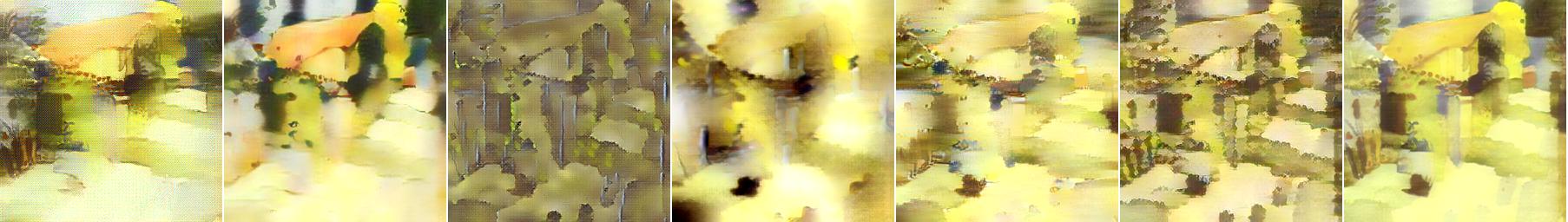}
    \vspace{-1.4em}\\
    \begin{tabular}
    {x{\w\linewidth}x{\w\linewidth}x{\w\linewidth}x{\w\linewidth}x{\w\linewidth}x{\w\linewidth}x{\w\linewidth}}
     ArtFlow~\cite{artflow} &  EFDM~\cite{efdm} &  StyleSwap~\cite{styleswap} &  Avatar-Net~\cite{avatar} &  SANet~\cite{sanet} &  AdaAttN~\cite{adaattn} &  StyTR2~\cite{stytr2}
    \end{tabular}
    \end{minipage}
    \\\hrulefill 
    \vspace{0.3em}

    \begin{minipage}{1\linewidth}
    \includegraphics[width=1\linewidth]{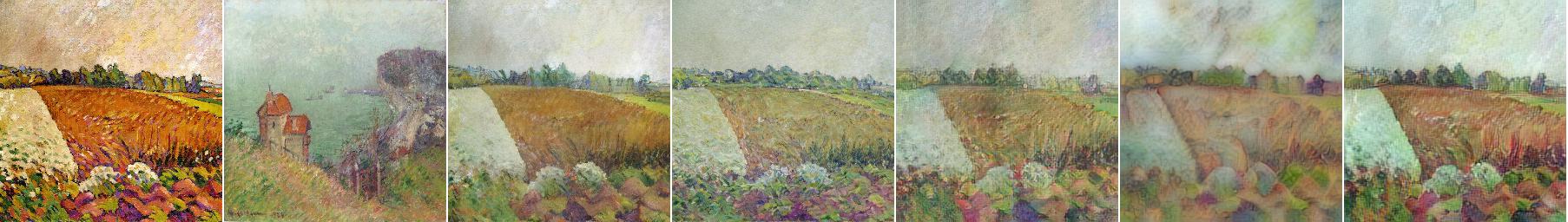}
    \setlength{\tabcolsep}{0pt}
    \vspace{-1.4em}\\
    \begin{tabular}
    {x{\w\linewidth}x{\w\linewidth}x{\w\linewidth}x{\w\linewidth}x{\w\linewidth}x{\w\linewidth}x{\w\linewidth}}
     Input &  Reference &  Ours ($\alpha=0$) & Ours ($\alpha=1$) &  AdaIN~\cite{adain} &  WCT~\cite{wct} &  LinearWCT~\cite{linearWCT} 
    \end{tabular}
    \includegraphics[width=1\linewidth]{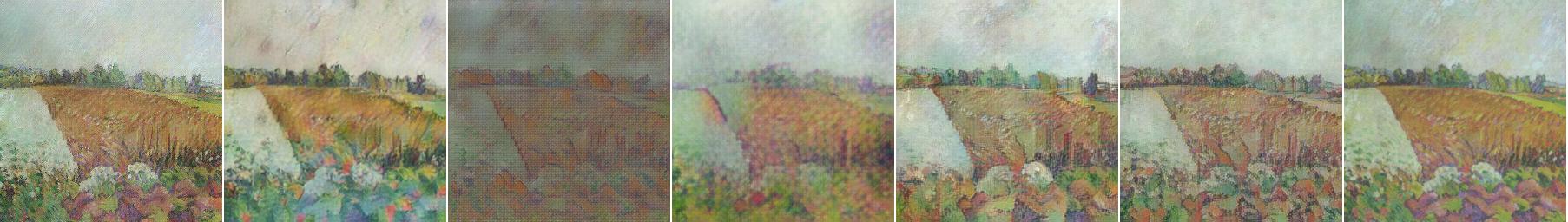}
    \vspace{-1.4em}\\
    \begin{tabular}
    {x{\w\linewidth}x{\w\linewidth}x{\w\linewidth}x{\w\linewidth}x{\w\linewidth}x{\w\linewidth}x{\w\linewidth}}
     ArtFlow~\cite{artflow} &  EFDM~\cite{efdm} &  StyleSwap~\cite{styleswap} &  Avatar-Net~\cite{avatar} &  SANet~\cite{sanet} &  AdaAttN~\cite{adaattn} &  StyTR2~\cite{stytr2}
    \end{tabular}
    \end{minipage}
    \\\hrulefill 
    \vspace{0.3em}

    \begin{minipage}{1\linewidth}
    \includegraphics[width=1\linewidth]{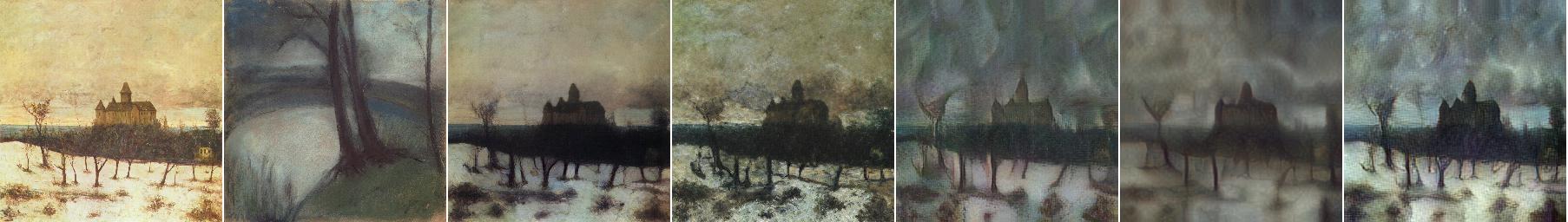}
    \setlength{\tabcolsep}{0pt}
    \vspace{-1.4em}\\
    \begin{tabular}
    {x{\w\linewidth}x{\w\linewidth}x{\w\linewidth}x{\w\linewidth}x{\w\linewidth}x{\w\linewidth}x{\w\linewidth}}
     Input &  Reference &  Ours ($\alpha=0$) & Ours ($\alpha=1$) &  AdaIN~\cite{adain} &  WCT~\cite{wct} &  LinearWCT~\cite{linearWCT} 
    \end{tabular}
    \includegraphics[width=1\linewidth]{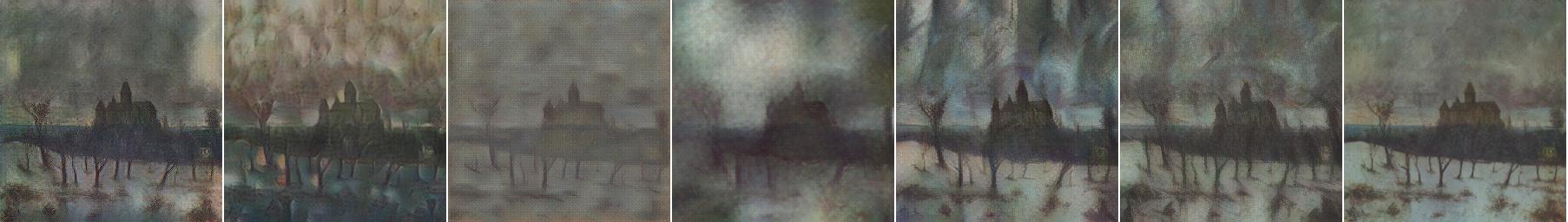}
    \vspace{-1.4em}\\
    \begin{tabular}
    {x{\w\linewidth}x{\w\linewidth}x{\w\linewidth}x{\w\linewidth}x{\w\linewidth}x{\w\linewidth}x{\w\linewidth}}
     ArtFlow~\cite{artflow} &  EFDM~\cite{efdm} &  StyleSwap~\cite{styleswap} &  Avatar-Net~\cite{avatar} &  SANet~\cite{sanet} &  AdaAttN~\cite{adaattn} &  StyTR2~\cite{stytr2}
    \end{tabular}
    \end{minipage}
    
    \caption{More \textbf{artwork-to-artwork} style transfer results by the state-of-the-art methods.  
    }
    \label{fig:art2art2}
\end{figure*}

\begin{figure*}[t]
    \centering
    \scriptsize
    \newcommand\w{0.1426}

    \includegraphics[width=1\linewidth]{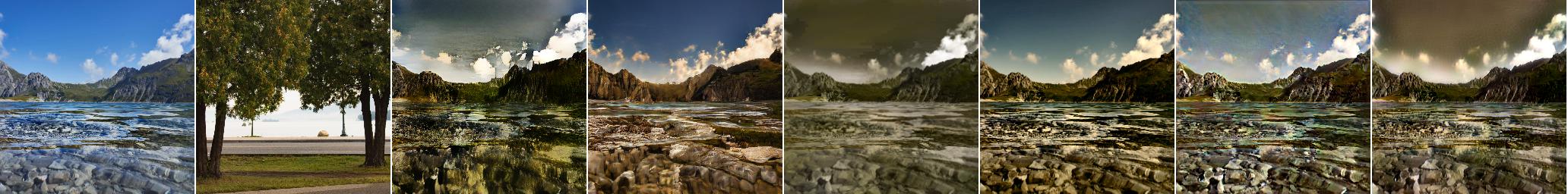}
    \\\vspace{0.3em}
    \includegraphics[width=1\linewidth]{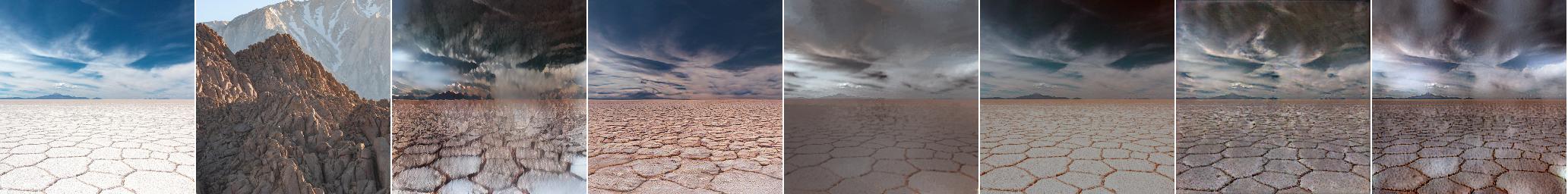}
    \\\vspace{0.3em}
    \includegraphics[width=1\linewidth]{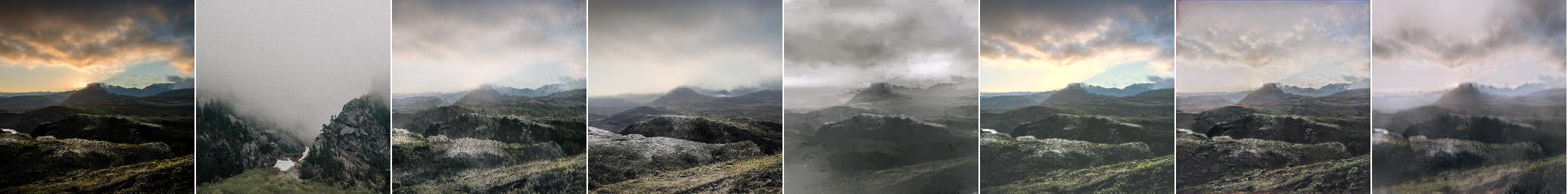}
    \\\vspace{0.3em}
    \includegraphics[width=1\linewidth]{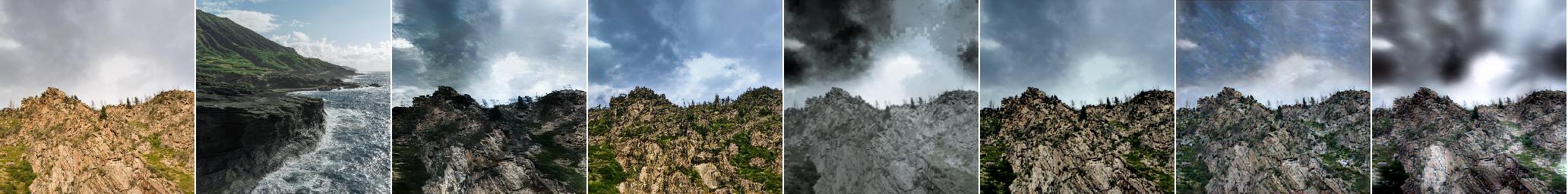}
    \\\vspace{0.3em}
    \includegraphics[width=1\linewidth]{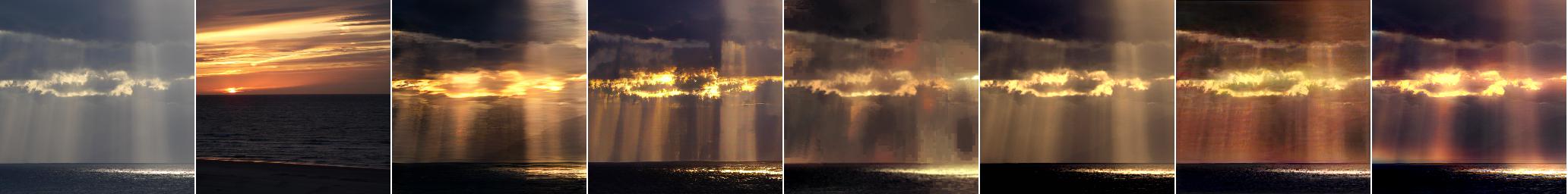}
    \\\vspace{0.3em}
    
    \begin{minipage}{1\linewidth}
    \includegraphics[width=1\linewidth]{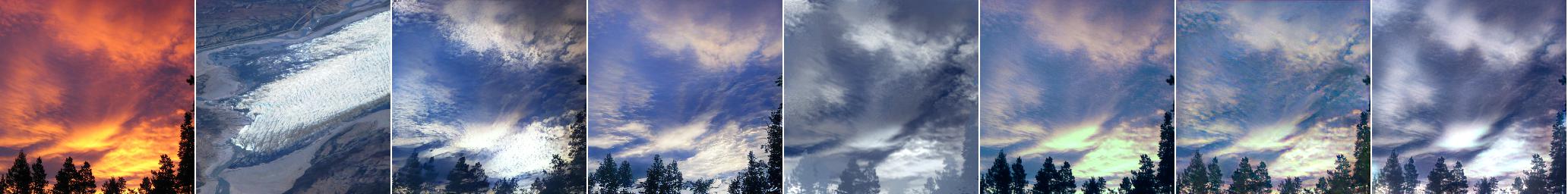}
    \setlength{\tabcolsep}{0pt}
    \vspace{-1.4em}\\
    \renewcommand\w{0.125}
    \begin{tabular}
    {x{\w\linewidth}x{\w\linewidth}x{\w\linewidth}x{\w\linewidth}x{\w\linewidth}x{\w\linewidth}x{\w\linewidth}x{\w\linewidth}}
     Input &  Reference &  Ours ($\alpha=0$) & Ours ($\alpha=1$) & PhotoWCT~\cite{photowct} &  WCT$^2$~\cite{wct2} & LinearWCT~\cite{linearWCT} & DSTN~\cite{dstn}  
    \end{tabular}
    \end{minipage}
    
    \caption{More \textbf{photo-to-photo} style transfer results by the state-of-the-art methods.
    }
    \label{fig:photo2photo}
\end{figure*}

\begin{figure*}[t]
    \centering
    \scriptsize

    \includegraphics[width=1\linewidth]{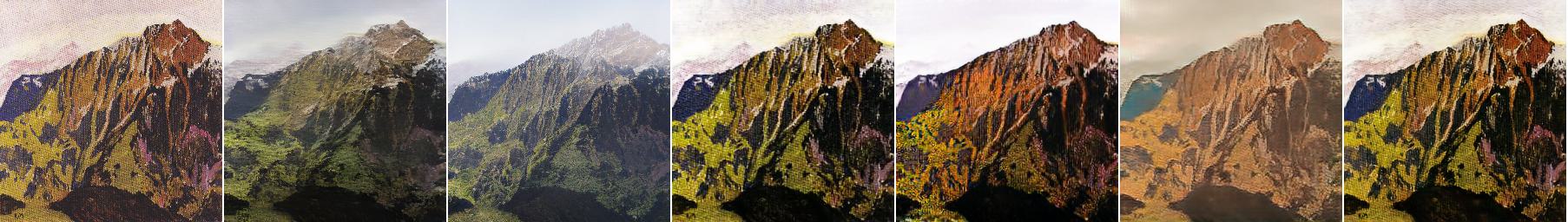}
    \\\vspace{0.3em}
    \includegraphics[width=1\linewidth]{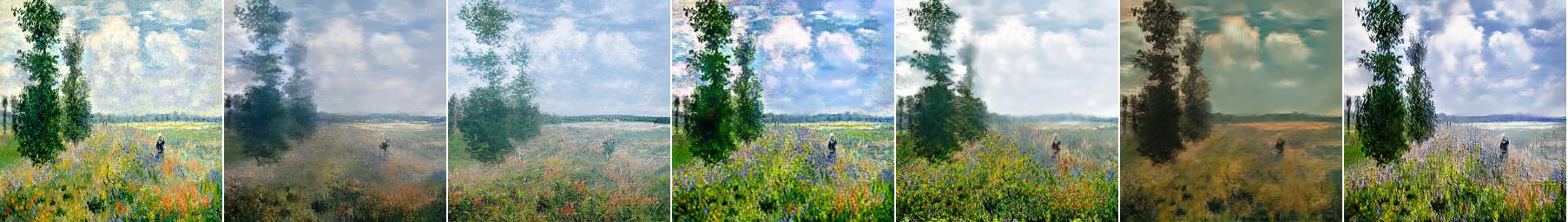}
    \\\vspace{0.3em}
    \includegraphics[width=1\linewidth]{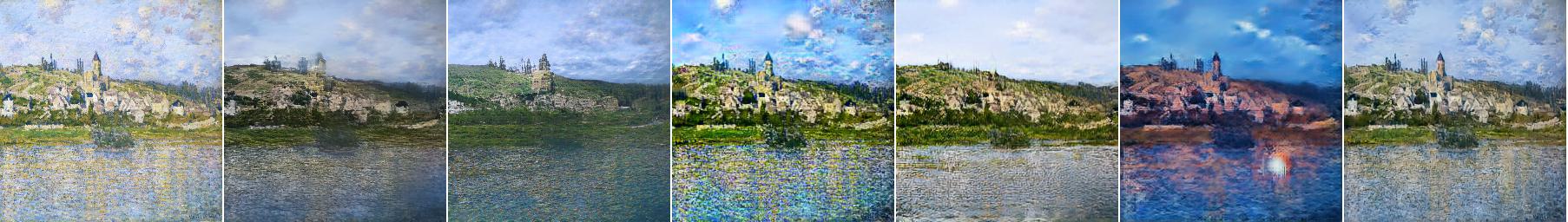}
    \\\vspace{0.3em}
    \includegraphics[width=1\linewidth]{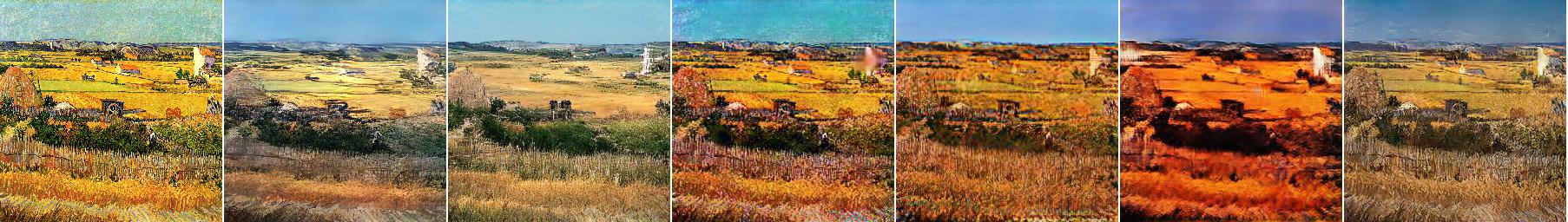}
    \\\vspace{0.3em}
    \includegraphics[width=1\linewidth]{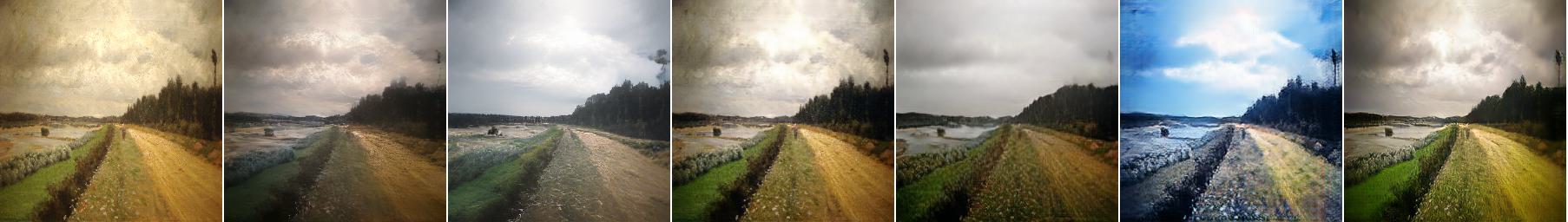}
    \\\vspace{0.3em}
    
    \begin{minipage}{1\linewidth}
    \includegraphics[width=1\linewidth]{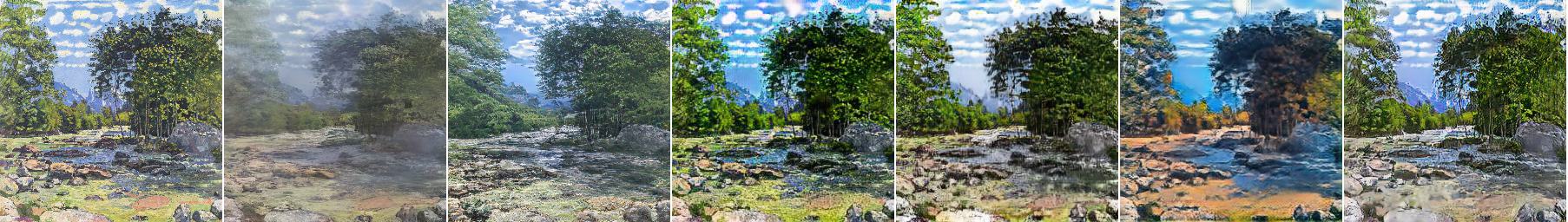}
    \setlength{\tabcolsep}{0pt}
    \vspace{-1.4em}\\
    \newcommand\w{0.14285}
    \begin{tabular}
    {x{\w\linewidth}x{\w\linewidth}x{\w\linewidth}x{\w\linewidth}x{\w\linewidth}x{\w\linewidth}x{\w\linewidth}}
     Input &  Ours ($\alpha=0$) & Ours ($\alpha=1$) & CycleGAN~\cite{cyclegan} & CUT~\cite{cut} &  DRIT~\cite{drit} & Art2Real~\cite{tomei2019art2real}
    \end{tabular}
    \end{minipage}
    
    \caption{More \textbf{artwork-to-photo} style transfer results by the state-of-the-art methods.
    }
    \label{fig:art2photo}
\end{figure*}

\begin{figure*}[t]
    \centering
    \scriptsize
    \newcommand\w{0.1426}
    
    \begin{minipage}{1\linewidth}
    \includegraphics[width=1\linewidth]{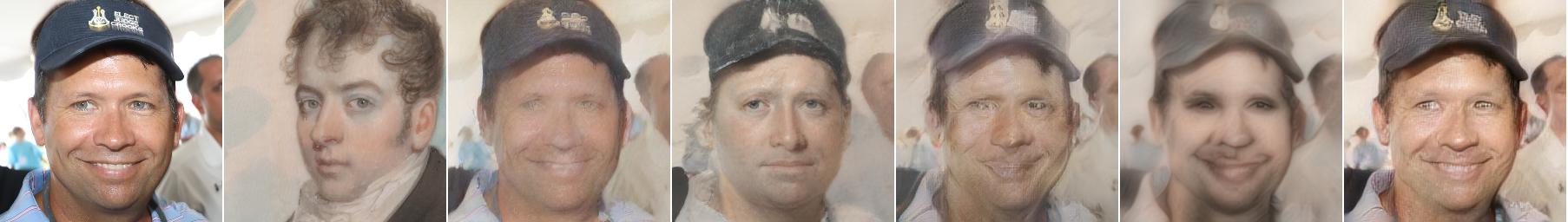}
    \setlength{\tabcolsep}{0pt}
    \vspace{-1.4em}\\
    \begin{tabular}
    {x{\w\linewidth}x{\w\linewidth}x{\w\linewidth}x{\w\linewidth}x{\w\linewidth}x{\w\linewidth}x{\w\linewidth}}
     Input &  Reference &  Ours ($\alpha=0$) & Ours ($\alpha=1$) &  AdaIN~\cite{adain} &  WCT~\cite{wct} &  LinearWCT~\cite{linearWCT} 
    \end{tabular}
    \includegraphics[width=1\linewidth]{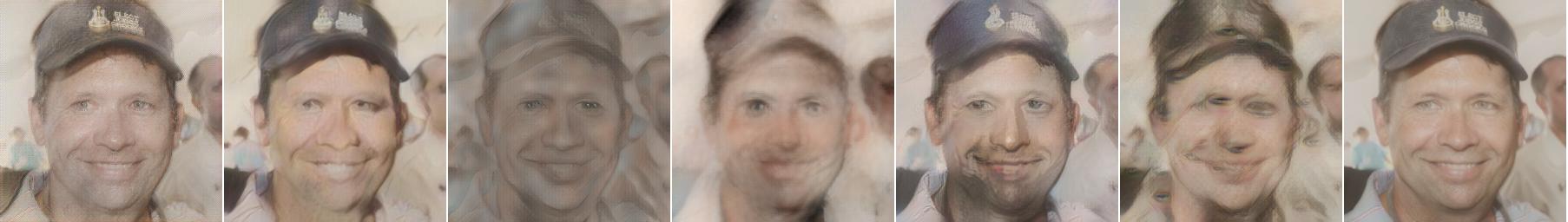}
    \vspace{-1.4em}\\
    \begin{tabular}
    {x{\w\linewidth}x{\w\linewidth}x{\w\linewidth}x{\w\linewidth}x{\w\linewidth}x{\w\linewidth}x{\w\linewidth}}
     ArtFlow~\cite{artflow} &  EFDM~\cite{efdm} &  StyleSwap~\cite{styleswap} &  Avatar-Net~\cite{avatar} &  SANet~\cite{sanet} &  AdaAttN~\cite{adaattn} &  StyTR2~\cite{stytr2}
    \end{tabular}
    \end{minipage}
    \\\hrulefill 
    \vspace{0.3em}

    \begin{minipage}{1\linewidth}
    \includegraphics[width=1\linewidth]{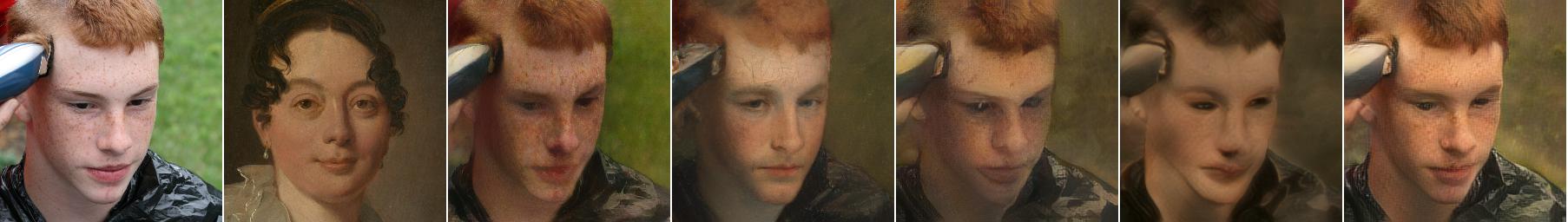}
    \setlength{\tabcolsep}{0pt}
    \vspace{-1.4em}\\
    \begin{tabular}
    {x{\w\linewidth}x{\w\linewidth}x{\w\linewidth}x{\w\linewidth}x{\w\linewidth}x{\w\linewidth}x{\w\linewidth}}
     Input &  Reference &  Ours ($\alpha=0$) & Ours ($\alpha=1$) &  AdaIN~\cite{adain} &  WCT~\cite{wct} &  LinearWCT~\cite{linearWCT} 
    \end{tabular}
    \includegraphics[width=1\linewidth]{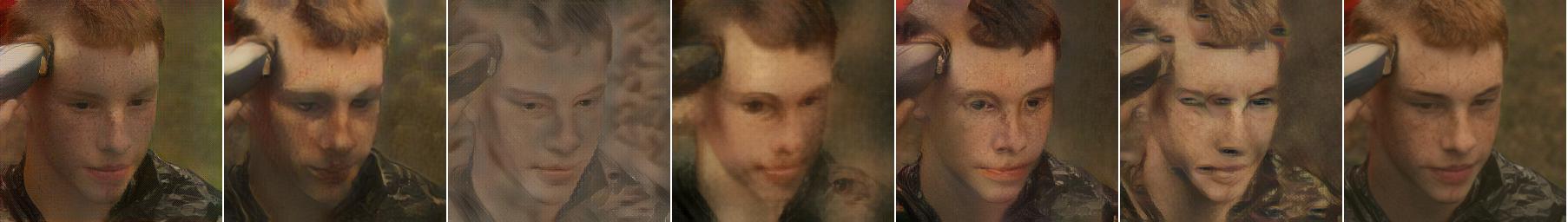}
    \vspace{-1.4em}\\
    \begin{tabular}
    {x{\w\linewidth}x{\w\linewidth}x{\w\linewidth}x{\w\linewidth}x{\w\linewidth}x{\w\linewidth}x{\w\linewidth}}
     ArtFlow~\cite{artflow} &  EFDM~\cite{efdm} &  StyleSwap~\cite{styleswap} &  Avatar-Net~\cite{avatar} &  SANet~\cite{sanet} &  AdaAttN~\cite{adaattn} &  StyTR2~\cite{stytr2}
    \end{tabular}
    \end{minipage}
    \\\hrulefill 
    \vspace{0.3em}

    \begin{minipage}{1\linewidth}
    \includegraphics[width=1\linewidth]{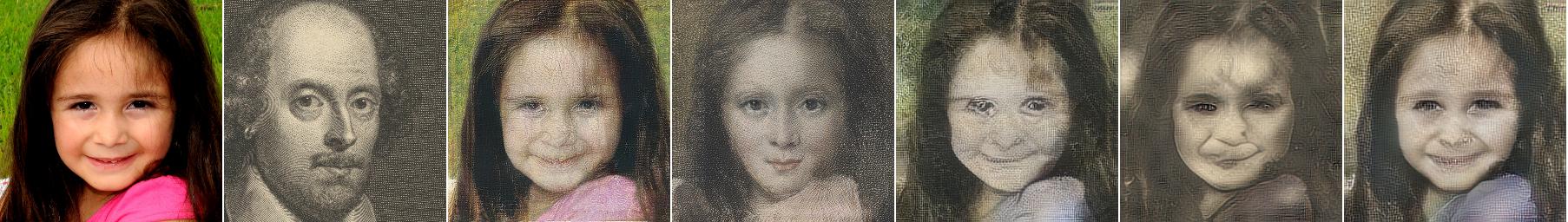}
    \setlength{\tabcolsep}{0pt}
    \vspace{-1.4em}\\
    \begin{tabular}
    {x{\w\linewidth}x{\w\linewidth}x{\w\linewidth}x{\w\linewidth}x{\w\linewidth}x{\w\linewidth}x{\w\linewidth}}
     Input &  Reference &  Ours ($\alpha=0$) & Ours ($\alpha=1$) &  AdaIN~\cite{adain} &  WCT~\cite{wct} &  LinearWCT~\cite{linearWCT} 
    \end{tabular}
    \includegraphics[width=1\linewidth]{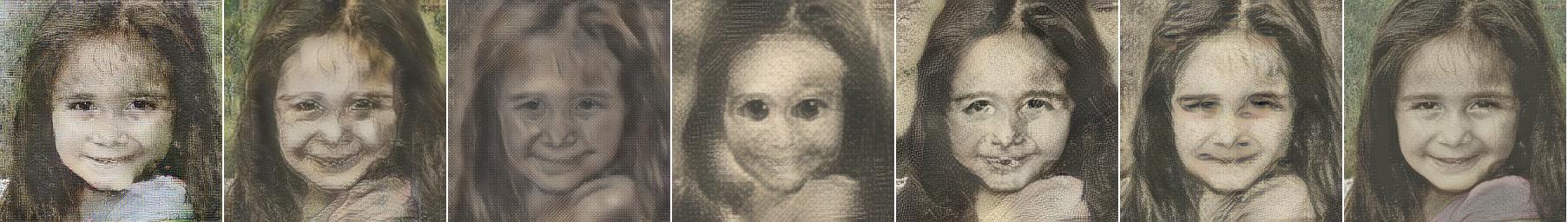}
    \vspace{-1.4em}\\
    \begin{tabular}
    {x{\w\linewidth}x{\w\linewidth}x{\w\linewidth}x{\w\linewidth}x{\w\linewidth}x{\w\linewidth}x{\w\linewidth}}
     ArtFlow~\cite{artflow} &  EFDM~\cite{efdm} &  StyleSwap~\cite{styleswap} &  Avatar-Net~\cite{avatar} &  SANet~\cite{sanet} &  AdaAttN~\cite{adaattn} &  StyTR2~\cite{stytr2}
    \end{tabular}
    \end{minipage}
    
    \caption{More \textbf{face-to-artwork} style transfer results by the state-of-the-art methods.  
    }
    \label{fig:photo2art1}
\end{figure*}

\begin{figure*}[t]
    \centering
    \scriptsize
    \newcommand\w{0.1426}
    
    \begin{minipage}{1\linewidth}
    \includegraphics[width=1\linewidth]{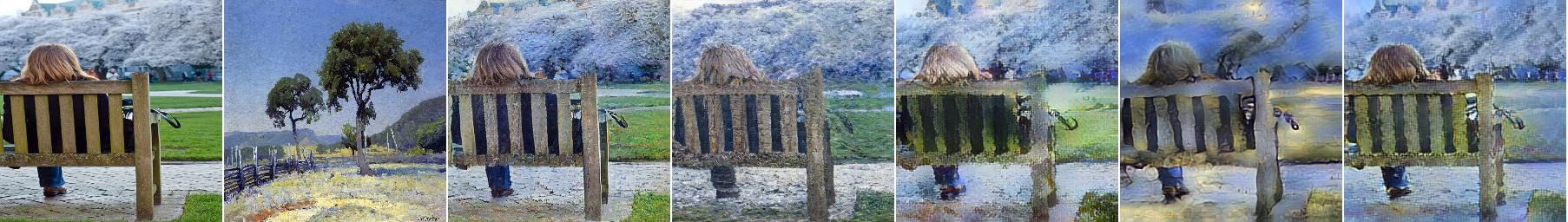}
    \setlength{\tabcolsep}{0pt}
    \vspace{-1.4em}\\
    \begin{tabular}
    {x{\w\linewidth}x{\w\linewidth}x{\w\linewidth}x{\w\linewidth}x{\w\linewidth}x{\w\linewidth}x{\w\linewidth}}
     Input &  Reference &  Ours ($\alpha=0$) & Ours ($\alpha=1$) &  AdaIN~\cite{adain} &  WCT~\cite{wct} &  LinearWCT~\cite{linearWCT} 
    \end{tabular}
    \includegraphics[width=1\linewidth]{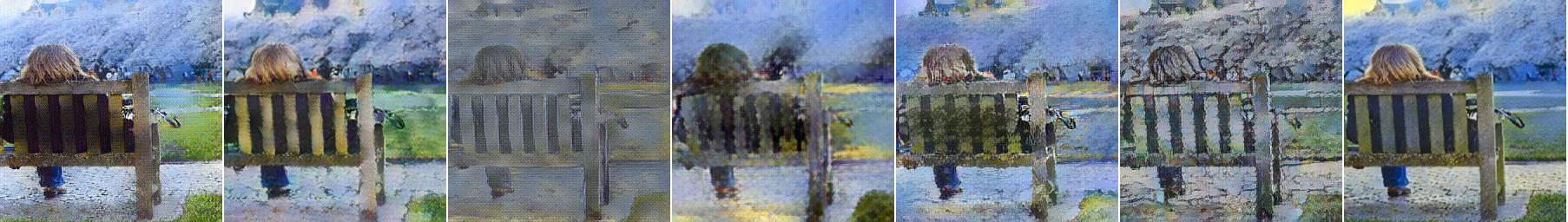}
    \vspace{-1.4em}\\
    \begin{tabular}
    {x{\w\linewidth}x{\w\linewidth}x{\w\linewidth}x{\w\linewidth}x{\w\linewidth}x{\w\linewidth}x{\w\linewidth}}
     ArtFlow~\cite{artflow} &  EFDM~\cite{efdm} &  StyleSwap~\cite{styleswap} &  Avatar-Net~\cite{avatar} &  SANet~\cite{sanet} &  AdaAttN~\cite{adaattn} &  StyTR2~\cite{stytr2}
    \end{tabular}
    \end{minipage}
    \\\hrulefill 
    \vspace{0.3em}

    \begin{minipage}{1\linewidth}
    \includegraphics[width=1\linewidth]{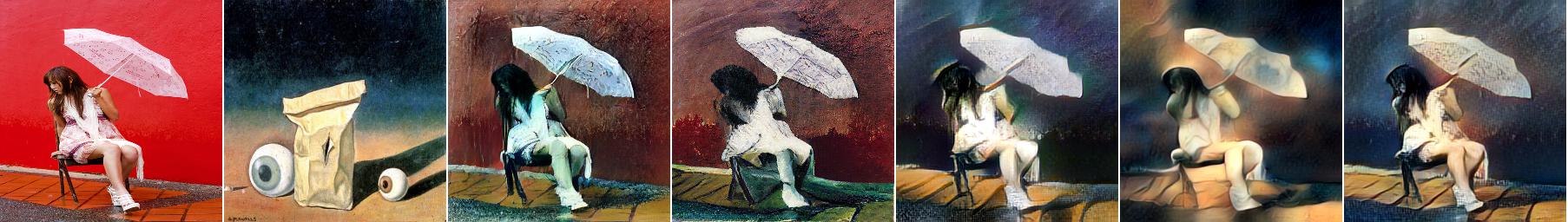}
    \setlength{\tabcolsep}{0pt}
    \vspace{-1.4em}\\
    \begin{tabular}
    {x{\w\linewidth}x{\w\linewidth}x{\w\linewidth}x{\w\linewidth}x{\w\linewidth}x{\w\linewidth}x{\w\linewidth}}
     Input &  Reference &  Ours ($\alpha=0$) & Ours ($\alpha=1$) &  AdaIN~\cite{adain} &  WCT~\cite{wct} &  LinearWCT~\cite{linearWCT} 
    \end{tabular}
    \includegraphics[width=1\linewidth]{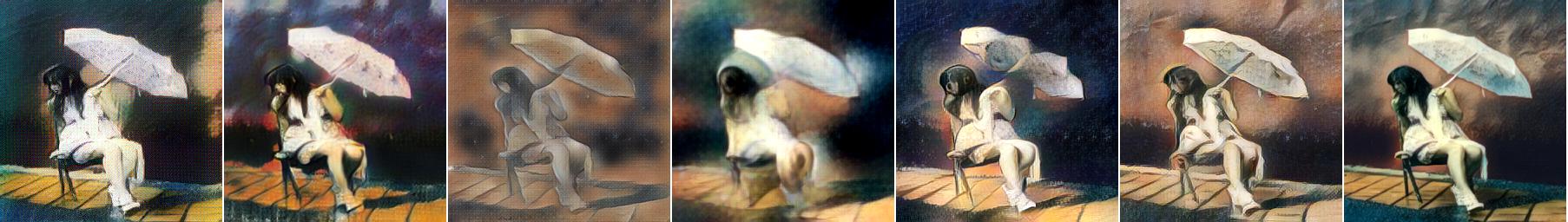}
    \vspace{-1.4em}\\
    \begin{tabular}
    {x{\w\linewidth}x{\w\linewidth}x{\w\linewidth}x{\w\linewidth}x{\w\linewidth}x{\w\linewidth}x{\w\linewidth}}
     ArtFlow~\cite{artflow} &  EFDM~\cite{efdm} &  StyleSwap~\cite{styleswap} &  Avatar-Net~\cite{avatar} &  SANet~\cite{sanet} &  AdaAttN~\cite{adaattn} &  StyTR2~\cite{stytr2}
    \end{tabular}
    \end{minipage}
    \\\hrulefill 
    \vspace{0.3em}

    \begin{minipage}{1\linewidth}
    \includegraphics[width=1\linewidth]{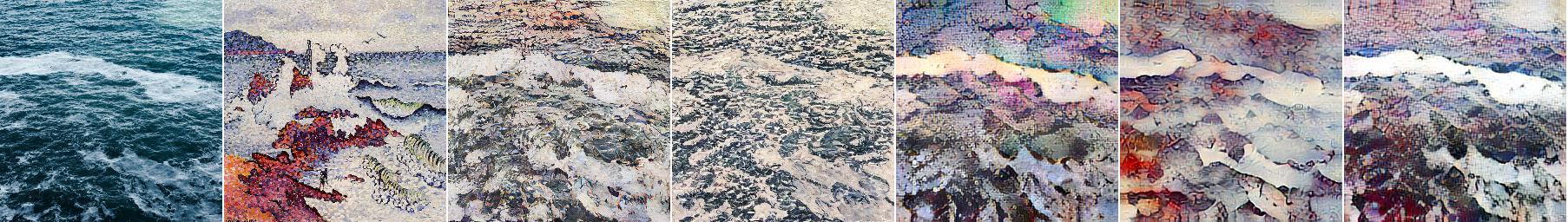}
    \setlength{\tabcolsep}{0pt}
    \vspace{-1.4em}\\
    \begin{tabular}
    {x{\w\linewidth}x{\w\linewidth}x{\w\linewidth}x{\w\linewidth}x{\w\linewidth}x{\w\linewidth}x{\w\linewidth}}
     Input &  Reference &  Ours ($\alpha=0$) & Ours ($\alpha=1$) &  AdaIN~\cite{adain} &  WCT~\cite{wct} &  LinearWCT~\cite{linearWCT} 
    \end{tabular}
    \includegraphics[width=1\linewidth]{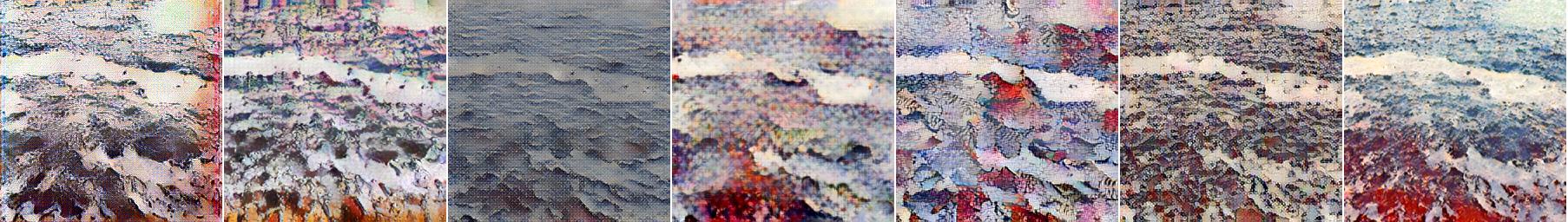}
    \vspace{-1.4em}\\
    \begin{tabular}
    {x{\w\linewidth}x{\w\linewidth}x{\w\linewidth}x{\w\linewidth}x{\w\linewidth}x{\w\linewidth}x{\w\linewidth}}
     ArtFlow~\cite{artflow} &  EFDM~\cite{efdm} &  StyleSwap~\cite{styleswap} &  Avatar-Net~\cite{avatar} &  SANet~\cite{sanet} &  AdaAttN~\cite{adaattn} &  StyTR2~\cite{stytr2}
    \end{tabular}
    \end{minipage}
    
    \caption{More \textbf{photo-to-artwork} style transfer results by the state-of-the-art methods.  
    }
    \label{fig:photo2art2}
\end{figure*}

\begin{figure*}[t]
    \centering
    \scriptsize
    \newcommand\w{0.1426}
    
    \begin{minipage}{1\linewidth}
    \includegraphics[width=1\linewidth]{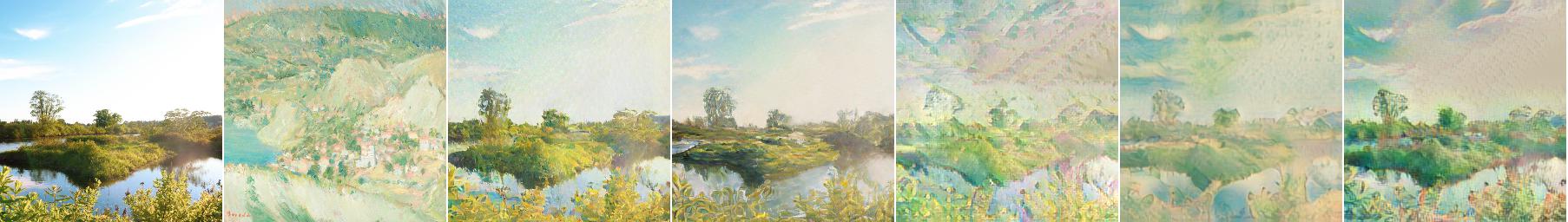}
    \setlength{\tabcolsep}{0pt}
    \vspace{-1.4em}\\
    \begin{tabular}
    {x{\w\linewidth}x{\w\linewidth}x{\w\linewidth}x{\w\linewidth}x{\w\linewidth}x{\w\linewidth}x{\w\linewidth}}
     Input &  Reference &  Ours ($\alpha=0$) & Ours ($\alpha=1$) &  AdaIN~\cite{adain} &  WCT~\cite{wct} &  LinearWCT~\cite{linearWCT} 
    \end{tabular}
    \includegraphics[width=1\linewidth]{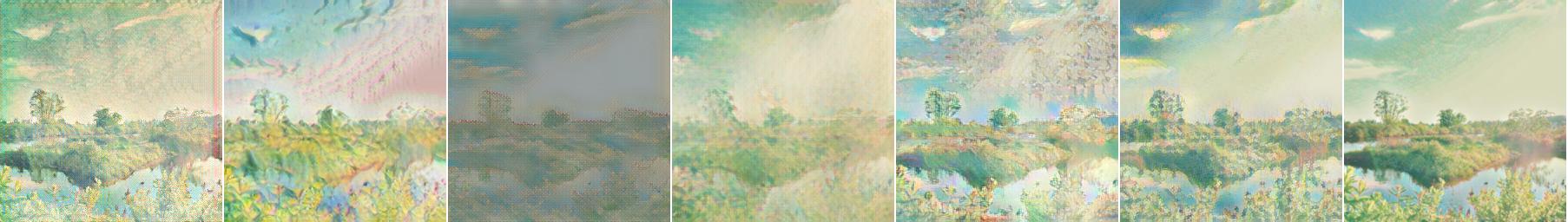}
    \vspace{-1.4em}\\
    \begin{tabular}
    {x{\w\linewidth}x{\w\linewidth}x{\w\linewidth}x{\w\linewidth}x{\w\linewidth}x{\w\linewidth}x{\w\linewidth}}
     ArtFlow~\cite{artflow} &  EFDM~\cite{efdm} &  StyleSwap~\cite{styleswap} &  Avatar-Net~\cite{avatar} &  SANet~\cite{sanet} &  AdaAttN~\cite{adaattn} &  StyTR2~\cite{stytr2}
    \end{tabular}
    \end{minipage}
    \\\hrulefill 
    \vspace{0.3em}

    \begin{minipage}{1\linewidth}
    \includegraphics[width=1\linewidth]{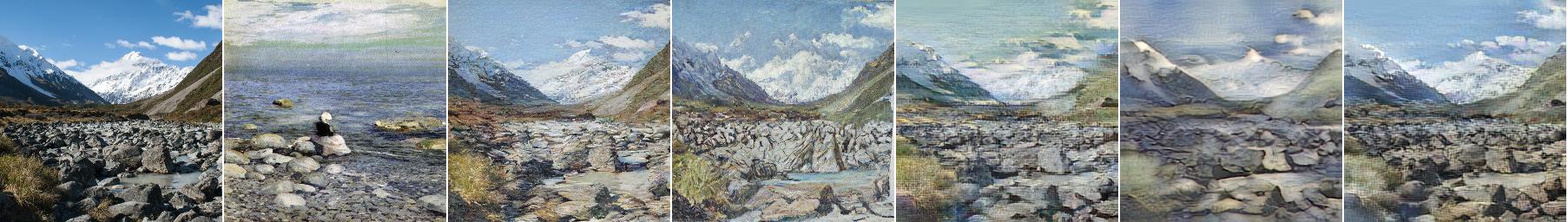}
    \setlength{\tabcolsep}{0pt}
    \vspace{-1.4em}\\
    \begin{tabular}
    {x{\w\linewidth}x{\w\linewidth}x{\w\linewidth}x{\w\linewidth}x{\w\linewidth}x{\w\linewidth}x{\w\linewidth}}
     Input &  Reference &  Ours ($\alpha=0$) & Ours ($\alpha=1$) &  AdaIN~\cite{adain} &  WCT~\cite{wct} &  LinearWCT~\cite{linearWCT} 
    \end{tabular}
    \includegraphics[width=1\linewidth]{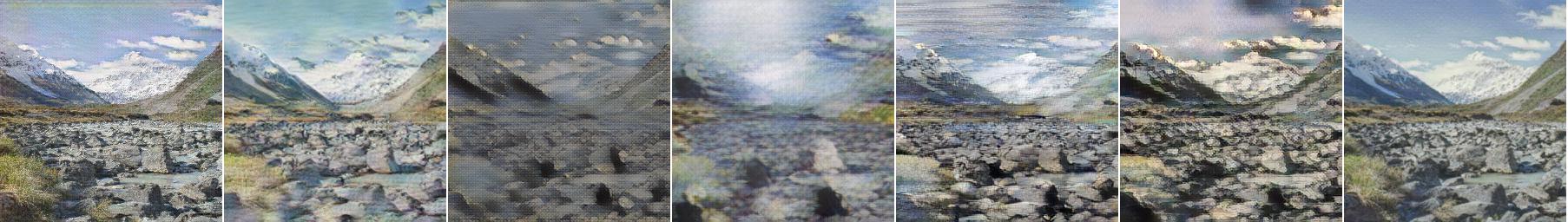}
    \vspace{-1.4em}\\
    \begin{tabular}
    {x{\w\linewidth}x{\w\linewidth}x{\w\linewidth}x{\w\linewidth}x{\w\linewidth}x{\w\linewidth}x{\w\linewidth}}
     ArtFlow~\cite{artflow} &  EFDM~\cite{efdm} &  StyleSwap~\cite{styleswap} &  Avatar-Net~\cite{avatar} &  SANet~\cite{sanet} &  AdaAttN~\cite{adaattn} &  StyTR2~\cite{stytr2}
    \end{tabular}
    \end{minipage}
    \\\hrulefill 
    \vspace{0.3em}

    \begin{minipage}{1\linewidth}
    \includegraphics[width=1\linewidth]{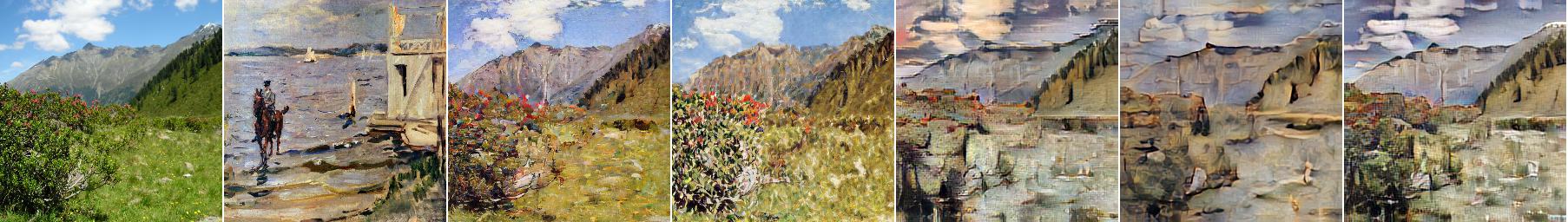}
    \setlength{\tabcolsep}{0pt}
    \vspace{-1.4em}\\
    \begin{tabular}
    {x{\w\linewidth}x{\w\linewidth}x{\w\linewidth}x{\w\linewidth}x{\w\linewidth}x{\w\linewidth}x{\w\linewidth}}
     Input &  Reference &  Ours ($\alpha=0$) & Ours ($\alpha=1$) &  AdaIN~\cite{adain} &  WCT~\cite{wct} &  LinearWCT~\cite{linearWCT} 
    \end{tabular}
    \includegraphics[width=1\linewidth]{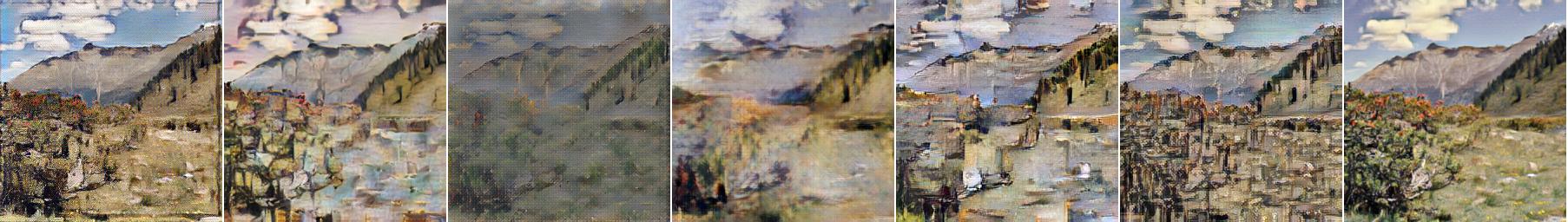}
    \vspace{-1.4em}\\
    \begin{tabular}
    {x{\w\linewidth}x{\w\linewidth}x{\w\linewidth}x{\w\linewidth}x{\w\linewidth}x{\w\linewidth}x{\w\linewidth}}
     ArtFlow~\cite{artflow} &  EFDM~\cite{efdm} &  StyleSwap~\cite{styleswap} &  Avatar-Net~\cite{avatar} &  SANet~\cite{sanet} &  AdaAttN~\cite{adaattn} &  StyTR2~\cite{stytr2}
    \end{tabular}
    \end{minipage}
    
    \caption{More \textbf{photo-to-artwork} style transfer results by the state-of-the-art methods.  
    }
    \label{fig:photo2art3}
\end{figure*}

\begin{figure*}[t]
    \centering
    \scriptsize
    \newcommand\w{0.1426}
    
    \begin{minipage}{1\linewidth}
    \includegraphics[width=1\linewidth]{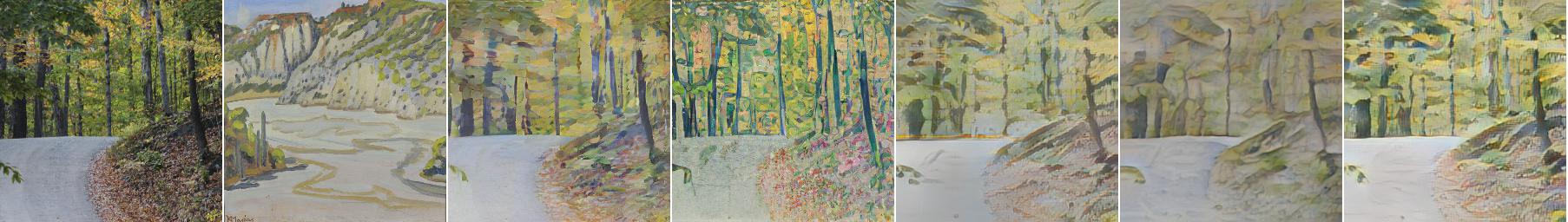}
    \setlength{\tabcolsep}{0pt}
    \vspace{-1.4em}\\
    \begin{tabular}
    {x{\w\linewidth}x{\w\linewidth}x{\w\linewidth}x{\w\linewidth}x{\w\linewidth}x{\w\linewidth}x{\w\linewidth}}
     Input &  Reference &  Ours ($\alpha=0$) & Ours ($\alpha=1$) &  AdaIN~\cite{adain} &  WCT~\cite{wct} &  LinearWCT~\cite{linearWCT} 
    \end{tabular}
    \includegraphics[width=1\linewidth]{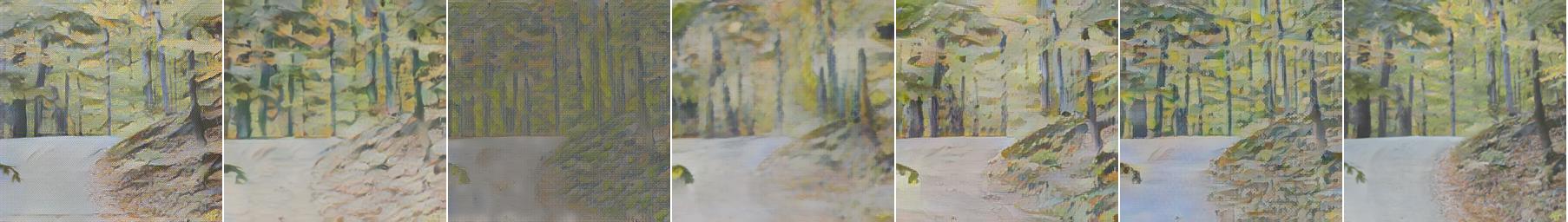}
    \vspace{-1.4em}\\
    \begin{tabular}
    {x{\w\linewidth}x{\w\linewidth}x{\w\linewidth}x{\w\linewidth}x{\w\linewidth}x{\w\linewidth}x{\w\linewidth}}
     ArtFlow~\cite{artflow} &  EFDM~\cite{efdm} &  StyleSwap~\cite{styleswap} &  Avatar-Net~\cite{avatar} &  SANet~\cite{sanet} &  AdaAttN~\cite{adaattn} &  StyTR2~\cite{stytr2}
    \end{tabular}
    \end{minipage}
    \\\hrulefill 
    \vspace{0.3em}

    \begin{minipage}{1\linewidth}
    \includegraphics[width=1\linewidth]{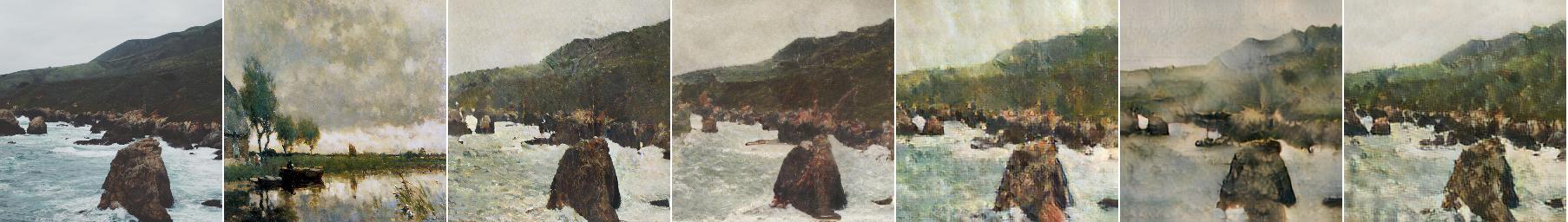}
    \setlength{\tabcolsep}{0pt}
    \vspace{-1.4em}\\
    \begin{tabular}
    {x{\w\linewidth}x{\w\linewidth}x{\w\linewidth}x{\w\linewidth}x{\w\linewidth}x{\w\linewidth}x{\w\linewidth}}
     Input &  Reference &  Ours ($\alpha=0$) & Ours ($\alpha=1$) &  AdaIN~\cite{adain} &  WCT~\cite{wct} &  LinearWCT~\cite{linearWCT} 
    \end{tabular}
    \includegraphics[width=1\linewidth]{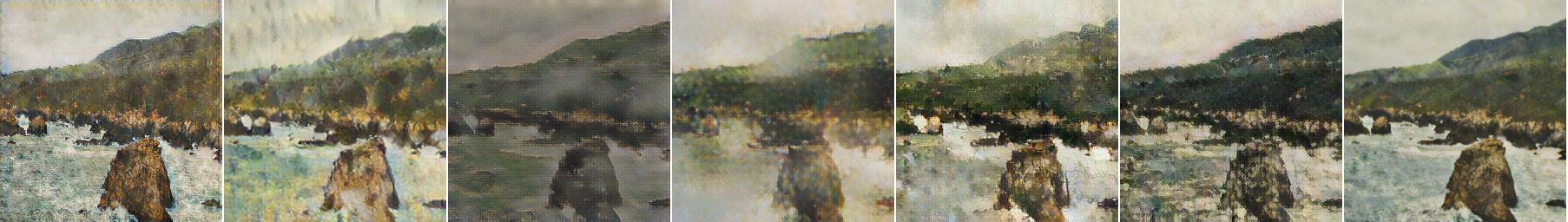}
    \vspace{-1.4em}\\
    \begin{tabular}
    {x{\w\linewidth}x{\w\linewidth}x{\w\linewidth}x{\w\linewidth}x{\w\linewidth}x{\w\linewidth}x{\w\linewidth}}
     ArtFlow~\cite{artflow} &  EFDM~\cite{efdm} &  StyleSwap~\cite{styleswap} &  Avatar-Net~\cite{avatar} &  SANet~\cite{sanet} &  AdaAttN~\cite{adaattn} &  StyTR2~\cite{stytr2}
    \end{tabular}
    \end{minipage}
    \\\hrulefill 
    \vspace{0.3em}

    \begin{minipage}{1\linewidth}
    \includegraphics[width=1\linewidth]{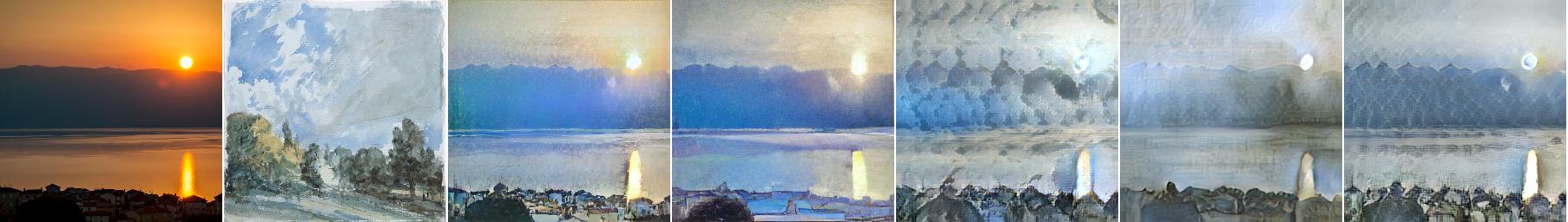}
    \setlength{\tabcolsep}{0pt}
    \vspace{-1.4em}\\
    \begin{tabular}
    {x{\w\linewidth}x{\w\linewidth}x{\w\linewidth}x{\w\linewidth}x{\w\linewidth}x{\w\linewidth}x{\w\linewidth}}
     Input &  Reference &  Ours ($\alpha=0$) & Ours ($\alpha=1$) &  AdaIN~\cite{adain} &  WCT~\cite{wct} &  LinearWCT~\cite{linearWCT} 
    \end{tabular}
    \includegraphics[width=1\linewidth]{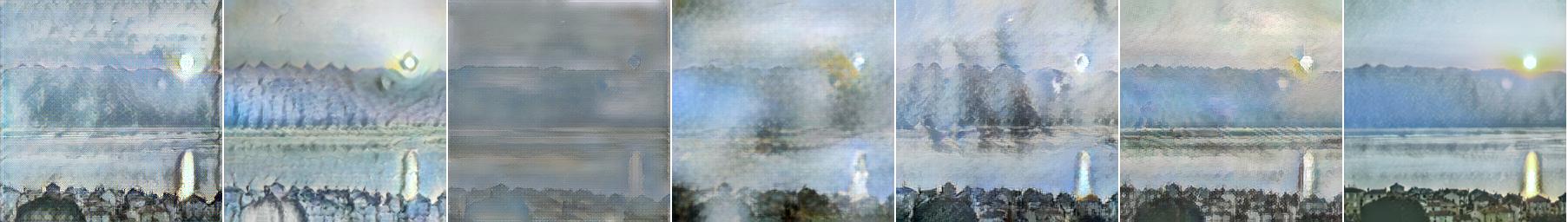}
    \vspace{-1.4em}\\
    \begin{tabular}
    {x{\w\linewidth}x{\w\linewidth}x{\w\linewidth}x{\w\linewidth}x{\w\linewidth}x{\w\linewidth}x{\w\linewidth}}
     ArtFlow~\cite{artflow} &  EFDM~\cite{efdm} &  StyleSwap~\cite{styleswap} &  Avatar-Net~\cite{avatar} &  SANet~\cite{sanet} &  AdaAttN~\cite{adaattn} &  StyTR2~\cite{stytr2}
    \end{tabular}
    \end{minipage}
    
    \caption{More \textbf{photo-to-artwork} style transfer results by the state-of-the-art methods.  
    }
    \label{fig:photo2art4}
\end{figure*}

\begin{figure*}[t]
    \centering
    \scriptsize
    \newcommand\w{0.1426}
    
    \begin{minipage}{1\linewidth}
    \includegraphics[width=1\linewidth]{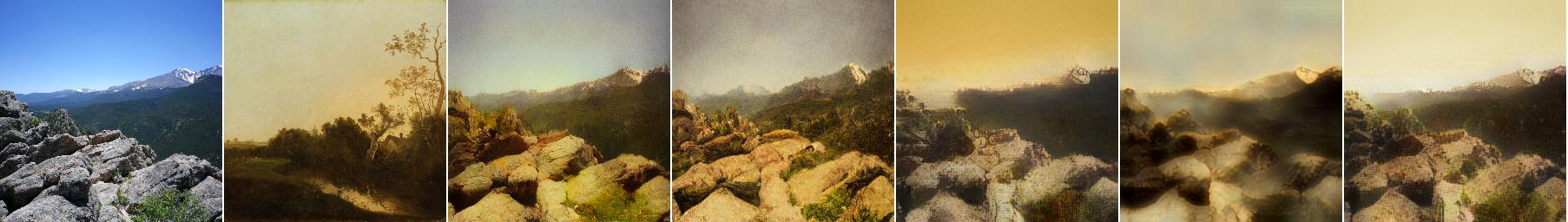}
    \setlength{\tabcolsep}{0pt}
    \vspace{-1.4em}\\
    \begin{tabular}
    {x{\w\linewidth}x{\w\linewidth}x{\w\linewidth}x{\w\linewidth}x{\w\linewidth}x{\w\linewidth}x{\w\linewidth}}
     Input &  Reference &  Ours ($\alpha=0$) & Ours ($\alpha=1$) &  AdaIN~\cite{adain} &  WCT~\cite{wct} &  LinearWCT~\cite{linearWCT} 
    \end{tabular}
    \includegraphics[width=1\linewidth]{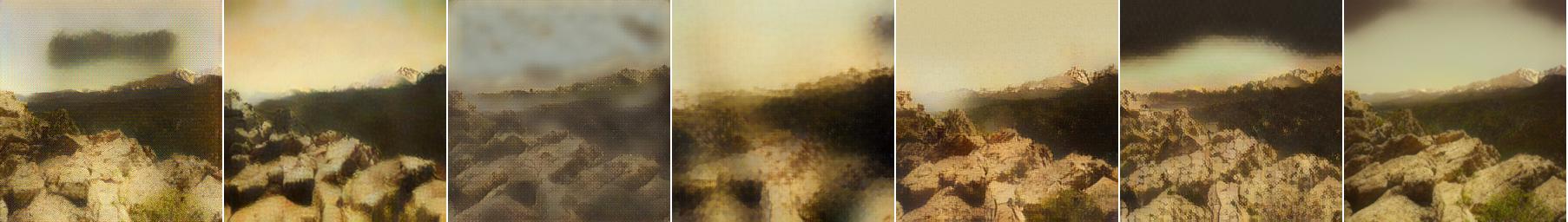}
    \vspace{-1.4em}\\
    \begin{tabular}
    {x{\w\linewidth}x{\w\linewidth}x{\w\linewidth}x{\w\linewidth}x{\w\linewidth}x{\w\linewidth}x{\w\linewidth}}
     ArtFlow~\cite{artflow} &  EFDM~\cite{efdm} &  StyleSwap~\cite{styleswap} &  Avatar-Net~\cite{avatar} &  SANet~\cite{sanet} &  AdaAttN~\cite{adaattn} &  StyTR2~\cite{stytr2}
    \end{tabular}
    \end{minipage}
    \\\hrulefill 
    \vspace{0.3em}

    \begin{minipage}{1\linewidth}
    \includegraphics[width=1\linewidth]{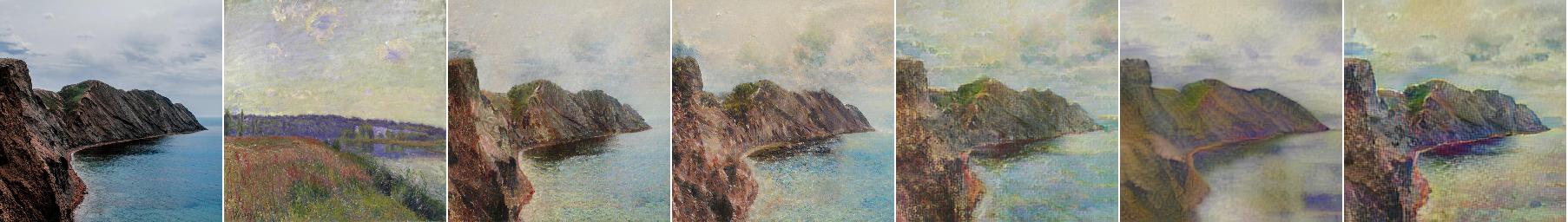}
    \setlength{\tabcolsep}{0pt}
    \vspace{-1.4em}\\
    \begin{tabular}
    {x{\w\linewidth}x{\w\linewidth}x{\w\linewidth}x{\w\linewidth}x{\w\linewidth}x{\w\linewidth}x{\w\linewidth}}
     Input &  Reference &  Ours ($\alpha=0$) & Ours ($\alpha=1$) &  AdaIN~\cite{adain} &  WCT~\cite{wct} &  LinearWCT~\cite{linearWCT} 
    \end{tabular}
    \includegraphics[width=1\linewidth]{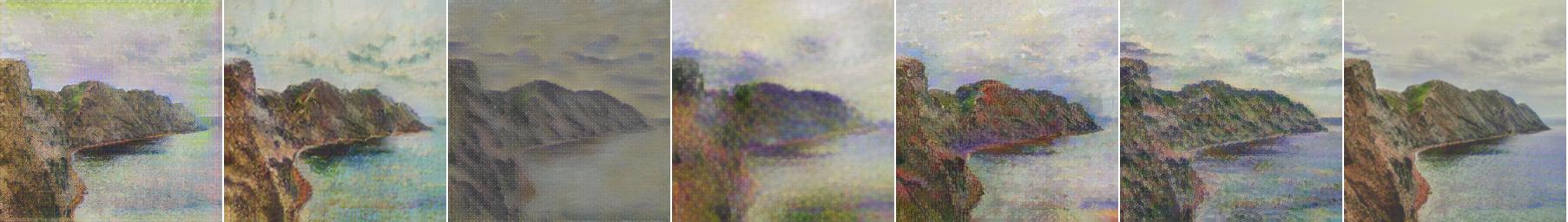}
    \vspace{-1.4em}\\
    \begin{tabular}
    {x{\w\linewidth}x{\w\linewidth}x{\w\linewidth}x{\w\linewidth}x{\w\linewidth}x{\w\linewidth}x{\w\linewidth}}
     ArtFlow~\cite{artflow} &  EFDM~\cite{efdm} &  StyleSwap~\cite{styleswap} &  Avatar-Net~\cite{avatar} &  SANet~\cite{sanet} &  AdaAttN~\cite{adaattn} &  StyTR2~\cite{stytr2}
    \end{tabular}
    \end{minipage}
    \\\hrulefill 
    \vspace{0.3em}

    \begin{minipage}{1\linewidth}
    \includegraphics[width=1\linewidth]{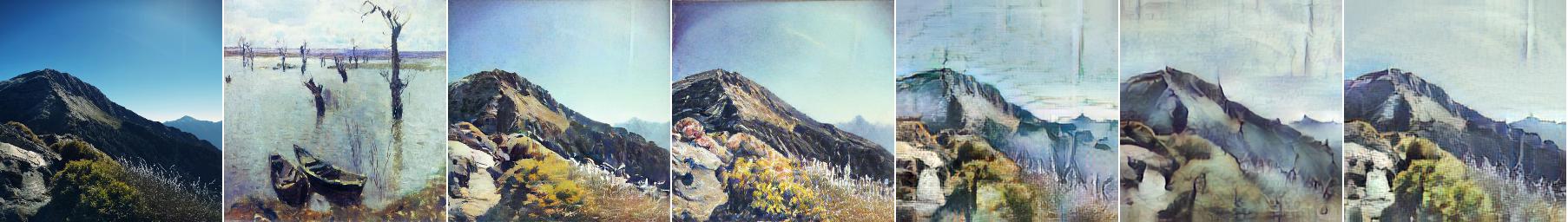}
    \setlength{\tabcolsep}{0pt}
    \vspace{-1.4em}\\
    \begin{tabular}
    {x{\w\linewidth}x{\w\linewidth}x{\w\linewidth}x{\w\linewidth}x{\w\linewidth}x{\w\linewidth}x{\w\linewidth}}
     Input &  Reference &  Ours ($\alpha=0$) & Ours ($\alpha=1$) &  AdaIN~\cite{adain} &  WCT~\cite{wct} &  LinearWCT~\cite{linearWCT} 
    \end{tabular}
    \includegraphics[width=1\linewidth]{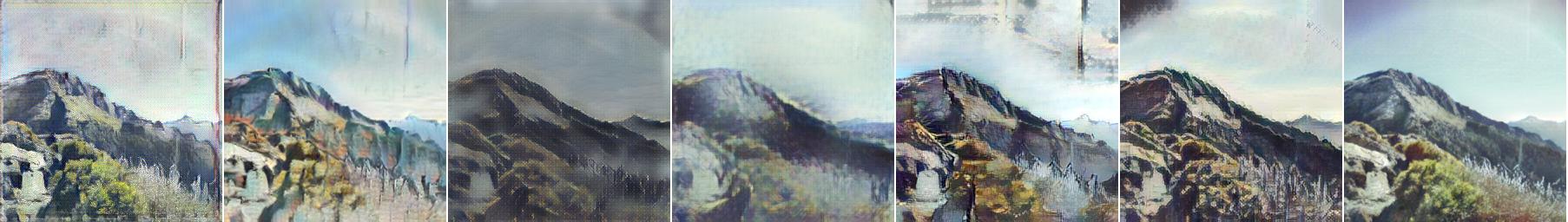}
    \vspace{-1.4em}\\
    \begin{tabular}
    {x{\w\linewidth}x{\w\linewidth}x{\w\linewidth}x{\w\linewidth}x{\w\linewidth}x{\w\linewidth}x{\w\linewidth}}
     ArtFlow~\cite{artflow} &  EFDM~\cite{efdm} &  StyleSwap~\cite{styleswap} &  Avatar-Net~\cite{avatar} &  SANet~\cite{sanet} &  AdaAttN~\cite{adaattn} &  StyTR2~\cite{stytr2}
    \end{tabular}
    \end{minipage}
    
    \caption{More \textbf{photo-to-artwork} style transfer results by the state-of-the-art methods.  
    }
    \label{fig:photo2art5}
\end{figure*}

Figs.~\ref{fig:art2art1} and~\ref{fig:art2art2} show more examples of artwork-to-artwork image style transfer.
Fig.~\ref{fig:photo2photo} shows more examples of photo-to-photo image style transfer.
Fig.~\ref{fig:art2photo} shows more examples of artwork-to-photo image style transfer.
Figs.~\ref{fig:photo2art1},~\ref{fig:photo2art2},~\ref{fig:photo2art3},~\ref{fig:photo2art4},~\ref{fig:photo2art5} show more examples of photo-to-artwork image style transfer.

\section{Examples of Artistic Style Transfer Confusion Test}
\label{sec:turing}

\begin{figure*}[t]
\label{fig:turing1}
\centering
\begin{minipage}[c]{0.195\linewidth}
\centering \small Real Artwork
\end{minipage}
\begin{minipage}[c]{0.195\linewidth}
\centering \small Fake Artwork (Ours)
\end{minipage}
\begin{minipage}[c]{0.195\linewidth}
\centering \small Content Image
\end{minipage}
\begin{minipage}[c]{0.195\linewidth}
\centering \small User Choice
\end{minipage}
\\ \vspace{0.1em}

\begin{minipage}[c]{0.195\linewidth}
\includegraphics[width=\textwidth]{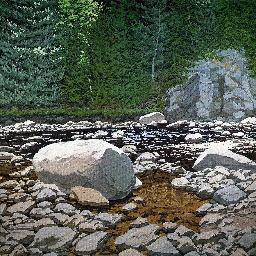}
\end{minipage}
\begin{minipage}[c]{0.195\linewidth}
\includegraphics[width=\textwidth]{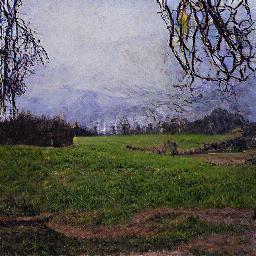}
\end{minipage}
\begin{minipage}[c]{0.195\linewidth}
\includegraphics[width=\textwidth]{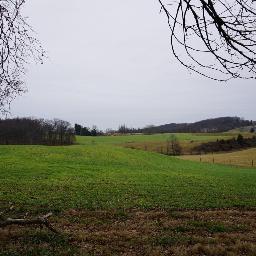}
\end{minipage}
\begin{minipage}[c]{0.195\linewidth}
\includegraphics[width=\textwidth]{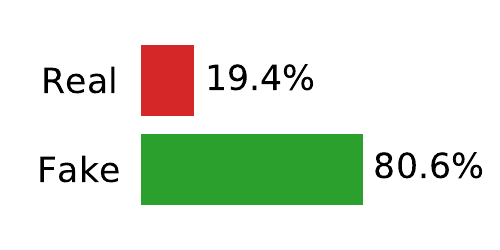}
\end{minipage}
\\ \vspace{0.4em}

\begin{minipage}[c]{0.195\linewidth}
\includegraphics[width=\textwidth]{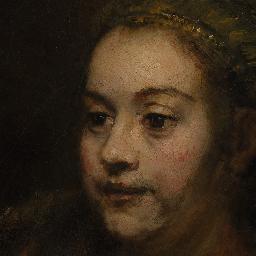}
\end{minipage}
\begin{minipage}[c]{0.195\linewidth}
\includegraphics[width=\textwidth]{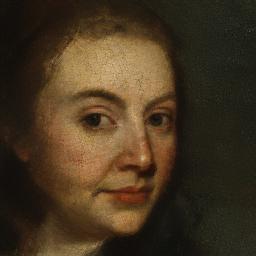}
\end{minipage}
\begin{minipage}[c]{0.195\linewidth}
\includegraphics[width=\textwidth]{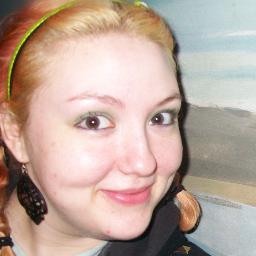}
\end{minipage}
\begin{minipage}[c]{0.195\linewidth}
\includegraphics[width=\textwidth]{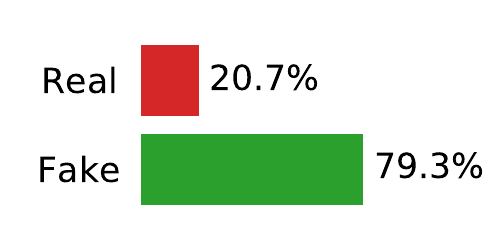}
\end{minipage}
\\ \vspace{0.4em}

\begin{minipage}[c]{0.195\linewidth}
\includegraphics[width=\textwidth]{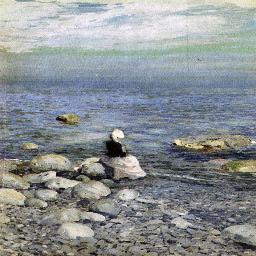}
\end{minipage}
\begin{minipage}[c]{0.195\linewidth}
\includegraphics[width=\textwidth]{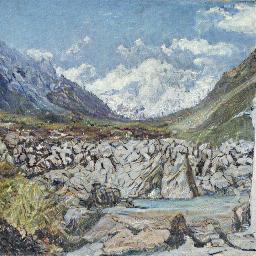}
\end{minipage}
\begin{minipage}[c]{0.195\linewidth}
\includegraphics[width=\textwidth]{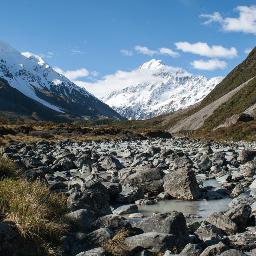}
\end{minipage}
\begin{minipage}[c]{0.195\linewidth}
\includegraphics[width=\textwidth]{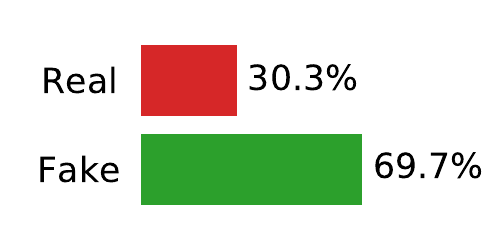}
\end{minipage}
\\ \vspace{0.4em}

\begin{minipage}[c]{0.195\linewidth}
\includegraphics[width=\textwidth]{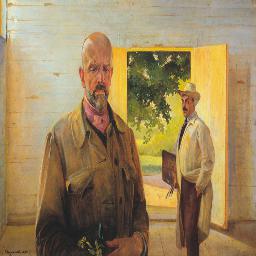}
\end{minipage}
\begin{minipage}[c]{0.195\linewidth}
\includegraphics[width=\textwidth]{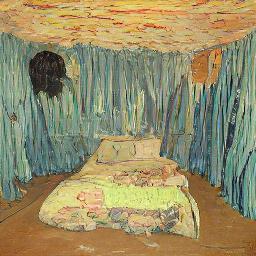}
\end{minipage}
\begin{minipage}[c]{0.195\linewidth}
\includegraphics[width=\textwidth]{figs/results/turing_all/3_content.jpg}
\end{minipage}
\begin{minipage}[c]{0.195\linewidth}
\includegraphics[width=\textwidth]{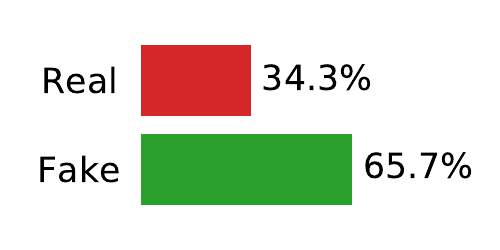}
\end{minipage}
\\ \vspace{0.4em}

\begin{minipage}[c]{0.195\linewidth}
\includegraphics[width=\textwidth]{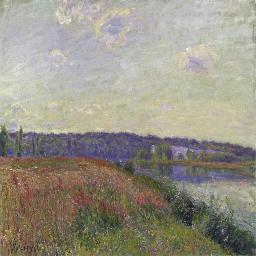}
\end{minipage}
\begin{minipage}[c]{0.195\linewidth}
\includegraphics[width=\textwidth]{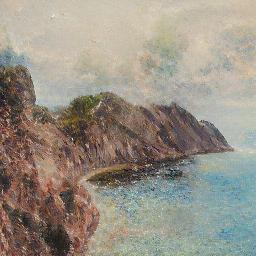}
\end{minipage}
\begin{minipage}[c]{0.195\linewidth}
\includegraphics[width=\textwidth]{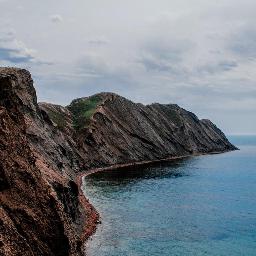}
\end{minipage}
\begin{minipage}[c]{0.195\linewidth}
\includegraphics[width=\textwidth]{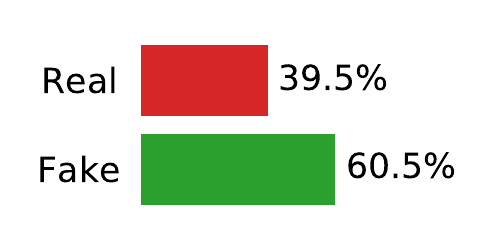}
\end{minipage}
\\ \vspace{0.4em}

\begin{minipage}[c]{0.195\linewidth}
\includegraphics[width=\textwidth]{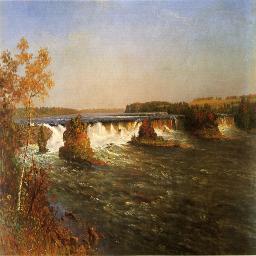}
\end{minipage}
\begin{minipage}[c]{0.195\linewidth}
\includegraphics[width=\textwidth]{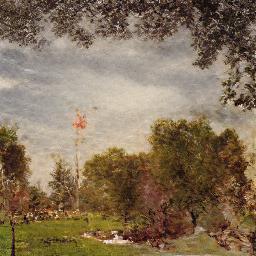}
\end{minipage}
\begin{minipage}[c]{0.195\linewidth}
\includegraphics[width=\textwidth]{figs/results/turing_all/13_content.jpg}
\end{minipage}
\begin{minipage}[c]{0.195\linewidth}
\includegraphics[width=\textwidth]{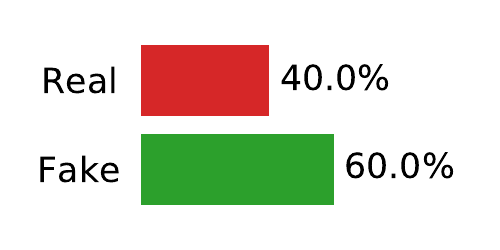}
\end{minipage}

\caption{Examples of artistic confusion test (1/6). The figures are sorted in an ascending order of the ratios of users choosing the real artwork. }
\label{fig:turing-1}
\end{figure*}

\begin{figure*}[t]
\centering
\begin{minipage}[c]{0.195\linewidth}
\centering \small Real Artwork
\end{minipage}
\begin{minipage}[c]{0.195\linewidth}
\centering \small Fake Artwork (Ours)
\end{minipage}
\begin{minipage}[c]{0.195\linewidth}
\centering \small Content Image
\end{minipage}
\begin{minipage}[c]{0.195\linewidth}
\centering \small User Choice
\end{minipage}
\\ \vspace{0.1em}

\begin{minipage}[c]{0.195\linewidth}
\includegraphics[width=\textwidth]{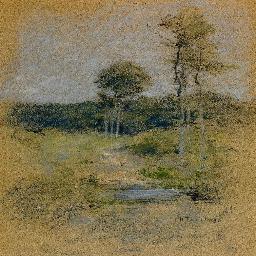}
\end{minipage}
\begin{minipage}[c]{0.195\linewidth}
\includegraphics[width=\textwidth]{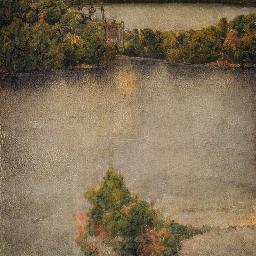}
\end{minipage}
\begin{minipage}[c]{0.195\linewidth}
\includegraphics[width=\textwidth]{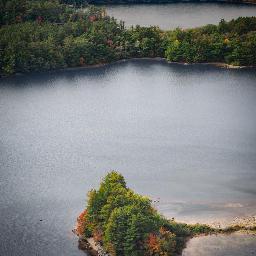}
\end{minipage}
\begin{minipage}[c]{0.195\linewidth}
\includegraphics[width=\textwidth]{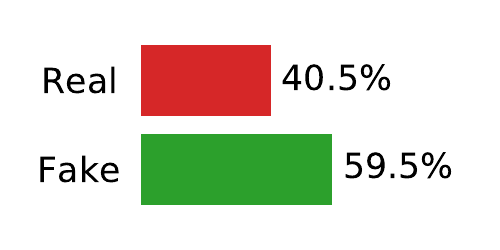}
\end{minipage}
\\ \vspace{0.4em}

\begin{minipage}[c]{0.195\linewidth}
\includegraphics[width=\textwidth]{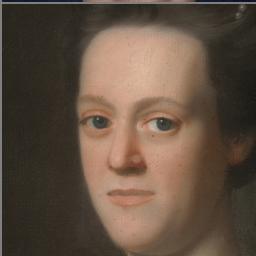}
\end{minipage}
\begin{minipage}[c]{0.195\linewidth}
\includegraphics[width=\textwidth]{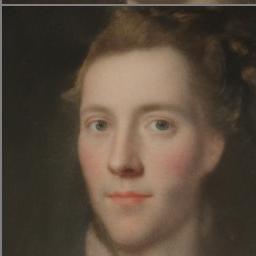}
\end{minipage}
\begin{minipage}[c]{0.195\linewidth}
\includegraphics[width=\textwidth]{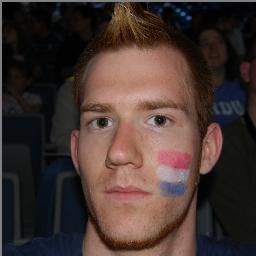}
\end{minipage}
\begin{minipage}[c]{0.195\linewidth}
\includegraphics[width=\textwidth]{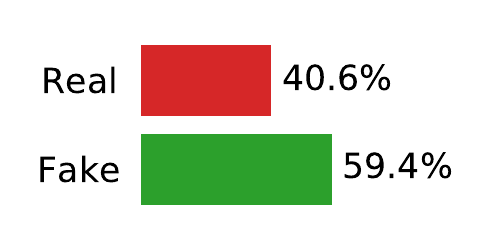}
\end{minipage}
\\ \vspace{0.4em}

\begin{minipage}[c]{0.195\linewidth}
\includegraphics[width=\textwidth]{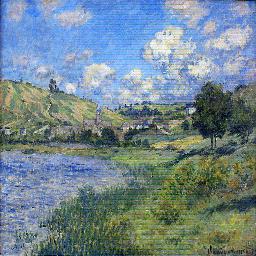}
\end{minipage}
\begin{minipage}[c]{0.195\linewidth}
\includegraphics[width=\textwidth]{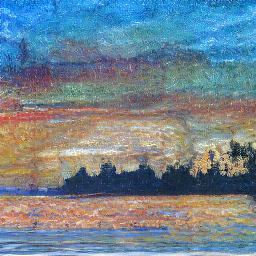}
\end{minipage}
\begin{minipage}[c]{0.195\linewidth}
\includegraphics[width=\textwidth]{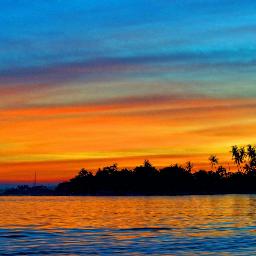}
\end{minipage}
\begin{minipage}[c]{0.195\linewidth}
\includegraphics[width=\textwidth]{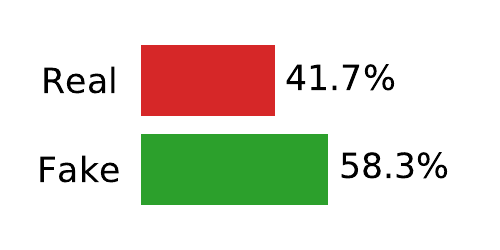}
\end{minipage}
\\ \vspace{0.4em}

\begin{minipage}[c]{0.195\linewidth}
\includegraphics[width=\textwidth]{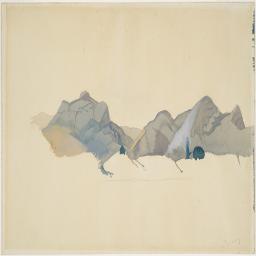}
\end{minipage}
\begin{minipage}[c]{0.195\linewidth}
\includegraphics[width=\textwidth]{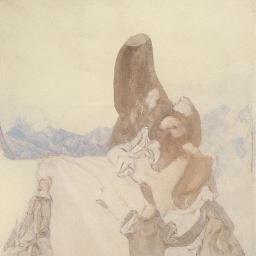}
\end{minipage}
\begin{minipage}[c]{0.195\linewidth}
\includegraphics[width=\textwidth]{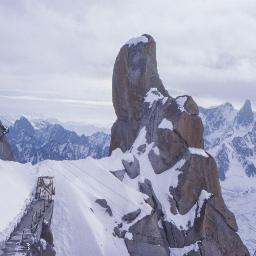}
\end{minipage}
\begin{minipage}[c]{0.195\linewidth}
\includegraphics[width=\textwidth]{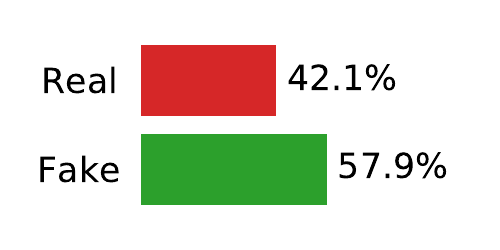}
\end{minipage}
\\ \vspace{0.4em}

\begin{minipage}[c]{0.195\linewidth}
\includegraphics[width=\textwidth]{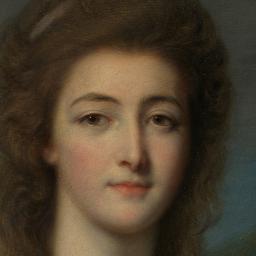}
\end{minipage}
\begin{minipage}[c]{0.195\linewidth}
\includegraphics[width=\textwidth]{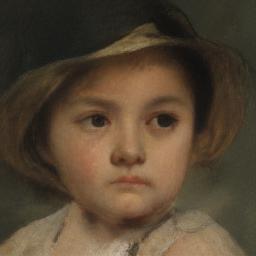}
\end{minipage}
\begin{minipage}[c]{0.195\linewidth}
\includegraphics[width=\textwidth]{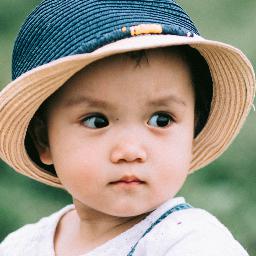}
\end{minipage}
\begin{minipage}[c]{0.195\linewidth}
\includegraphics[width=\textwidth]{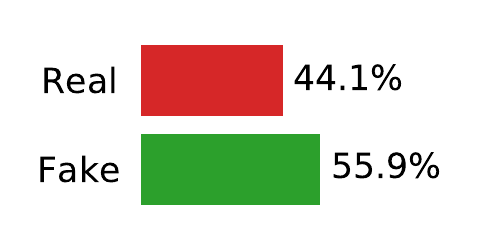}
\end{minipage}
\\ \vspace{0.4em}

\begin{minipage}[c]{0.195\linewidth}
\includegraphics[width=\textwidth]{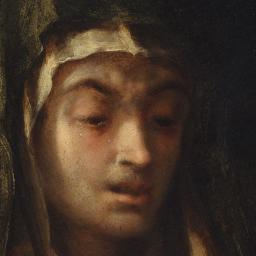}
\end{minipage}
\begin{minipage}[c]{0.195\linewidth}
\includegraphics[width=\textwidth]{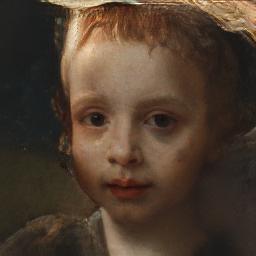}
\end{minipage}
\begin{minipage}[c]{0.195\linewidth}
\includegraphics[width=\textwidth]{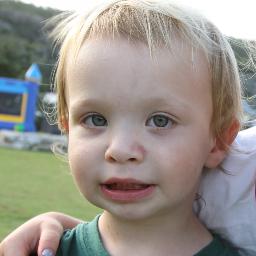}
\end{minipage}
\begin{minipage}[c]{0.195\linewidth}
\includegraphics[width=\textwidth]{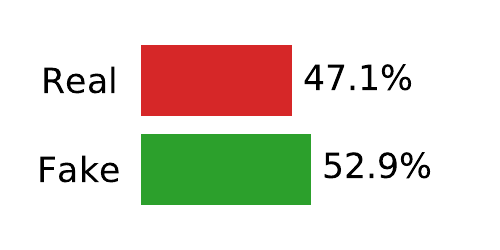}
\end{minipage}

\caption{Examples of artistic confusion test (2/6). The figures are sorted in an ascending order of the ratios of users choosing the real artwork. }
\label{fig:turing-2}
\end{figure*}

\begin{figure*}[t]
\centering
\begin{minipage}[c]{0.195\linewidth}
\centering \small Real Artwork
\end{minipage}
\begin{minipage}[c]{0.195\linewidth}
\centering \small Fake Artwork (Ours)
\end{minipage}
\begin{minipage}[c]{0.195\linewidth}
\centering \small Content Image
\end{minipage}
\begin{minipage}[c]{0.195\linewidth}
\centering \small User Choice
\end{minipage}
\\ \vspace{0.1em}

\begin{minipage}[c]{0.195\linewidth}
\includegraphics[width=\textwidth]{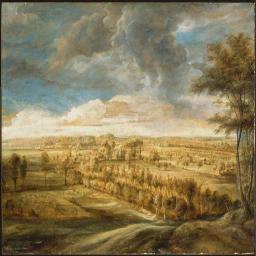}
\end{minipage}
\begin{minipage}[c]{0.195\linewidth}
\includegraphics[width=\textwidth]{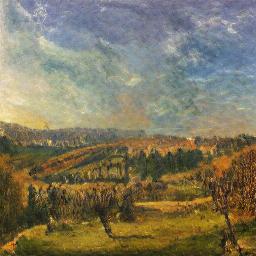}
\end{minipage}
\begin{minipage}[c]{0.195\linewidth}
\includegraphics[width=\textwidth]{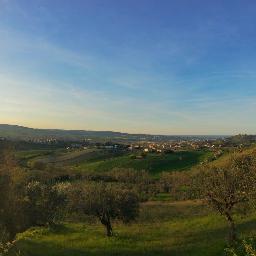}
\end{minipage}
\begin{minipage}[c]{0.195\linewidth}
\includegraphics[width=\textwidth]{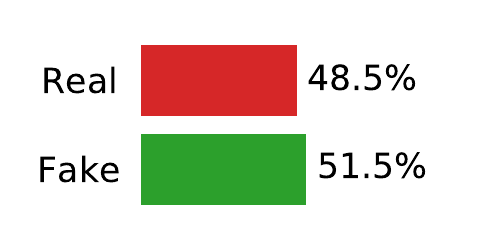}
\end{minipage}
\\ \vspace{0.4em}

\begin{minipage}[c]{0.195\linewidth}
\includegraphics[width=\textwidth]{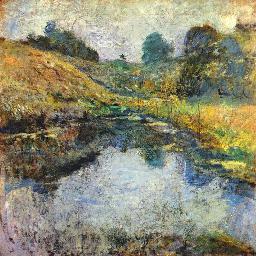}
\end{minipage}
\begin{minipage}[c]{0.195\linewidth}
\includegraphics[width=\textwidth]{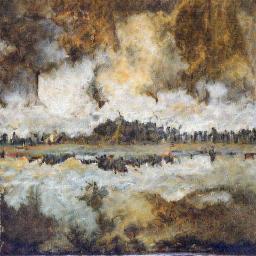}
\end{minipage}
\begin{minipage}[c]{0.195\linewidth}
\includegraphics[width=\textwidth]{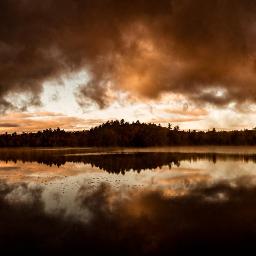}
\end{minipage}
\begin{minipage}[c]{0.195\linewidth}
\includegraphics[width=\textwidth]{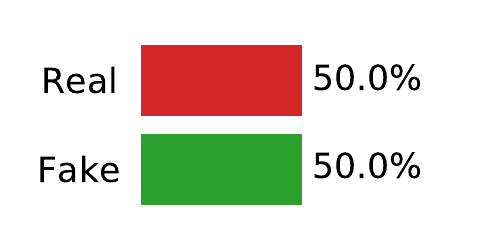}
\end{minipage}
\\ \vspace{0.4em}

\begin{minipage}[c]{0.195\linewidth}
\includegraphics[width=\textwidth]{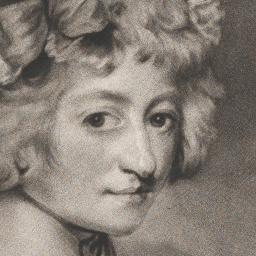}
\end{minipage}
\begin{minipage}[c]{0.195\linewidth}
\includegraphics[width=\textwidth]{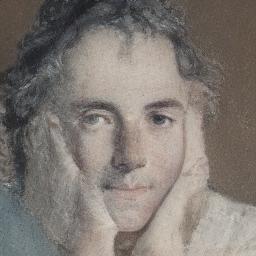}
\end{minipage}
\begin{minipage}[c]{0.195\linewidth}
\includegraphics[width=\textwidth]{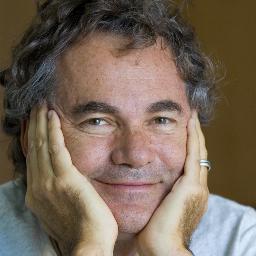}
\end{minipage}
\begin{minipage}[c]{0.195\linewidth}
\includegraphics[width=\textwidth]{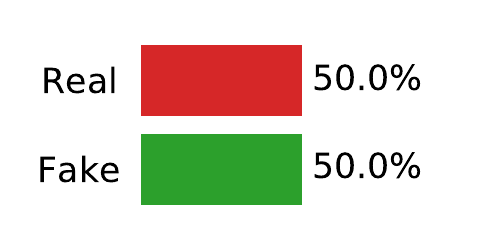}
\end{minipage}
\\ \vspace{0.4em}

\begin{minipage}[c]{0.195\linewidth}
\includegraphics[width=\textwidth]{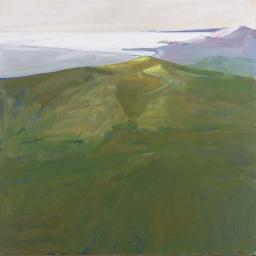}
\end{minipage}
\begin{minipage}[c]{0.195\linewidth}
\includegraphics[width=\textwidth]{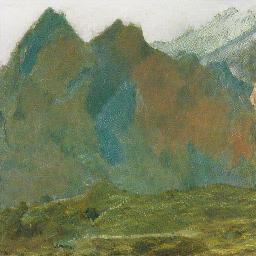}
\end{minipage}
\begin{minipage}[c]{0.195\linewidth}
\includegraphics[width=\textwidth]{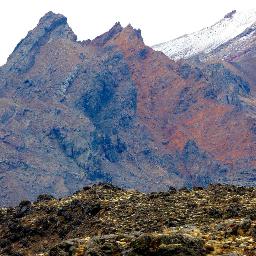}
\end{minipage}
\begin{minipage}[c]{0.195\linewidth}
\includegraphics[width=\textwidth]{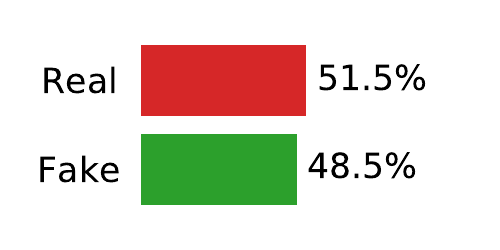}
\end{minipage}
\\ \vspace{0.4em}

\begin{minipage}[c]{0.195\linewidth}
\includegraphics[width=\textwidth]{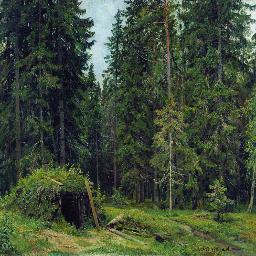}
\end{minipage}
\begin{minipage}[c]{0.195\linewidth}
\includegraphics[width=\textwidth]{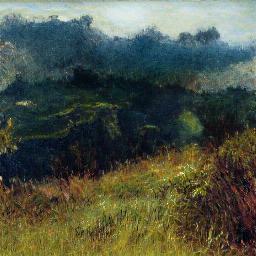}
\end{minipage}
\begin{minipage}[c]{0.195\linewidth}
\includegraphics[width=\textwidth]{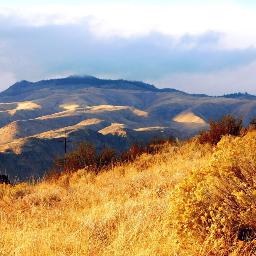}
\end{minipage}
\begin{minipage}[c]{0.195\linewidth}
\includegraphics[width=\textwidth]{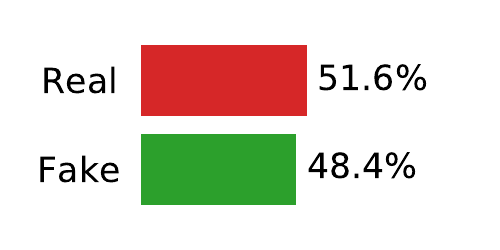}
\end{minipage}
\\ \vspace{0.4em}

\begin{minipage}[c]{0.195\linewidth}
\includegraphics[width=\textwidth]{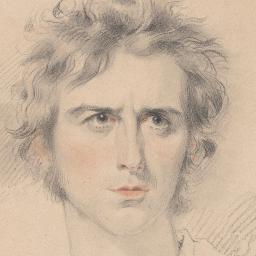}
\end{minipage}
\begin{minipage}[c]{0.195\linewidth}
\includegraphics[width=\textwidth]{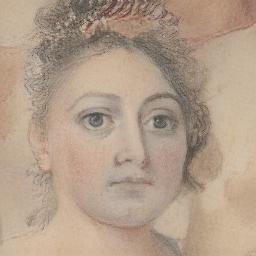}
\end{minipage}
\begin{minipage}[c]{0.195\linewidth}
\includegraphics[width=\textwidth]{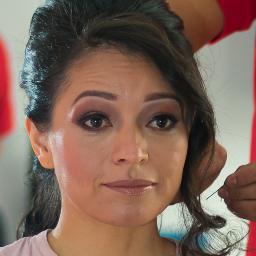}
\end{minipage}
\begin{minipage}[c]{0.195\linewidth}
\includegraphics[width=\textwidth]{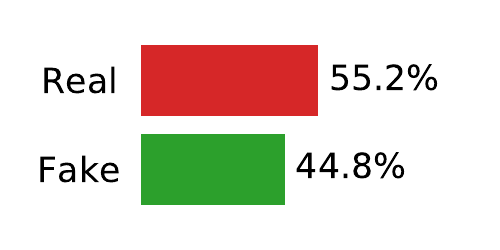}
\end{minipage}

\caption{Examples of artistic confusion test (3/6). The figures are sorted in an ascending order of the ratios of users choosing the real artwork. }
\label{fig:turing-3}
\end{figure*}

\begin{figure*}[t]
\centering
\begin{minipage}[c]{0.195\linewidth}
\centering \small Real Artwork
\end{minipage}
\begin{minipage}[c]{0.195\linewidth}
\centering \small Fake Artwork (Ours)
\end{minipage}
\begin{minipage}[c]{0.195\linewidth}
\centering \small Content Image
\end{minipage}
\begin{minipage}[c]{0.195\linewidth}
\centering \small User Choice
\end{minipage}
\\ \vspace{0.1em}

\begin{minipage}[c]{0.195\linewidth}
\includegraphics[width=\textwidth]{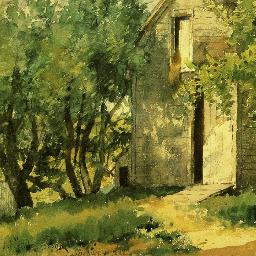}
\end{minipage}
\begin{minipage}[c]{0.195\linewidth}
\includegraphics[width=\textwidth]{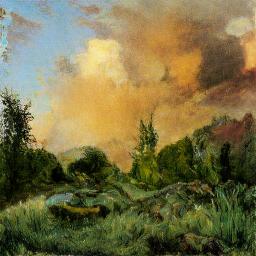}
\end{minipage}
\begin{minipage}[c]{0.195\linewidth}
\includegraphics[width=\textwidth]{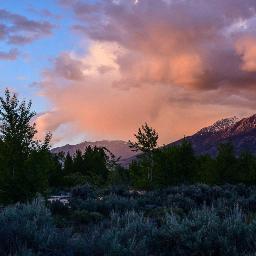}
\end{minipage}
\begin{minipage}[c]{0.195\linewidth}
\includegraphics[width=\textwidth]{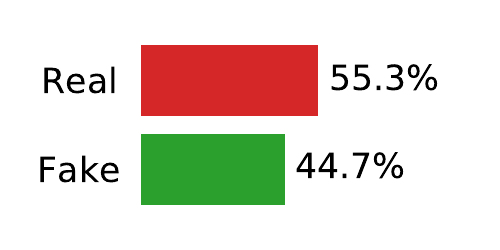}
\end{minipage}
\\ \vspace{0.4em}

\begin{minipage}[c]{0.195\linewidth}
\includegraphics[width=\textwidth]{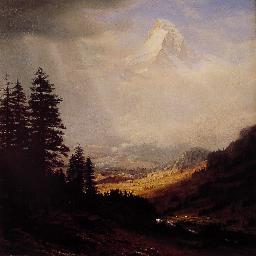}
\end{minipage}
\begin{minipage}[c]{0.195\linewidth}
\includegraphics[width=\textwidth]{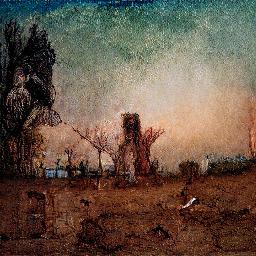}
\end{minipage}
\begin{minipage}[c]{0.195\linewidth}
\includegraphics[width=\textwidth]{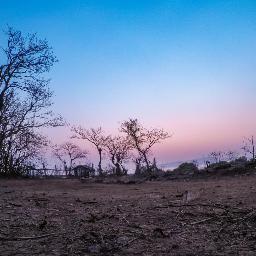}
\end{minipage}
\begin{minipage}[c]{0.195\linewidth}
\includegraphics[width=\textwidth]{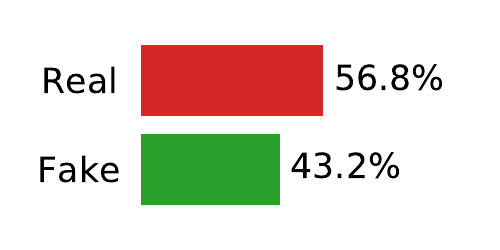}
\end{minipage}
\\ \vspace{0.4em}

\begin{minipage}[c]{0.195\linewidth}
\includegraphics[width=\textwidth]{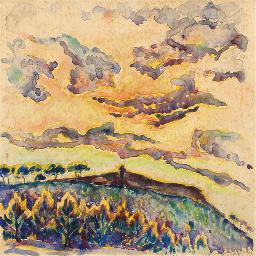}
\end{minipage}
\begin{minipage}[c]{0.195\linewidth}
\includegraphics[width=\textwidth]{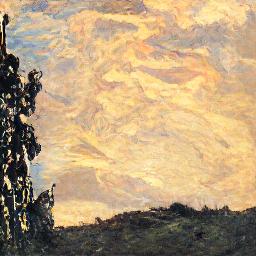}
\end{minipage}
\begin{minipage}[c]{0.195\linewidth}
\includegraphics[width=\textwidth]{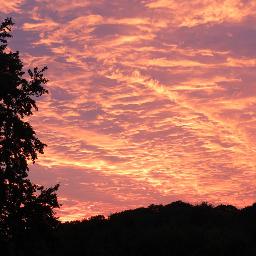}
\end{minipage}
\begin{minipage}[c]{0.195\linewidth}
\includegraphics[width=\textwidth]{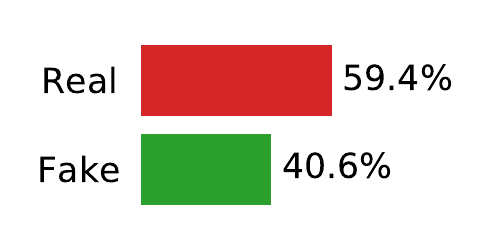}
\end{minipage}
\\ \vspace{0.4em}

\begin{minipage}[c]{0.195\linewidth}
\includegraphics[width=\textwidth]{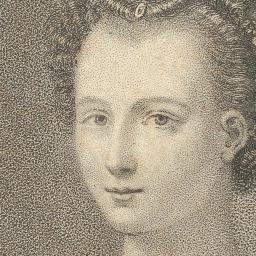}
\end{minipage}
\begin{minipage}[c]{0.195\linewidth}
\includegraphics[width=\textwidth]{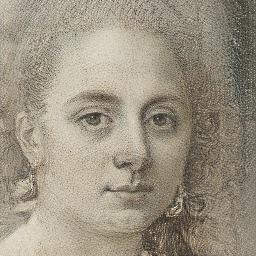}
\end{minipage}
\begin{minipage}[c]{0.195\linewidth}
\includegraphics[width=\textwidth]{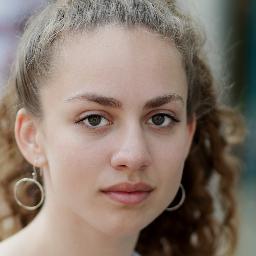}
\end{minipage}
\begin{minipage}[c]{0.195\linewidth}
\includegraphics[width=\textwidth]{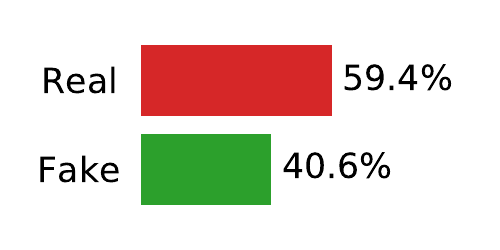}
\end{minipage}
\\ \vspace{0.4em}

\begin{minipage}[c]{0.195\linewidth}
\includegraphics[width=\textwidth]{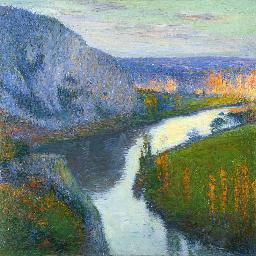}
\end{minipage}
\begin{minipage}[c]{0.195\linewidth}
\includegraphics[width=\textwidth]{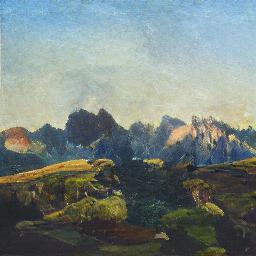}
\end{minipage}
\begin{minipage}[c]{0.195\linewidth}
\includegraphics[width=\textwidth]{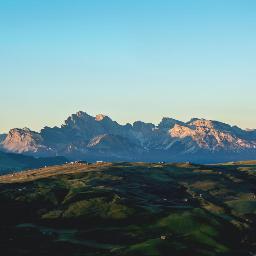}
\end{minipage}
\begin{minipage}[c]{0.195\linewidth}
\includegraphics[width=\textwidth]{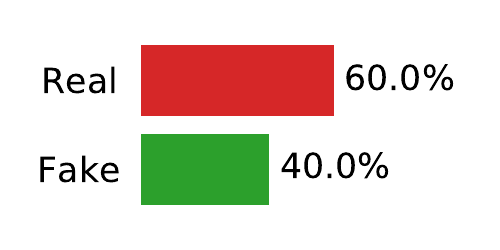}
\end{minipage}
\\ \vspace{0.4em}

\begin{minipage}[c]{0.195\linewidth}
\includegraphics[width=\textwidth]{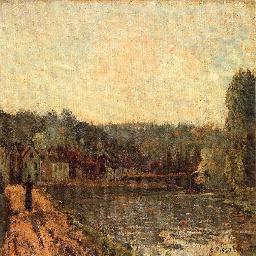}
\end{minipage}
\begin{minipage}[c]{0.195\linewidth}
\includegraphics[width=\textwidth]{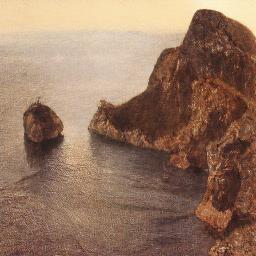}
\end{minipage}
\begin{minipage}[c]{0.195\linewidth}
\includegraphics[width=\textwidth]{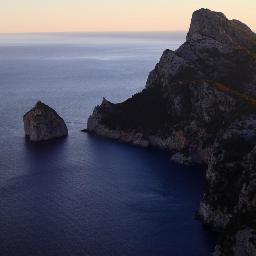}
\end{minipage}
\begin{minipage}[c]{0.195\linewidth}
\includegraphics[width=\textwidth]{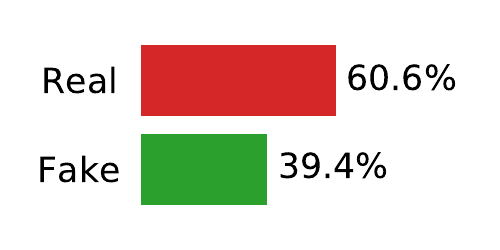}
\end{minipage}

\caption{Examples of artistic confusion test (4/6). The figures are sorted in an ascending order of the ratios of users choosing the real artwork. }
\label{fig:turing-4}
\end{figure*}

\begin{figure*}[t]
\centering
\begin{minipage}[c]{0.195\linewidth}
\centering \small Real Artwork
\end{minipage}
\begin{minipage}[c]{0.195\linewidth}
\centering \small Fake Artwork (Ours)
\end{minipage}
\begin{minipage}[c]{0.195\linewidth}
\centering \small Content Image
\end{minipage}
\begin{minipage}[c]{0.195\linewidth}
\centering \small User Choice
\end{minipage}
\\ \vspace{0.1em}

\begin{minipage}[c]{0.195\linewidth}
\includegraphics[width=\textwidth]{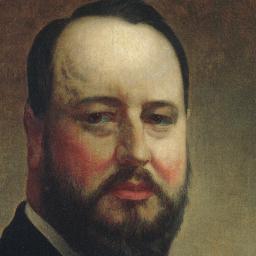}
\end{minipage}
\begin{minipage}[c]{0.195\linewidth}
\includegraphics[width=\textwidth]{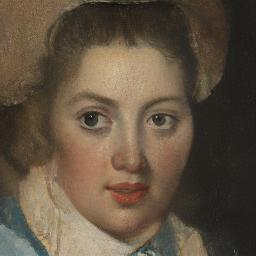}
\end{minipage}
\begin{minipage}[c]{0.195\linewidth}
\includegraphics[width=\textwidth]{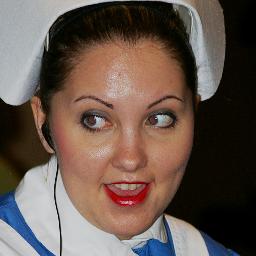}
\end{minipage}
\begin{minipage}[c]{0.195\linewidth}
\includegraphics[width=\textwidth]{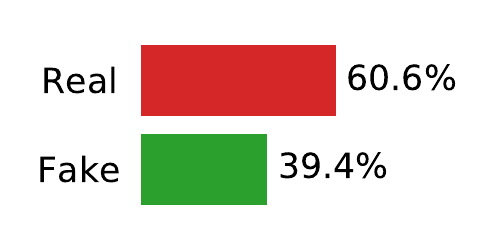}
\end{minipage}
\\ \vspace{0.4em}

\begin{minipage}[c]{0.195\linewidth}
\includegraphics[width=\textwidth]{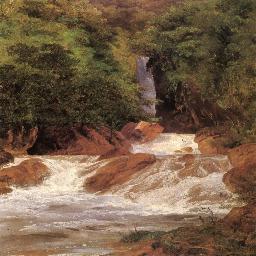}
\end{minipage}
\begin{minipage}[c]{0.195\linewidth}
\includegraphics[width=\textwidth]{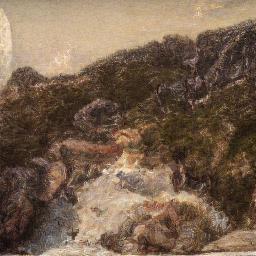}
\end{minipage}
\begin{minipage}[c]{0.195\linewidth}
\includegraphics[width=\textwidth]{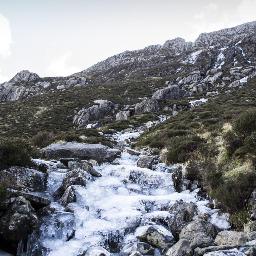}
\end{minipage}
\begin{minipage}[c]{0.195\linewidth}
\includegraphics[width=\textwidth]{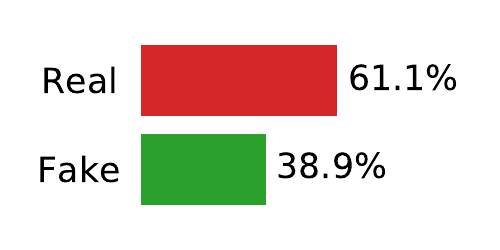}
\end{minipage}
\\ \vspace{0.4em}

\begin{minipage}[c]{0.195\linewidth}
\includegraphics[width=\textwidth]{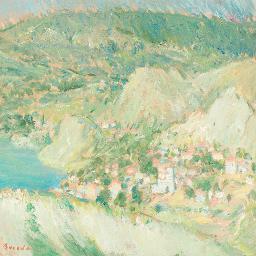}
\end{minipage}
\begin{minipage}[c]{0.195\linewidth}
\includegraphics[width=\textwidth]{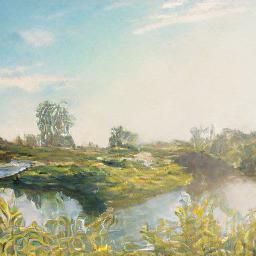}
\end{minipage}
\begin{minipage}[c]{0.195\linewidth}
\includegraphics[width=\textwidth]{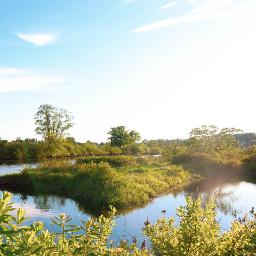}
\end{minipage}
\begin{minipage}[c]{0.195\linewidth}
\includegraphics[width=\textwidth]{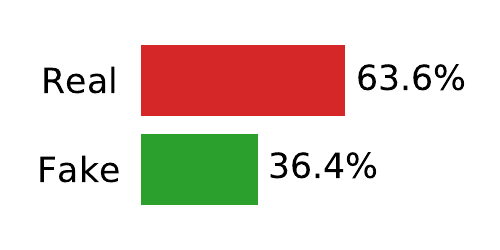}
\end{minipage}
\\ \vspace{0.4em}

\begin{minipage}[c]{0.195\linewidth}
\includegraphics[width=\textwidth]{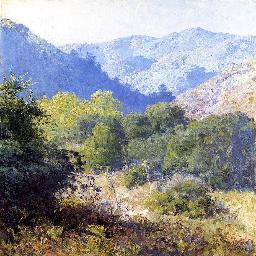}
\end{minipage}
\begin{minipage}[c]{0.195\linewidth}
\includegraphics[width=\textwidth]{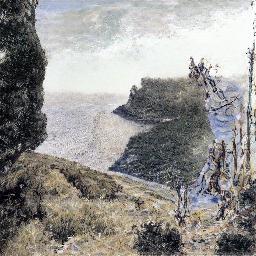}
\end{minipage}
\begin{minipage}[c]{0.195\linewidth}
\includegraphics[width=\textwidth]{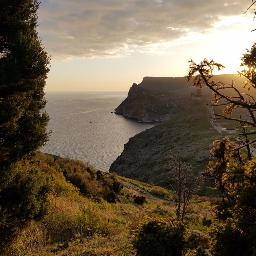}
\end{minipage}
\begin{minipage}[c]{0.195\linewidth}
\includegraphics[width=\textwidth]{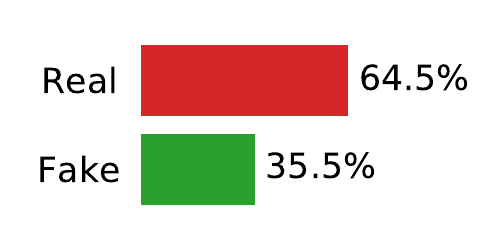}
\end{minipage}
\\ \vspace{0.4em}

\begin{minipage}[c]{0.195\linewidth}
\includegraphics[width=\textwidth]{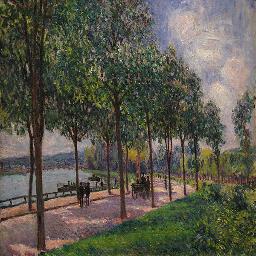}
\end{minipage}
\begin{minipage}[c]{0.195\linewidth}
\includegraphics[width=\textwidth]{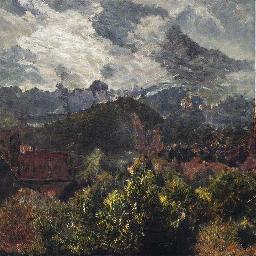}
\end{minipage}
\begin{minipage}[c]{0.195\linewidth}
\includegraphics[width=\textwidth]{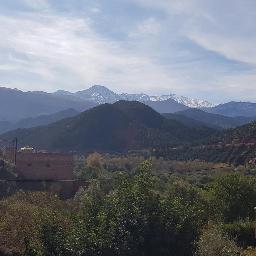}
\end{minipage}
\begin{minipage}[c]{0.195\linewidth}
\includegraphics[width=\textwidth]{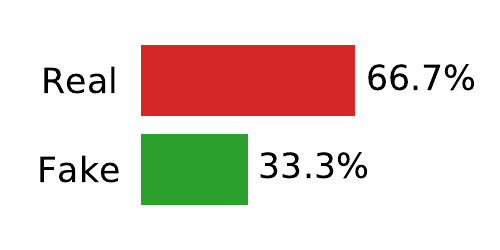}
\end{minipage}
\\ \vspace{0.4em}

\begin{minipage}[c]{0.195\linewidth}
\includegraphics[width=\textwidth]{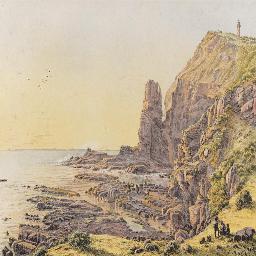}
\end{minipage}
\begin{minipage}[c]{0.195\linewidth}
\includegraphics[width=\textwidth]{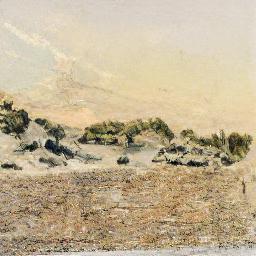}
\end{minipage}
\begin{minipage}[c]{0.195\linewidth}
\includegraphics[width=\textwidth]{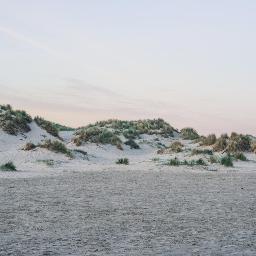}
\end{minipage}
\begin{minipage}[c]{0.195\linewidth}
\includegraphics[width=\textwidth]{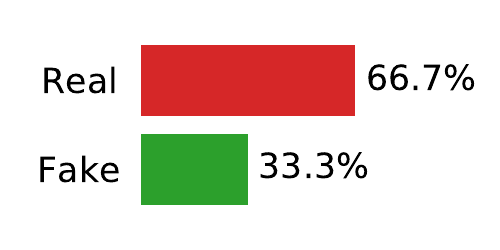}
\end{minipage}

\caption{Examples of artistic confusion test (5/6). The figures are sorted in an ascending order of the ratios of users choosing the real artwork. }
\label{fig:turing-5}
\end{figure*}

\begin{figure*}[t]
\centering
\begin{minipage}[c]{0.195\linewidth}
\centering \small Real Artwork
\end{minipage}
\begin{minipage}[c]{0.195\linewidth}
\centering \small Fake Artwork (Ours)
\end{minipage}
\begin{minipage}[c]{0.195\linewidth}
\centering \small Content Image
\end{minipage}
\begin{minipage}[c]{0.195\linewidth}
\centering \small User Choice
\end{minipage}
\\ \vspace{0.1em}

\begin{minipage}[c]{0.195\linewidth}
\includegraphics[width=\textwidth]{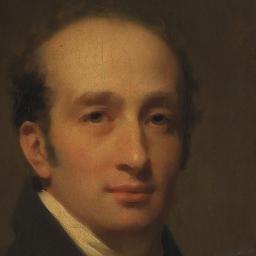}
\end{minipage}
\begin{minipage}[c]{0.195\linewidth}
\includegraphics[width=\textwidth]{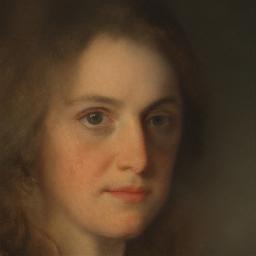}
\end{minipage}
\begin{minipage}[c]{0.195\linewidth}
\includegraphics[width=\textwidth]{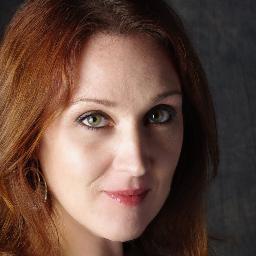}
\end{minipage}
\begin{minipage}[c]{0.195\linewidth}
\includegraphics[width=\textwidth]{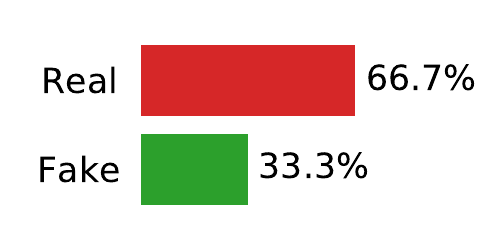}
\end{minipage}
\\ \vspace{0.4em}

\begin{minipage}[c]{0.195\linewidth}
\includegraphics[width=\textwidth]{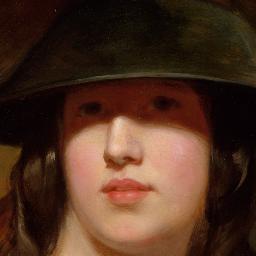}
\end{minipage}
\begin{minipage}[c]{0.195\linewidth}
\includegraphics[width=\textwidth]{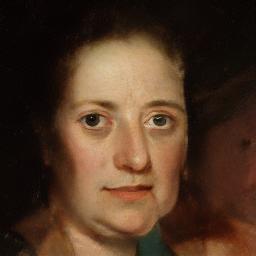}
\end{minipage}
\begin{minipage}[c]{0.195\linewidth}
\includegraphics[width=\textwidth]{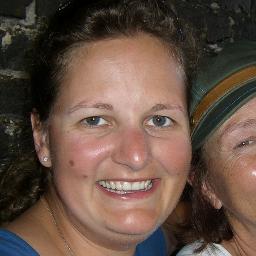}
\end{minipage}
\begin{minipage}[c]{0.195\linewidth}
\includegraphics[width=\textwidth]{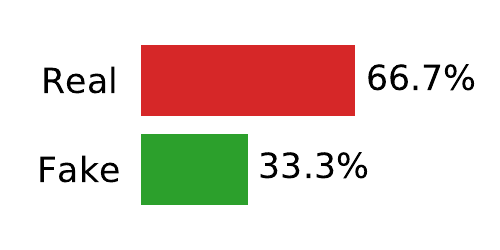}
\end{minipage}
\\ \vspace{0.4em}

\begin{minipage}[c]{0.195\linewidth}
\includegraphics[width=\textwidth]{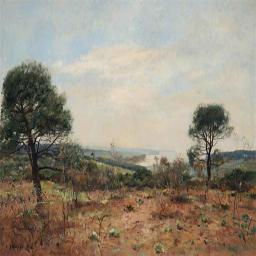}
\end{minipage}
\begin{minipage}[c]{0.195\linewidth}
\includegraphics[width=\textwidth]{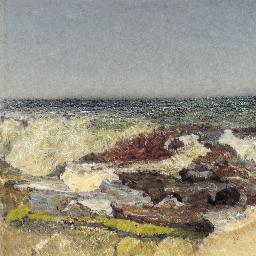}
\end{minipage}
\begin{minipage}[c]{0.195\linewidth}
\includegraphics[width=\textwidth]{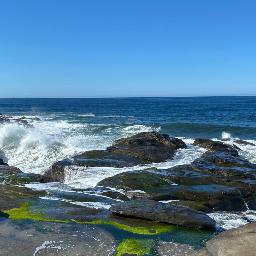}
\end{minipage}
\begin{minipage}[c]{0.195\linewidth}
\includegraphics[width=\textwidth]{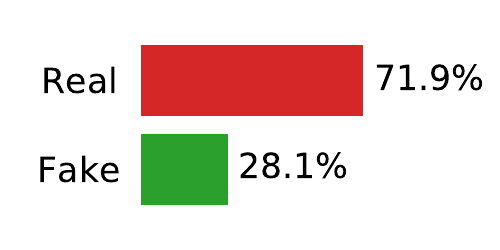}
\end{minipage}
\\ \vspace{0.4em}

\begin{minipage}[c]{0.195\linewidth}
\includegraphics[width=\textwidth]{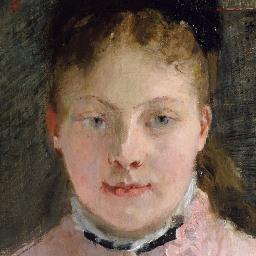}
\end{minipage}
\begin{minipage}[c]{0.195\linewidth}
\includegraphics[width=\textwidth]{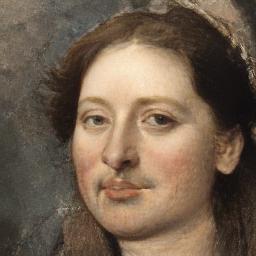}
\end{minipage}
\begin{minipage}[c]{0.195\linewidth}
\includegraphics[width=\textwidth]{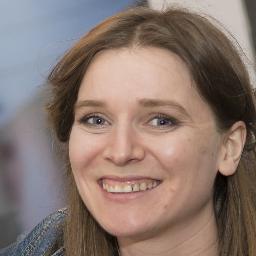}
\end{minipage}
\begin{minipage}[c]{0.195\linewidth}
\includegraphics[width=\textwidth]{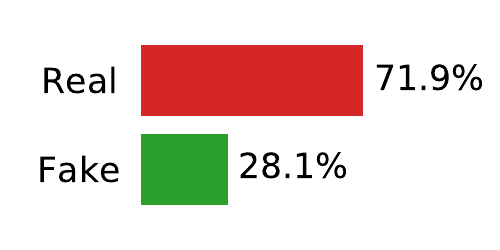}
\end{minipage}
\\ \vspace{0.4em}

\begin{minipage}[c]{0.195\linewidth}
\includegraphics[width=\textwidth]{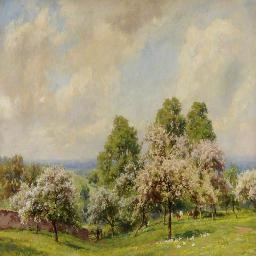}
\end{minipage}
\begin{minipage}[c]{0.195\linewidth}
\includegraphics[width=\textwidth]{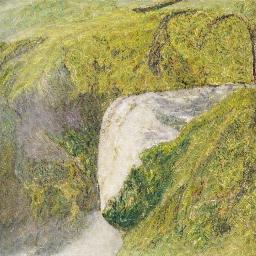}
\end{minipage}
\begin{minipage}[c]{0.195\linewidth}
\includegraphics[width=\textwidth]{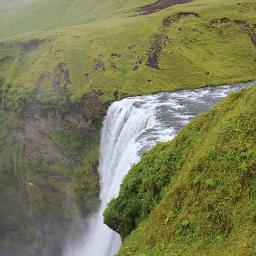}
\end{minipage}
\begin{minipage}[c]{0.195\linewidth}
\includegraphics[width=\textwidth]{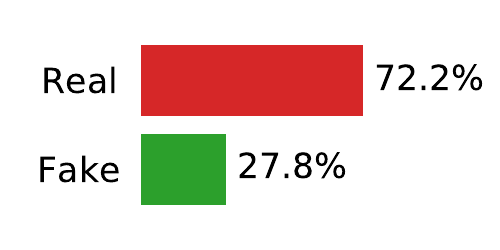}
\end{minipage}
\\ \vspace{0.4em}

\begin{minipage}[c]{0.195\linewidth}
\includegraphics[width=\textwidth]{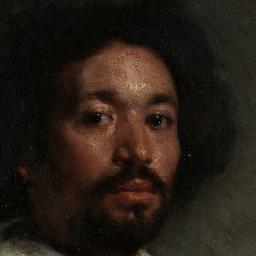}
\end{minipage}
\begin{minipage}[c]{0.195\linewidth}
\includegraphics[width=\textwidth]{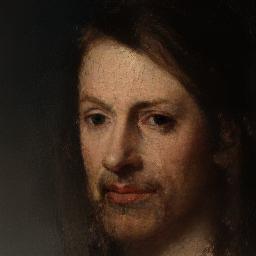}
\end{minipage}
\begin{minipage}[c]{0.195\linewidth}
\includegraphics[width=\textwidth]{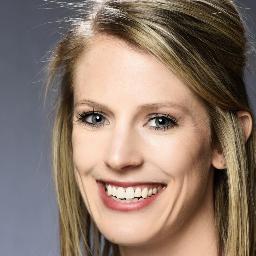}
\end{minipage}
\begin{minipage}[c]{0.195\linewidth}
\includegraphics[width=\textwidth]{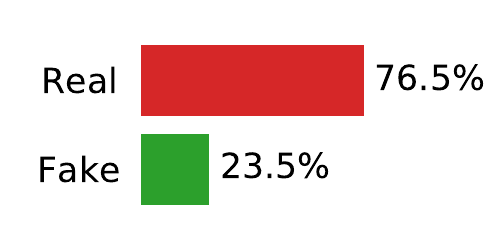}
\end{minipage}

\caption{Examples of artistic confusion test (6/6). The figures are sorted in an ascending order of the ratios of users choosing the real artwork. }
\label{fig:turing-6}
\end{figure*}

We present all 36 examples and corresponding user feedbacks of our Artistic Style Transfer Confusion Test in Figs. \ref{fig:turing-1}, \ref{fig:turing-2}, \ref{fig:turing-3}, \ref{fig:turing-4}, \ref{fig:turing-5}, and \ref{fig:turing-6}. The ratios of correct selections with respect to individual examples range from 19.4\% to 76.5\% with an overall average of 51.6\%. The results indicate that our QuantArt method can generate highly realistic artistic images, and most of the generations are difficult for humans to identify from the real artworks. 

\begin{table*}[h]
\centering
\caption{Quantitative comparison of the state-of-the-art methods for \textbf{artwork-to-artwork} image style transfer. The \textbf{best} and \underline{second best} results are highlighted, respectively.}%
\vspace{-1em}
     \resizebox{1\hsize}{!}{
    \begin{tabular}{lccccccccccccccc}
        \hline
       \multirow{2}{*}{Metric $\downarrow$} & \multirow{2}{*}{AdaIN} &  \multirow{2}{*}{WCT} &  \multirow{2}{*}{LinearWCT} & \multirow{2}{*}{ArtFlow} &  \multirow{2}{*}{EFDM} & \multirow{2}{*}{StyleSwap} &  \multirow{2}{*}{AvatarNet} &  \multirow{2}{*}{SANet} &  \multirow{2}{*}{AdaAttN} &  \multirow{2}{*}{StyTR2} & \multicolumn{3}{c}{Ours ($\alpha,\beta$)}\\ \cline{12-14}
        &&&&&&&&&&&$(0,0)$&$(0,1)$&$(1,1)$\\\hline
        LPIPS $\downarrow$ & 0.600 & 0.695 & 0.591 & 0.522 & 0.630 & 0.616 & 0.725 & 0.622 & 0.587 & 0.494 & \textit{0.264} & \textbf{0.407} & \underline{0.437} \\ 
        Gram loss\begin{scriptsize} (${\times 10^{3}}$) \end{scriptsize}  $\downarrow$  & 0.159 & 0.267 & 0.160 & \textbf{0.105} & 0.395 & 1.367 & 0.712 & \underline{0.123} & 0.242 & 0.384 & - & 0.677 & 0.965 \\
        FID $\downarrow$ & 31.668 & 73.170 & 42.248 & 25.642 & 47.264 & 66.375 & 50.234 & 25.721 & 24.827 & \underline{14.363} & - & 15.763 & \textbf{10.973} \\ 
        ArtFID $\downarrow$ & 52.266 & 125.719 & 68.793 & 40.544 & 78.663 & 108.878 & 88.369 & 43.343 & 40.980 & \underline{22.959} & - & 23.587 & \textbf{17.209} \\ \hline 
    \end{tabular}
    }
    \label{tab:art2art}
\end{table*}

\begin{table*}[h]
\centering
\vspace{-1em}
\caption{Quantitative comparison of the state-of-the-art methods for \textbf{photo-to-photo} image style transfer.}%
\vspace{-1em}
     \resizebox{0.66\hsize}{!}{
    \begin{tabular}{lccccccc}
        \hline
       \multirow{2}{*}{Metric $\downarrow$} & \multirow{2}{*}{PhotoWCT} &  \multirow{2}{*}{WCT2} &  \multirow{2}{*}{LinearWCT} & \multirow{2}{*}{DSTN} & \multicolumn{3}{c}{Ours ($\alpha,\beta$)}\\ \cline{6-8}
        &&&&&$(0,0)$&$(0,1)$&$(1,1)$\\\hline
        LPIPS $\downarrow$ & 0.509 & \textbf{0.348} & 0.353 & 0.415 & \textit{0.158} & 0.494 & 0.388 \\ 
        Gram loss\begin{scriptsize} (${\times 10^{3}}$) \end{scriptsize}  $\downarrow$  & 0.650 & 0.372 & 0.499 & 0.460 & - & \textbf{0.355} & 1.114 \\
        FID $\downarrow$ & 15.462 & 8.946 & 17.775 & 20.583 & - & 19.651 & \textbf{4.261} \\ 
        ArtFID $\downarrow$ & 24.848 & 13.409 & 25.402 & 30.536 & - & 30.847 & \textbf{7.300} \\ \hline 
    \end{tabular}
    }
    \label{tab:real2real}
\end{table*}

\begin{table*}[h]
\centering
\vspace{-1em}
\caption{Quantitative comparison of the state-of-the-art methods for \textbf{artwork-to-photo} image style transfer.}%
\vspace{-1em}
     \resizebox{0.53\hsize}{!}{
    \begin{tabular}{lccccccc}
        \hline
       \multirow{2}{*}{Metric $\downarrow$} & \multirow{2}{*}{CycleGAN} & \multirow{2}{*}{CUT} &  \multirow{2}{*}{DRIT} &  \multirow{2}{*}{Art2Real} & \multicolumn{2}{c}{Ours}\\ \cline{6-7}
        &&&&&$\alpha=0$&$\alpha=1$\\\hline
        LPIPS $\downarrow$ & \textbf{0.226} & 0.427 & 0.573 & 0.280 & 0.457 & 0.457 \\ 
        FID  $\downarrow$  & 42.899 & 35.237 & 41.419 & 34.608 & 31.002 & \textbf{20.864} \\
        ArtFID $\downarrow$ & 53.827 & 51.724 & 66.717 & 45.585 & 46.641 & \textbf{31.847} \\ \hline 
    \end{tabular}
    }
    \label{tab:art2real}
\end{table*}

\clearpage

\clearpage

\clearpage

\end{document}